\documentclass[lettersize,journal]{IEEEtran}
\usepackage{amsmath,amsfonts}
\usepackage{algorithm}
\usepackage{algpseudocode}

\usepackage{caption}
\usepackage{hyperref}

\usepackage{array}
\usepackage[dvipsnames]{xcolor}

\usepackage{textcomp}
\usepackage{stfloats}
\usepackage{url}
\usepackage{verbatim}
\usepackage{graphicx}
\usepackage{cite}

\usepackage{booktabs}
\usepackage{epsfig}
\usepackage{enumitem}
\usepackage{multirow}
\usepackage{subcaption}
\usepackage{wrapfig}
\usepackage{placeins}
\usepackage{setspace}
\usepackage{microtype}
\usepackage[american]{babel}

\usepackage{bm}
\usepackage{color}
\usepackage[most]{tcolorbox}
\usepackage{cuted}

\newtcolorbox{finding}[1][]{
  colback=black!5!white,    
  colframe=black!60!white,  
  fonttitle=\bfseries,     
  title=Finding,           
  arc=4pt,                 
  outer arc=4pt,
  #1                       
}

\definecolor{cvprblue}{HTML}{DAE8F5} 
\definecolor{cvprgray}{HTML}{EAEAEA}
\newtcolorbox{findingbox}{
    enhanced,
    colback=gray!8,          
    colframe=cvprblue,  
    arc=3pt,                 
    boxrule=1pt,           
    drop shadow=gray!50,     
    left=4pt, right=4pt, top=2pt, bottom=2pt, 
    fontupper=\small 
}

\graphicspath{{figure/}}

\DeclareCaptionLabelSeparator{myperiod}{.\hspace{1.1em}}
\DeclareCaptionLabelFormat{ieeetable}{TABLE~\Roman{table}}
\begin{document}

\title{Benchmarking Large Vision-Language Models on Fine-Grained Image Tasks: From Evaluation to Diagnosis}

\author{Hong-Tao~Yu, Chen-Wei~Xie, Yuxin~Peng,~\IEEEmembership{Fellow,~IEEE}, Serge~Belongie, and~Xiu-Shen~Wei,~\IEEEmembership{Senior Member,~IEEE}%
\IEEEcompsocitemizethanks{\IEEEcompsocthanksitem {Hong-Tao Yu is with the School of Computer Science and Engineering, Southeast University, China. E-mail: yuht\_seu@seu.edu.cn.}
\IEEEcompsocthanksitem Chen-Wei~Xie is with Alibaba Group. E-mail: xiecw.mail@gmail.com.
\IEEEcompsocthanksitem {Xiu-Shen Wei is with the School of Computer Science and Engineering, School of Intelligence Science and Engineering, and Key Laboratory of New Generation Artificial Intelligence Technology and Its Interdisciplinary Applications, Southeast University, China. E-mail: weixs@seu.edu.cn.}
\IEEEcompsocthanksitem {Yuxin Peng is with the Wangxuan Institute of Computer Technology, National
Key Laboratory for Multimedia Information Processing, Peking University, China. E-mail: pengyuxin@pku.edu.cn.}
\IEEEcompsocthanksitem {Serge  Belongie is with the University of Copenhagen, Denmark. E-mail: s.belongie@di.ku.dk.}
\IEEEcompsocthanksitem {Xiu-Shen Wei is the corresponding author.}}
}

\markboth{SUBMITTED TO IEEE TPAMI}%
{Yu~\MakeLowercase{\textit{et~al.}}: Benchmarking Large Vision-Language Models on Fine-Grained Image Tasks}

\IEEEtitleabstractindextext{%
\begin{abstract}
Recent advancements in Large Vision-Language Models (LVLMs) have demonstrated remarkable multimodal perception and reasoning capabilities. While numerous benchmarks have evaluated LVLMs from holistic or task-specific perspectives, their capabilities on fine-grained image tasks---fundamental to computer vision---remain insufficiently understood. To address this gap, we introduce \texttt{FG-BMK}, a comprehensive fine-grained evaluation benchmark containing 1.01 million questions and 0.28 million images, covering diverse scenarios from common object-centric domains to specialized domains. \texttt{FG-BMK} jointly evaluates dialogue-level fine-grained semantic recognition and feature-level visual discriminability through human-oriented and machine-oriented paradigms, enabling diagnostic analysis of whether LVLM failures arise from insufficient visual representations, weak visual-to-semantic grounding, or limited fine-grained knowledge. Through extensive experiments on a diverse set of representative LVLMs/VLMs, we find that current LVLMs remain inadequate fine-grained recognizers, with failures arising from intertwined bottlenecks in visual representations, semantic grounding, modality alignment, and category-level knowledge. We further analyze training design factors for improving fine-grained capabilities and examine how visual and linguistic perturbations affect LVLM predictions. These findings provide diagnostic insights into the limitations of current LVLMs and offer guidance for future data construction and model design in developing more reliable LVLMs for fine-grained visual tasks. Our code is open-source and available at \url{https://fg-bmk.github.io/}.
\end{abstract}

}

\maketitle


\IEEEdisplaynontitleabstractindextext
\IEEEpeerreviewmaketitle

\section{Introduction}

\IEEEPARstart{L}{arge} language models (LLMs) have made substantial progress in recent years, with models such as GPT~\cite{GPT4} exhibiting strong language understanding, reasoning, and generation abilities across a broad range of tasks. These advances have further stimulated the development of Large Vision-Language Models (LVLMs), which extend language-centric intelligence toward multimodal perception and interaction. Representative models, including GPT-5.4~\cite{GPT4}, Qwen~\cite{Qwen-VL}, InternVL~\cite{chen2024internvl}, and LLaVA-1.5~\cite{LLaVA15}, have achieved impressive performance in multimodal perception and reasoning. More recently, unified multimodal models have further expanded this paradigm by integrating visual understanding and generation within a single framework to make these capabilities mutually reinforcing.

\begin{figure*}[t!]
    \centering
    {\includegraphics[width=0.99\textwidth]{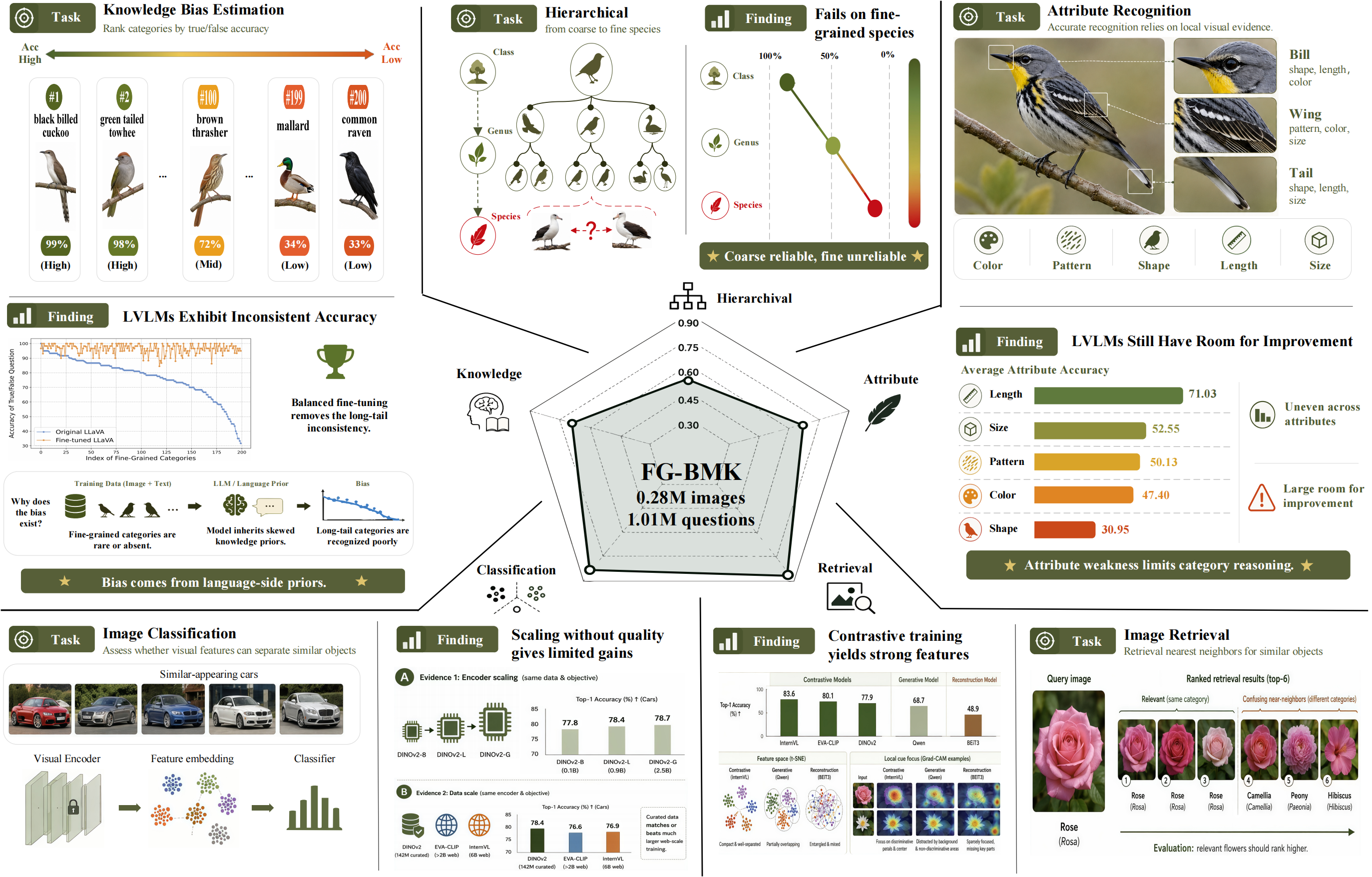}}
    \caption{Overview of \texttt{FG-BMK}. \texttt{FG-BMK} evaluates LVLMs on fine-grained visual tasks from five diagnostic dimensions: hierarchical recognition, knowledge bias estimation, attribute recognition, image classification, and image retrieval. The teaser illustrates both the task formats and representative findings, showing that current LVLMs still suffer from degraded fine-level recognition, biased category knowledge, uneven attribute understanding, and insufficient fine-grained visual discriminability.}
    \label{fig:teaser}
    \vspace{-1.3em}
\end{figure*}

These rapid advances have also driven increasingly systematic evaluations of LVLM capabilities. Existing holistic and specialized benchmarks have been proposed to examine LVLMs from different perspectives. For instance, LVLM-eHub~\cite{Lvlmehub} and MMBench~\cite{MMBench} offer broad evaluations of multimodal perception and reasoning, whereas specialized evaluations such as DocVQA~\cite{DocVQA} and GQA~\cite{GQA} target specific tasks, including document visual perception and visual reasoning. More recently, several studies~\cite{geigle2024african,zhang2024visually,tan2025vision} have begun to examine LVLMs on fine-grained image tasks, which require analyzing visual objects at the subordinate-category level and are fundamental to computer vision~\cite{XSsurveyPAMI}. However, these evaluations remain limited in scope, mainly focusing on classification-style tasks with limited domain diversity, task coverage, and diagnostic depth. As a result, the capability boundaries of LVLMs in fine-grained tasks remain poorly understood.

To address this gap, we introduce \texttt{FG-BMK}, a comprehensive benchmark for evaluating LVLMs on fine-grained image tasks. The benchmark contains 1.01 million questions and 0.28 million images, covering diverse fine-grained scenarios from common object-centric domains to specialized domains. Rather than treating fine-grained evaluation as a single classification problem, \texttt{FG-BMK} is organized around two complementary paradigms: human-oriented and machine-oriented evaluation. The human-oriented evaluation uses dialogue-like questions to assess fine-grained semantic recognition, including attribute perception, category-level knowledge bias, and hierarchical granularity understanding. The machine-oriented evaluation directly probes visual representations through two core fine-grained vision tasks---image retrieval and image recognition---measuring whether LVLM visual features preserve fine-grained similarity and category separability. By jointly examining dialogue-level semantic recognition and feature-level visual discriminability, \texttt{FG-BMK} enables a more diagnostic evaluation of whether LVLM failures arise from insufficient visual representations, weak visual-to-semantic grounding, or insufficient domain-specific or fine-grained category knowledge.

Building on the diagnostic design of \texttt{FG-BMK}, we organize our evaluation as a progressive analysis of fine-grained LVLM capabilities, rather than merely reporting aggregate benchmark scores. We begin by asking whether current LVLMs can serve as reliable fine-grained recognizers. To this end, we evaluate their performance across different taxonomy granularities, compare them with fine-grained tailored models, and further examine their ability to recognize discriminative visual attributes.

We then move from measuring this gap to diagnosing its underlying causes. By jointly considering dialogue-level semantic recognition and feature-level visual discriminability, we distinguish whether LVLM failures arise from insufficient visual representations, weak visual-to-semantic grounding, or limited fine-grained knowledge. We further investigate this issue through unified understanding-generation models, visual-to-textual alignment analysis, and category-level long-tail behavior, revealing how visual representations, semantic grounding, alignment strategies, and training-data coverage jointly shape fine-grained recognition performance.

Beyond failure diagnosis, we examine which training design factors can improve fine-grained LVLM capabilities. Specifically, we analyze how training objectives, visual feature quality, vision-encoder scale, training-data scale, and supervised fine-tuning data composition affect fine-grained visual discriminability and downstream recognition. Finally, we evaluate the robustness of fine-grained LVLM recognition under visual and linguistic perturbations, testing whether these capabilities remain stable when visual evidence is degraded or misleading language priors are introduced. Overall, this evaluation protocol moves from performance assessment to failure diagnosis, improvement analysis, and robustness verification, leading to the following key findings:

\begin{itemize}[leftmargin=*, itemsep=0pt, topsep=0pt]

\item The contrastive training paradigm in LVLMs proves more effective in enhancing the fine-grained discriminability of visual features, whereas generative and reconstruction-based training paradigms tend to yield weaker discriminability.

\item Aligning visual features with textual features in LVLMs can impair their fine-grained discriminability when image-text granularity is mismatched; however, content-level alignment improves general visual understanding, whereas category-level alignment strengthens fine-grained semantic grounding.

\item LVLMs and LVMs are more vulnerable to feature perturbations in fine-grained tasks than in generic vision tasks, while language-side perturbations can override visual evidence more effectively than visual-side perturbations.

\item LVLMs demonstrate relatively stronger capabilities in perceiving visual appearances but face challenges in fine-grained category reasoning (which depends on the recognition of visual attributes).

\item Unified understanding-generation models can exhibit fine-grained visual discriminability without truly grounding fine-grained category concepts, as their category-conditioned generations often miss defining visual characteristics.

\item In specialized domains such as remote sensing, semantic understanding rather than visual discrimination becomes the major bottleneck of LVLMs.

\item Despite their advancements, LVLMs still lag behind fine-grained tailored models in handling fine-grained visual tasks.

\end{itemize}

Note that a preliminary version of this work was published as a conference paper~\cite{FG-BMK} in the International Conference on Learning Representations (ICLR) 2026. In this journal version, we make substantial extensions in both evaluation coverage and diagnostic depth. Rather than simply extending the benchmark results, we reorganize the evaluation into a progressive diagnostic framework that moves from capability assessment to failure diagnosis, training-factor analysis, and robustness verification. More specifically, we expand the evaluation scope to more diverse and recent model architectures, including unified understanding-generation models, as well as specialized fine-grained domains, revealing new limitations in fine-grained concept grounding and domain-specific semantic understanding. Second, we design complementary qualitative analyses from both global and local perspectives, providing intuitive evidence of how different training paradigms shape fine-grained category separability and discriminative visual cues. Third, we extend the alignment analysis from a simple feature comparison to a controlled study of alignment-data granularity, revealing how textual supervision at different granularities shapes visual feature quality and downstream capabilities. Fourth, we further analyze how instruction-tuning data composition affects fine-grained capability, showing that a balanced mixture of general and fine-grained instruction data enables LVLMs to acquire fine-grained recognition ability while preserving general multimodal capabilities. Finally, we expand the robustness study across feature, image, and language levels, revealing how different perturbations affect fine-grained LVLM predictions and showing that language-side priors can more easily override visual evidence. Together, these extensions advance \texttt{FG-BMK} from a benchmark-centered evaluation toward a more comprehensive diagnostic study of LVLMs on fine-grained visual tasks.

\section{Related Work}

We provide a concise review of the relevant literature in three main areas: large vision-language model development, benchmark evaluation for LVLMs, and fine-grained image tasks, which respectively contextualize the evaluated models, existing evaluation protocols, and the visual challenges targeted by our benchmark.

\subsection{Large Vision-Language Models} 

Large Language Models (LLMs), exemplified by GPT-5.4~\cite{GPT4}, have shown substantial progress in text comprehension, reasoning, and generation. Extending this progress beyond language, Large Vision-Language Models (LVLMs) have developed strong multimodal perception and reasoning abilities across a wide range of tasks. Existing LVLMs and vision-language foundation models enhance multimodal capabilities through different technical routes. BLIP~\cite{BLIP} leverages noisy web data with bootstrapped captions for vision-language pre-training, while BLIP-2~\cite{BLIP2} bridges frozen image encoders and large language models through a lightweight querying module. LLaVA~\cite{LLaVA15} introduces visual instruction tuning with GPT-generated multimodal instruction data to enable effective visual-language interaction. The Qwen-VL series~\cite{Qwen-VL} extends Qwen language models with visual receptors and multi-stage multimodal training, where early image-text pre-training optimizes visual components via a generative language-modeling objective. Later variants improve dynamic-resolution perception, spatial-temporal modeling, and long-context interleaved understanding. The InternVL series~\cite{chen2024internvl,zhu2025internvl3} scales multimodal learning with large vision encoders and integrated multimodal pre-training, with recent versions further improving reasoning and efficiency through advanced post-training and inference recipes. In parallel, BEiT3~\cite{BEiT3} treats images as a foreign language and performs masked data modeling over images, texts, and image-text pairs with a shared multimodal backbone. More recently, unified multimodal models, such as BLIP3-o~\cite{blip3o}, UniWorld-V1~\cite{UniWorld-V1}, and BAGEL~\cite{bagel}, further integrate visual understanding and generation within a single framework. Despite these advances, most existing evaluations still emphasize general multimodal perception, reasoning, or generation, leaving their capabilities on fine-grained visual tasks less comprehensively understood.

\subsection{Large Vision-Language Model Benchmarks}


Alongside the rapid progress of LVLMs, numerous benchmarks have been introduced to characterize their multimodal capabilities from different perspectives. General and holistic benchmarks, such as LVLM-eHub~\cite{Lvlmehub} and MMBench~\cite{MMBench}, aim to provide broad assessments of multimodal perception, reasoning, and instruction-following abilities. In addition, task-specific benchmarks focus on particular capabilities or application scenarios. For example, ChartQA~\cite{ChartQA} evaluates chart understanding, DocVQA~\cite{DocVQA} focuses on document visual question answering, GQA~\cite{GQA} assesses compositional visual reasoning, CAPability~\cite{CAPability} evaluates image captioning quality, and OCRBench~\cite{OCRBench} measures optical character recognition ability. Other benchmarks, such as MathVista~\cite{lu2024mathvista} and MMMU~\cite{MMMU}, further introduce expert-level multimodal reasoning problems across multiple disciplines, while robustness-oriented evaluations~\cite{PGD} investigate model behavior under adversarial or corrupted inputs.

Nevertheless, existing LVLM benchmarks are still not sufficient for fine-grained tasks, since they rarely probe subordinate-category recognition or attribute-level discrimination. Recent fine-grained-related evaluations have begun to examine LVLMs on fine-grained classification tasks~\cite{geigle2024african,zhang2024visually,tan2025vision}, but they are limited in task coverage, question diversity, or diagnostic depth. In contrast, our \texttt{FG-BMK} jointly evaluates dialogue-level semantic recognition and feature-level visual discriminability across diverse fine-grained domains, providing a more comprehensive test bed for analyzing the capability boundaries of LVLMs on fine-grained image tasks.

\subsection{Fine-Grained Image Tasks}

Fine-grained visual tasks~\cite{XSsurveyPAMI,xu2022dual,wei2024mecom,FSCIL-EACA,jing2024fineclip,ECDNet,WLL-LMDM,FGM-SPCL} aim to distinguish subordinate categories that often share similar global appearances but differ in subtle local attributes or discriminative parts. Such tasks are pivotal in applications including biodiversity monitoring~\cite{jinganimal}, object retrieval~\cite{shen2022semicon}, product recommendation~\cite{wei2022rpc}, and specialized domains such as remote sensing and medical image analysis, where category distinctions often require domain-specific knowledge. Despite the strong general-purpose performance of LVLMs such as GPT-5.4, InternVL, and Qwen, their fine-grained capabilities remain insufficiently understood. Motivated by this issue, we develop a comprehensive benchmark and perform extensive experiments to assess LVLMs on fine-grained tasks. Our analysis reveals their key limitations and provides practical implications for improving future model design and training.

\section{The Evaluation Benchmark}\label{sec:Methods}

\begin{figure*}[t!]
    \centering
    {\includegraphics[width=0.99\textwidth]{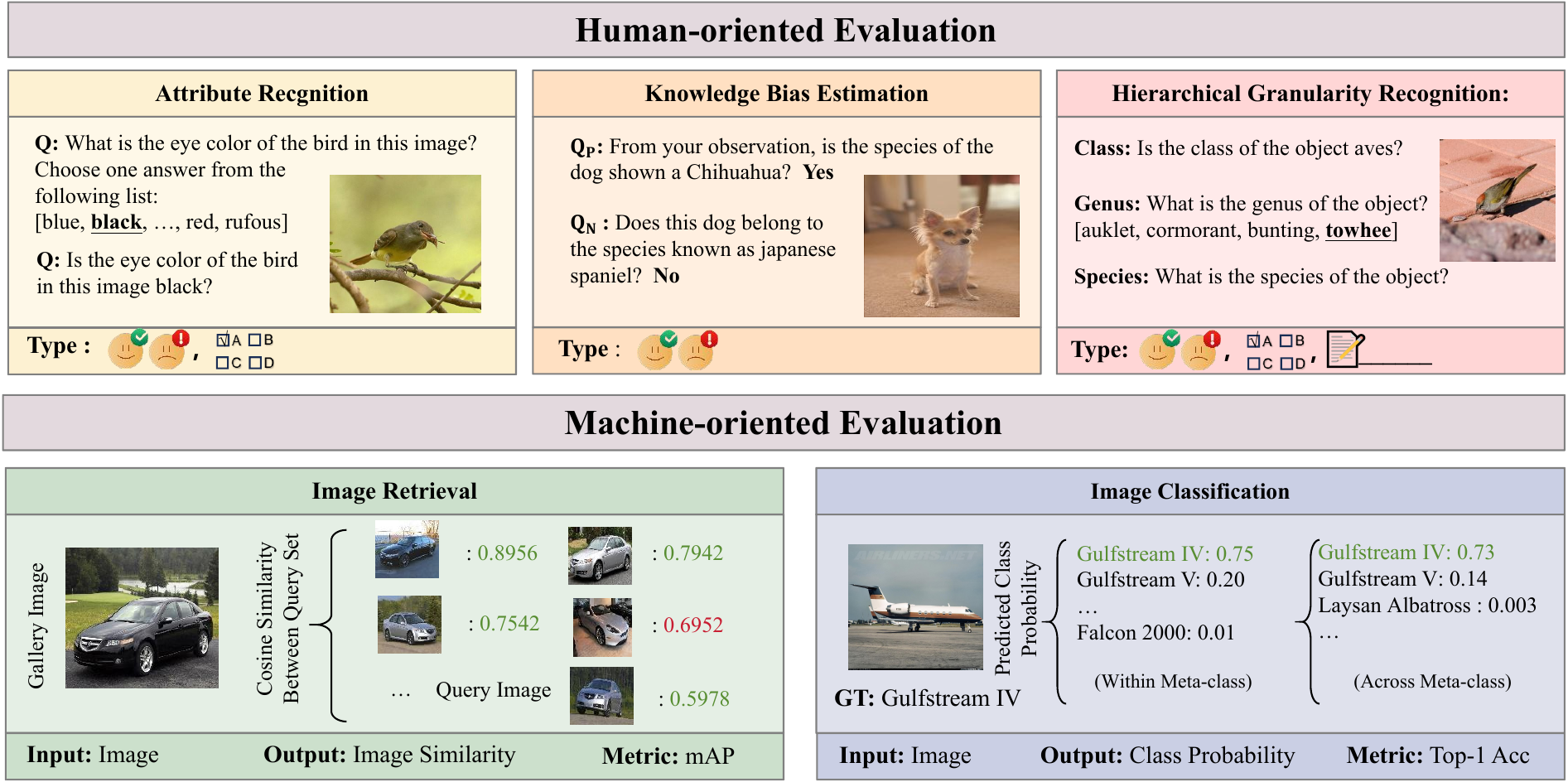}}
    \caption{Our proposed benchmark: The human-oriented evaluation tests the model’s ability to handle fine-grained visual queries (true/false, multiple-choice, short-answer), while the machine-oriented evaluation directly assesses visual feature representation through image retrieval and classification tasks. \includegraphics[height=1em]{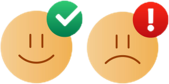}=true/false question, \includegraphics[height=1em]{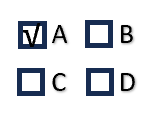}=multiple-choice question, \includegraphics[height=1em]{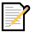}=short-answer question.}
    \label{fig:benchmark}
    \vspace{-1.3em}
\end{figure*}

In this section, we first provide an overview of the benchmark, including its data scale, domain coverage, and two complementary evaluation paradigms. We then describe the evaluation paradigms, tasks and metrics under the human-oriented and machine-oriented settings. Finally, we detail the data collection, question construction, and quality verification procedures used to ensure reliable fine-grained evaluation.

\subsection{FG-BMK Overview}

To systematically evaluate LVLMs on fine-grained image tasks, we construct a comprehensive benchmark termed \texttt{FG-BMK}, containing 1.01 million questions and 0.28 million images collected from 13 fine-grained datasets, covering diverse scenarios from common object-centric domains to specialized domains. Unlike existing benchmarks that mainly focus on classification-style tasks, \texttt{FG-BMK} consists of two complementary evaluation paradigms: human-oriented evaluation measures fine-grained semantic understanding through visual question answering, while machine-oriented evaluation probes visual feature discriminability through image retrieval and classification tasks. As illustrated in Figure~\ref{fig:benchmark}, each evaluation paradigm contains multiple fine-grained tasks with different question formats and evaluation perspectives, enabling \texttt{FG-BMK} to support diagnostic analysis of LVLM limitations across different tasks, granularities, and domains.

\subsection{Evaluation Paradigms, Tasks, and Metrics}
\label{subsec:task_and_metrics}

\paragraph{Evaluation Paradigms} Rather than treating fine-grained capability as a single classification problem, \texttt{FG-BMK} evaluates it through two complementary paradigms: human-oriented evaluation for dialogue-level semantic grounding and machine-oriented evaluation for feature-level visual discriminability. The former reflects the practical interaction form of LVLMs, where answers are jointly influenced by visual perception, language priors, domain knowledge, and prompts; the latter removes the language-generation interface and directly examines whether visual representations can distinguish fine-grained categories. Comparing these two paradigms allows us to diagnose whether LVLM failures mainly arise from weak visual discriminability, insufficient visual-to-semantic grounding, or limited fine-grained knowledge.

\paragraph{Evaluation Tasks} Within each paradigm, we further design tasks to probe different aspects of fine-grained capability. For example, in the human-oriented evaluation, we go beyond category recognition by introducing attribute recognition for subtle local cues critical to subordinate-category discrimination. In the machine-oriented evaluation, we adopt two fundamental vision tasks---image retrieval and classification---and further evaluate classification under both within- and across-meta-category settings to test whether visual representations remain discriminative across single-domain and mixed-domain scenarios.

\paragraph{Evaluation Metrics} For human-oriented tasks, we use three question formats with different answer-space constraints: true/false, multiple-choice, and short-answer. True/false questions are framed as semantic verification, where the model must judge whether a given fine-grained statement is correct. Multiple-choice questions provide a constrained candidate set, allowing us to test whether the model can discriminate among plausible fine-grained options through relative comparison. Short-answer questions remove explicit answer candidates, thereby evaluating fine-grained recognition in a more open-ended setting. For all questions, the response is considered correct if it matches the expected option or contains the ground-truth answer. For machine-oriented tasks, following DINOv2~\cite{DINOv2}, we use mean Average Precision (mAP) for image retrieval and Top-1 accuracy for image classification. The detailed tasks are summarized below:

\textbf{\emph{Human-oriented Evaluation}}: 

\begin{itemize}[leftmargin=*, itemsep=0pt, topsep=0pt]

    \item \textit{Attribute Recognition}: This task consists of true/false and multiple-choice questions that assess whether the model can recognize fine-grained visual attributes, such as size, color, length, shape, and pattern. These attributes often serve as key discriminative cues for distinguishing subordinate categories.

    \item \textit{Knowledge Bias Estimation}: This section uses category-level true/false questions to examine whether LVLMs exhibit uneven recognition ability across different fine-grained categories. By measuring category-wise accuracy, it reveals whether models recognize certain fine-grained concepts more reliably than others.

    \item \textit{Hierarchical Granularity Recognition}: This section consists of true/false, multiple-choice, and short-answer questions that assess whether LVLMs can leverage domain-specific knowledge to recognize object categories at different levels of hierarchical taxonomies. It examines whether models remain reliable as the category granularity increases from coarse to fine levels.
\end{itemize}

\textbf{\emph{Machine-oriented Evaluation}}: 

\begin{itemize}[leftmargin=*, itemsep=0pt, topsep=0pt]
    \item \textit{Image Retrieval}: This task retrieves images from multiple subordinate categories within the same meta-category according to visual feature similarity. It evaluates whether the learned visual representations preserve fine-grained similarity structures.

    \item \textit{Image Classification}: This task recognizes images into fine-grained categories, either within a single meta-category (\emph{e.g.}, species of animals, models of cars) or across multiple meta-categories. It assesses whether visual features are sufficiently discriminative under both category-specific and mixed-domain classification settings.

\end{itemize}

More details about the evaluation tasks are presented in Appendix~\ref{app:eval_details}.

\subsection{Data Curation} \label{subsec:data_curation}

\paragraph{Data Collection.} To ensure both data quality and domain coverage, we source images for \texttt{FG-BMK} from 13 well-established fine-grained datasets. These datasets cover common object-centric domains, such as birds, dogs, cars, and aircraft, as well as specialized domains that require domain-specific visual knowledge, including remote sensing images from MTARSI, enabling us to compare LVLM performance across both common and less frequently studied fine-grained domains. Compared with web-crawled images~\cite{CLIP}, curated fine-grained datasets provide more reliable category boundaries, hierarchical taxonomies, and annotation quality, which are critical for constructing controlled fine-grained evaluation tasks. The statistics and meta-class information of these datasets are summarized in Table~\ref{table:appendix_fg-datasets}.

\paragraph{Question Construction.} For the human-oriented evaluation, we construct questions from the original annotations using task-specific rule-based templates. Depending on the task, the source annotations include attribute labels, category labels, and hierarchical taxonomy information. The construction follows two principles. First, the questions should explicitly target fine-grained visual understanding rather than coarse object recognition. Second, negative labels and distractor options should be visually or semantically close to the ground truth whenever possible, so that the questions require fine-grained discrimination rather than trivial rejection. Specifically, we select negative samples from the same attribute space, taxonomy level, or parent/meta-category according to the task type. For multiple-choice questions, the correct answer and distractor options are randomly ordered to reduce positional bias. To facilitate automatic evaluation, we further append task-specific answer-format instructions to the questions, such as ``Answer with yes or no.'' for true/false questions.

\begin{itemize}[leftmargin=*, itemsep=0pt, topsep=0pt]
    \item \textit{Attribute Recognition}: We design true/false and multiple-choice questions based on fine-grained attribute annotations. For multiple-choice questions, the options include all possible attribute candidates; for true/false questions, we construct balanced positive and negative pairs by matching images with correct or incorrect attribute labels.

    \item \textit{Knowledge Bias Estimation}: We construct category-level true/false questions for each fine-grained category. Positive samples are generated by pairing each image with its ground-truth fine-grained label, while negative samples are generated by pairing the image with a label sampled from other subcategories within the same super-category, ensuring that negative labels remain semantically close to the ground truth. Each image is paired with a positive and a negtive question.

    \item \textit{Hierarchical Granularity Recognition}: We construct true/false, multiple-choice, and short-answer questions across different granularity levels using the hierarchical taxonomy labels associated with each image. For true/false questions, we generate negative samples by matching an image with an incorrect label from the same hierarchical level (\emph{e.g.}, pairing an image of Aves (birds) with Insecta (insects)).  For multiple-choice questions, options are drawn from different categories within the same parent category of the hierarchical taxonomy (\emph{e.g.}, species-level options such as \emph{Black-footed Albatross} and \emph{Laysan Albatross} within the genus \emph{Albatross}). For short-answer questions, the model is asked to directly produce the category label.

    \item \textit{Image Retrieval and Classification}: For the machine-oriented evaluation, we directly use the original fine-grained category labels from each dataset. In image retrieval, images from the same subordinate category are treated as relevant matches. In image classification, we evaluate both within-meta-category and across-meta-category settings. For the across-meta-category setting, we combine fine-grained categories from different datasets into a unified training/testing set, and then evaluating the trained classifier on each individual dataset.
\end{itemize}

\paragraph{Question Quality Verification.} Since automatically generated questions may be sensitive to template wording, we further examine whether the linguistic diversity of question templates affects the evaluation results. Specifically, we expand the original template set to 10 diverse human-written prompts and reconstruct the corresponding questions in the human-oriented benchmark. We then evaluate InternVL3 on the \emph{CUB-200-2011} dataset under both the original and extended template settings. As shown in Table~\ref{table:attributes_InternVL3_appendix} and Table~\ref{table:tf_mc_InternVL3_appendix}, the extended templates lead to only minor accuracy changes across attribute recognition and hierarchical granularity recognition, while the overall model behavior and observed trends remain consistent. This suggests that the evaluation results are not dominated by template-specific artifacts, as long as the questions clearly specify the intended visual concept and answer format.

\section{Observations and Discussions}

This section presents the main observations and discussions based on \texttt{FG-BMK}. We first introduce the evaluated models, and then analyze fine-grained LVLM behavior along a progressive diagnostic path: assessing their fine-grained recognition gaps, diagnosing the bottlenecks behind these failures, examining training design factors for improving fine-grained capabilities, and evaluating robustness under visual and linguistic perturbations.

\subsection{Models under Evaluation}

\begin{table*}[t!]
    \centering
    \vspace{-1em}
    \setlength{\tabcolsep}{10pt}  
    \caption{Training Strategies of the Open-Source Evaluated Models. ``DINOv2'' Is a Purely Visual Model. ``Con'' Denotes Contrastive Loss, ``Gen'' Generative Loss, ``Mat'' Image-Text Matching Loss, ``Rec'' Reconstruction Loss Used in BEiT3, and ``Dis'' Distillation Loss Used in DINOv2.}
    \small 
    \label{table:model_detail}
    \begin{tabular}{|c|c|c|c|c|c|c|c|c|c|}
        \hline
        \multirow{2}{*}{Model}       & \multirow{2}{*}{Vision Size} & \multicolumn{5}{c|}{\rule{0pt}{8pt}Loss Function}                                   & \multicolumn{3}{c|}{Training Data}       \\
        \cline{3-10}
                               & & \rule{0pt}{8pt}Con & Gen & Mat & Rec & Dis & $<$ 0.1B & 0.1B $\sim$ 1B & $>$ 1B \\ \hline
        \hline
        \rule{0pt}{8pt}InternVL3-7B~\cite{zhu2025internvl3}     & ViT-L           & \checkmark           & \checkmark          & \checkmark        &                & \checkmark             &                &         & \checkmark            \\
        InternVL-Chat~\cite{chen2024internvl}     & ViT-6B           & \checkmark           & \checkmark          & \checkmark        &                &              &                &         & \checkmark            \\
        LLaVA-1.5-7B~\cite{LLaVA15}           & ViT-L           &             & \checkmark          &          &                &              & \checkmark              &         &              \\
        Qwen2.5-VL-7B~\cite{Qwen2.5-VL}                & ViT-600M       &  \checkmark           & \checkmark          &  \checkmark        &                &   \checkmark           &                &         & \checkmark            \\
        Qwen-VL-Chat~\cite{Qwen-VL}                & ViT-G       &             & \checkmark          &          &                &              &                &         & \checkmark            \\
        BLIP-2-XL~\cite{BLIP2}      & ViT-G       & \checkmark            & \checkmark           &  \checkmark        &                &              &                & \checkmark       &              \\
        EVA-CLIP~\cite{EVA-CLIP}           & ViT-L          & \checkmark           &            &          &                &              &                &         & \checkmark            \\
        BEiT3~\cite{BEiT3}            & ViT-L           &             &            &          & \checkmark              &              & \checkmark              &         &              \\
        CoCa~\cite{CoCa}                 & ViT-L           & \checkmark           & \checkmark          &          &                &              &                &         & \checkmark            \\
        DINOv2~\cite{DINOv2}               & ViT-L           & \checkmark           &            &          &                & \checkmark            &                & \checkmark       &              \\
        BAGEL~\cite{bagel}               & ViT-L           & \checkmark           & \checkmark           &          &   \checkmark             &             &                &        &  \checkmark            \\
        BLIP3o~\cite{blip3o}               & ViT-L           & \checkmark           & \checkmark           &          &   \checkmark             &             &                &        &    \checkmark          \\
        UniWorld-V1~\cite{UniWorld-V1}               & ViT-L           & \checkmark           & \checkmark           &          &  \checkmark              &             &                &        &  \checkmark            \\
        \hline
    \end{tabular}
    \vspace{-7pt}
\end{table*}

Given the diverse landscape of existing LVLMs and vision-language foundation models, we select a representative set of models covering different model families, access types, architecture designs, training objectives, visual encoder scales, and training data scales, as summarized in Table~\ref{table:model_detail}. Our evaluation includes widely used open-source LVLMs, closed-source models such as GPT-5.4~\cite{GPT4} and Gemini-3.5-flash~\cite{Gemini}, unified understanding-generation models, and a purely visual foundation model. This selection allows us to analyze both dialogue-level fine-grained semantic recognition and feature-level visual discriminability.

For human-oriented evaluation, we evaluate instruction-tuned LVLMs and closed-source models through dialogue-style questions. For machine-oriented evaluation, we focus on models with accessible visual features, since image retrieval and classification require extracting visual representations. The purely visual model provides a feature-level reference, while unified multimodal models are included to cover the emerging paradigm that integrates visual understanding and generation. To better isolate the effects of model architecture and training strategy, we use representative versions from each model family in machine-oriented evaluation, where their visual encoders and training objectives are more transparent. Further details about the evaluated models can be found in Appendix~\ref{app:evaluated_models}.

\subsection{LVLMs Remain Inadequate Fine-Grained Recognizers}

\begin{figure}[t]
    \centering
    \includegraphics[width=0.8\columnwidth]{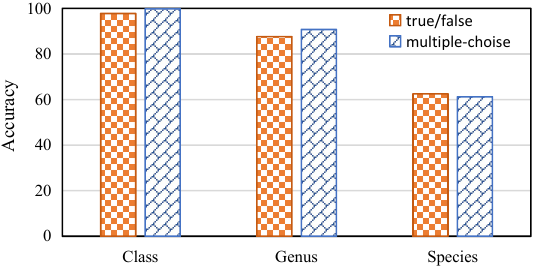}
    \caption{Results of InternVL3~\cite{zhu2025internvl3} on true/false and multiple-choice questions across different levels of granularity on the \emph{CUB-200-2011}~\cite{CUB} dataset. The $x$-axis denotes the granularity of the recognition questions.}
    \label{fig:sec42_granularity}
    \vspace{-1em}
\end{figure}

\begin{figure}[t]
    \centering
    \includegraphics[width=0.8\columnwidth, height=10em]{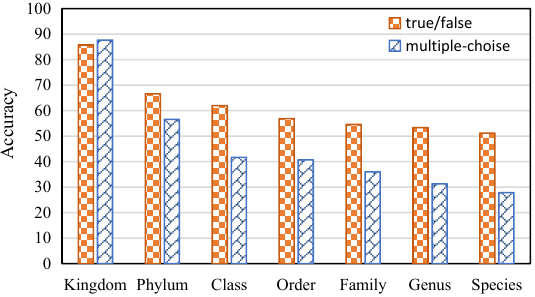}
    \caption{Results of LLaVA~\cite{LLaVA15} on true/false and multiple-choice questions across different levels of granularity on the \emph{iNat2021}~\cite{CUB} dataset. The $x$-axis denotes the granularity of the recognition questions.}
    \label{fig:sec42_granularity_iNat}
    \vspace{-0.1em}
\end{figure}

After introducing the evaluated models, we first ask a direct question: to what extent can current LVLMs recognize fine-grained visual categories? A single aggregate accuracy is insufficient to characterize this ability, since fine-grained recognition involves multiple levels of difficulty. We first examine how model performance changes as category labels move from coarse to increasingly fine levels, revealing whether LVLMs can preserve recognition ability under finer semantic distinctions. We then compare LVLMs with fine-grained tailored models, using specialized recognizers as a reference to assess the gap between general-purpose LVLMs and models explicitly designed for fine-grained recognition. Finally, since fine-grained category decisions often depend on subtle combinations of visual attributes, we further evaluate attribute-level recognition as intermediate evidence for category-level understanding. These analyses are supported by the granularity results in Figure~\ref{fig:sec42_granularity} and Figure~\ref{fig:sec42_granularity_iNat}, the tailored-model comparison in Table~\ref{table:fg-tailored}, and the attribute-recognition results in Table~\ref{table:attributes_InternVL3}. Together, they reveal a consistent recognition gap: current LVLMs remain inadequate fine-grained recognizers.

\begin{findingbox}
    \textit{Finding 1:} LVLMs struggle to distinguish excessively fine-grained categories.
\end{findingbox}

To examine how recognition performance changes with category granularity, we evaluate questions at multiple taxonomy levels, ranging from coarse taxonomic levels such as kingdom or class to fine-grained species. As shown in Figure~\ref{fig:sec42_granularity} and Figure~\ref{fig:sec42_granularity_iNat}, we take InternVL3~\cite{zhu2025internvl3} as a representative example and observe a consistent decline in its true/false and multiple-choice accuracy as the category granularity becomes finer. At the class level (\emph{e.g.}, \textit{``Is the class of the object in this image an Insecta/Aves?''}), the model achieves 99.76\% accuracy on multiple-choice questions and 99.77\% on true/false questions.\footnote{When questions are relatively simple, LVLMs achieve very high accuracy. The slight difference between multiple-choice and true/false accuracy may be caused by answer-space differences and randomness.} However, when the granularity narrows to the genus level, where competing labels are selected from different genera within the same class (\emph{e.g.}, \textit{``Is the object in this image an albatross or a gull?''}), its multiple-choice accuracy decreases to 90.75\%, corresponding to a 9.01\% drop. When moving further to the species level, where negative labels are drawn from different species within the same genus (\emph{e.g.}, \textit{``Is the object in this image a black-footed albatross/Laysan albatross?''}), the accuracy further decreases to 62.48\% on true/false questions and 61.18\% on multiple-choice questions. This indicates that LVLMs can handle coarse semantic distinctions reasonably well, but become much less reliable when distinguishing closely related subordinate categories. Similar degradation is observed across other LVLMs. Additional examples of multiple-choice and true/false questions can be found in Appendix~\ref{app:res_hierar}.

\begin{findingbox}
    \textit{Finding 2:} LVLMs do not outperform fine-grained tailored models in fine-grained tasks.
\end{findingbox}

To further contextualize the fine-grained recognition ability of LVLMs, we compare them with models specifically designed for fine-grained recognition. As shown in Table~\ref{table:fg-tailored}, although LVLMs achieve competitive results on several datasets, their performance remains below that of fine-grained tailored models under both short-answer evaluation and linear probing. For example, on FGVC Aircraft, LVLMs achieve 66.19\% accuracy with short-answer questions and 78.88\% with linear probing, whereas the fine-grained tailored model reaches 95.40\%. Similar gaps can also be observed on Stanford Dogs and Stanford Cars.

\begin{table}[t]
    \vspace{0.2em}
    \caption{Comparison of LVLMs and Fine-Grained Tailored Models on Classification Tasks. ``SA'' Denotes LVLMs Fine-Tuned on Fine-Grained Datasets for Short-Answer Questions, ``LC'' Represents Linear Classifiers Using LVLM Visual Features, and ``FG-Tailored'' Refers to State-of-the-Art Fine-Grained Tailored Models.}
    \label{table:fg-tailored}
    \centering
    \small
    \setlength{\tabcolsep}{9pt}
    \begin{tabular}{|c|c|c|c|}
        \hline
        \rule{0pt}{8pt}Datasets & SA & LC & FG-Tailored \\
        \hline
        \hline
        \emph{\rule{0pt}{8pt}CUB-200-2011} & 85.60 & 91.65 & 93.10~\cite{diao2022metaformer} \\
        \emph{Stanford Dogs} & 86.49 & 90.50 & 97.30~\cite{bera2022sr} \\
        \emph{Stanford Cars} & 90.55 & 94.30 & 97.10~\cite{liu2024progressive} \\
        \emph{Food-101} & 95.25 & 95.67 & 98.60~\cite{behera2021context} \\
        \emph{FGVC Aircraft} & 66.19 & 78.88 & 95.40~\cite{sikdar2024interweaving} \\
        \hline
    \end{tabular}
    \vspace{-1em}
\end{table}

This gap may be partly attributed to the different optimization goals of the two types of models. Fine-grained tailored models are usually designed for specific recognition domains and often introduce mechanisms to capture local, part-level, or hierarchical visual details. For example, CAP~\cite{behera2021context} employs context-aware attentional pooling to aggregate hierarchical contextual information from pixels to regions and images, which benefits fine-grained classification. In contrast, LVLMs are primarily optimized for general multimodal understanding and instruction following, and their standard architecture (\emph{e.g.}, ViT + MLP + LLM) does not explicitly emphasize such fine-grained discriminative cues. Although these specialized components cannot be directly transferred to LVLMs, their core idea of strengthening local and hierarchical visual evidence remains relevant for improving fine-grained recognition while preserving general-purpose multimodal capabilities.

\begin{table*}[!t]
    \caption{Attribute Recognition Accuracy of InternVL3~\cite{zhu2025internvl3} on the \emph{CUB-200-2011}~\cite{CUB} Dataset (Values in Parentheses Represent the Average Accuracy for Each Attribute).}
    \setlength{\tabcolsep}{9pt}
    \label{table:attributes_InternVL3}
    \centering
    \small
    \begin{tabular}{|c|c|c|c|c|c|c|c|} 
        \hline
        \multicolumn{8}{|c|}{\textbf{\rule{0pt}{8pt}Color Attribute (47.40)}} \\
        \hline 
        \hline 
        \rule{0pt}{8pt}belly color        & 58.49 & back color        & 34.98 & bill color        & 51.31 & breast color      & 54.25 \\ 
        crown color        & 55.30 & eye color         & 84.59 & forehead color    & 53.32 & leg color         & 44.01 \\ 
        nape color         & 39.24 & throat color      & 52.77 & under tail color  & 34.69 & underparts color  & 56.20 \\ 
        upper tail color   & 37.30 & upperparts color  & 28.75 & wing color        & 30.16 & primary color     & 43.05 \\ 
        \hline
        \hline
        \multicolumn{8}{|c|}{\textbf{\rule{0pt}{8pt}Pattern Attribute (50.13)}} \\
        \hline 
        \hline 
        \rule{0pt}{8pt}back pattern       & 40.94 & belly pattern     & 68.13 & breast pattern    & 65.12 & head pattern      & 35.92 \\ 
        tail pattern       & 41.64 & wing pattern      & 49.04 &                   &       &                   &       \\ 
        \hline
        \hline
        \multicolumn{8}{|c|}{\textbf{\rule{0pt}{8pt}Shape Attribute (30.95)}} \\
        \hline 
        \hline 
        \rule{0pt}{8pt}bill shape         & 37.61 & shape             & 52.37 & tail shape        & 10.42 & wing shape        & 23.39 \\ 
        \hline
        \hline
        \multicolumn{4}{|c|}{\textbf{\rule{0pt}{8pt}Length Attribute (71.03)}} & \multicolumn{4}{c|}{\textbf{Size Attribute (52.55)}} \\
        \hline 
        \hline 
        \multicolumn{2}{|c|}{\rule{0pt}{8pt}bill length}   & \multicolumn{2}{c|}{71.03}  & \multicolumn{2}{c|}{size} & \multicolumn{2}{c|}{52.55}    \\ 
        \hline
    \end{tabular}
\end{table*}

\begin{findingbox}
    \textit{Finding 3:} LVLMs exhibit significant room for improvement in recognizing fine-grained attributes.
\end{findingbox}

Since fine-grained category recognition often relies on local visual evidence, we further examine whether LVLMs can recognize the attributes that distinguish similar categories. As shown in Table~\ref{table:attributes_InternVL3} and Table~\ref{table:app_attributes_Qwen2.5_VL}, LVLMs exhibit uneven performance across different attribute types. InternVL3~\cite{zhu2025internvl3} and Qwen2.5-VL~\cite{Qwen2.5-VL} achieve 50.13\% and 45.12\% average accuracy for pattern recognition, respectively, but only 30.95\% and 29.30\% for shape recognition. Although a few attributes achieve relatively high accuracy, most attributes remain far from being reliably recognized, and some part-level attributes can be as low as around 10\%. These results indicate that LVLMs still have substantial room for improvement in fine-grained attribute recognition.

Such attribute-level weaknesses can directly limit fine-grained category reasoning, where the correct category often depends on subtle combinations of color, shape, pattern, and part-level cues. We also observe that attribute-wise performance varies across models: for example, InternVL3 struggles more with pattern recognition than with size, whereas Gemini-3.5-flash~\cite{Gemini} shows the opposite trend. Additionally, our comparison across model versions suggests that recent LVLMs have made more substantial progress in recognizing pattern and length, but their gains in color and shape recognition are comparatively limited. Detailed results of the attribute recognition task are provided in Appendix~\ref{app:res_attribute}.

\vspace{1em}

Overall, the above observations show that current LVLMs remain inadequate fine-grained recognizers: their performance degrades under increasingly fine category granularity, they still lag behind fine-grained tailored models, and they exhibit limited and uneven attribute-level recognition. However, these performance gaps alone do not reveal where the failures originate. We therefore next move from measuring the recognition gap to diagnosing its underlying bottlenecks.

\subsection{Bottlenecks Behind LVLM Failures in Fine-Grained Tasks.}

Having established that current LVLMs remain inadequate fine-grained recognizers, we next diagnose where these failures originate. Fine-grained recognition depends not only on whether visual features are discriminative, but also on whether such visual evidence can be aligned with language semantics and grounded into correct fine-grained categories. Therefore, we analyze LVLM failures from the perspectives of visual representation, semantic grounding, visual-to-textual alignment, and category-level long-tail behavior.

We first compare feature-level discriminability with dialogue-based recognition accuracy to determine whether failures arise from insufficient visual representations or from the inability to use these representations in semantic recognition. We then leverage unified models to further examine the relation between visual discriminability and semantic grounding, using their generation capability to inspect whether fine-grained category names are grounded into corresponding visual concepts.  Next, to understand how visual features are connected with language semantics, we focus on the visual-to-textual alignment stage and examine how alignment affects both visual feature separability and fine-grained semantic grounding. Finally, we examine whether recognition failures are concentrated on long-tail fine-grained categories, and further trace these category-level disparities through balanced fine-tuning and training-data coverage analysis.

\begin{figure*}[t!]
    \centering
    {\includegraphics[width=0.94\textwidth,height=8em]{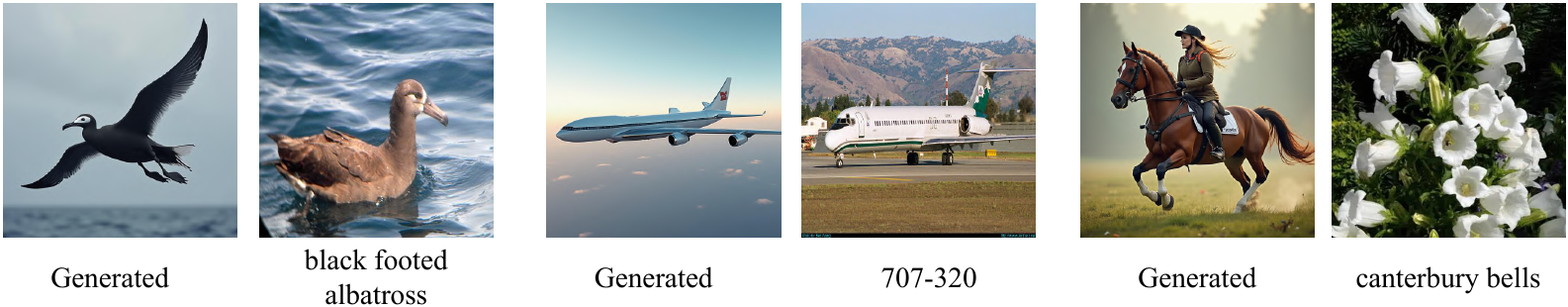}}
    \caption{Qualitative comparison between real images and fine-grained category-conditioned generated images.}
    \label{fig:generated_fg_samples}
    \vspace{-1.2em}
\end{figure*}

\begin{figure*}[t]
    \centering

    \begin{minipage}[b]{0.30\textwidth}
        \centering
        \includegraphics[width=\textwidth, height=10em]{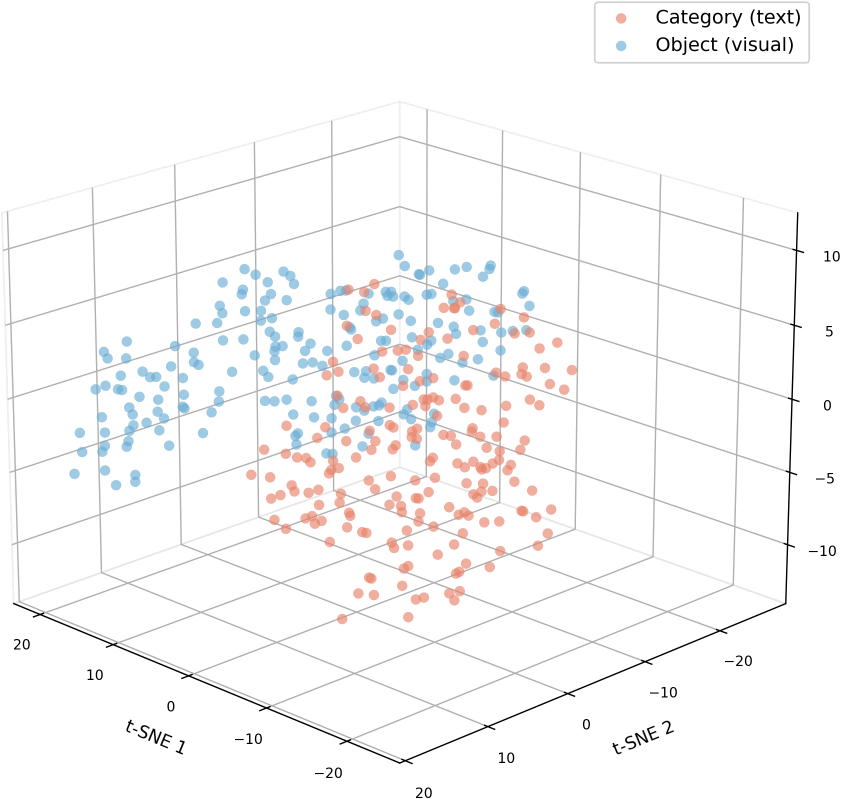}
        \vspace{-0.7em}
        \small (a) Original
    \end{minipage}
    \hspace{0.02\textwidth}
    \begin{minipage}[b]{0.30\textwidth}
        \centering
        \includegraphics[width=\textwidth, height=10em]{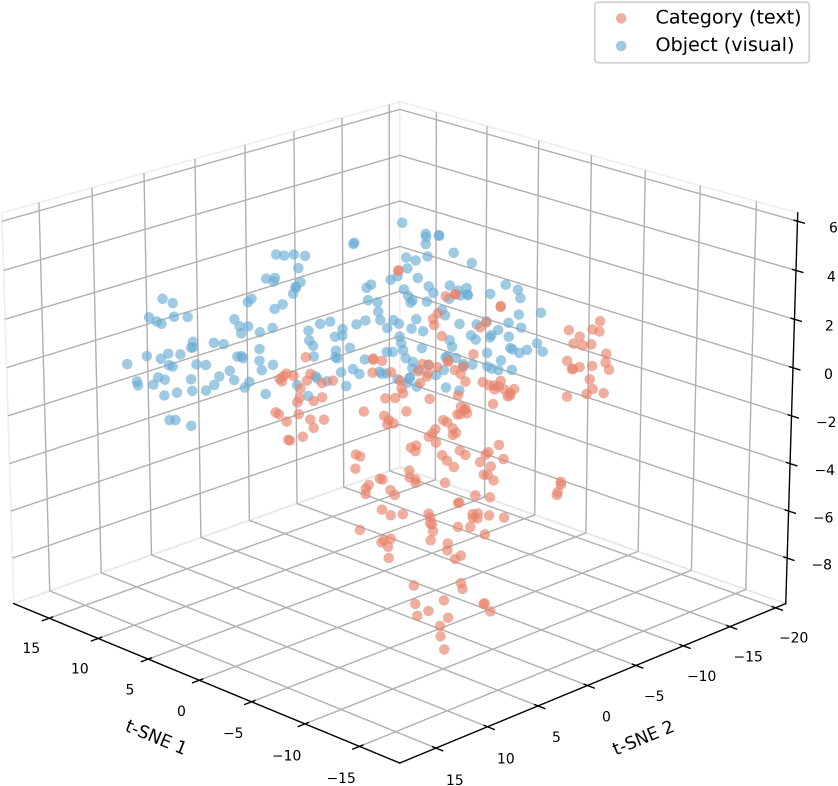}
        \vspace{-0.7em}
        \small (b) Aligned-Recap
    \end{minipage}
    \hspace{0.02\textwidth}
    \begin{minipage}[b]{0.30\textwidth}
        \centering
        \includegraphics[width=\textwidth, height=10em]{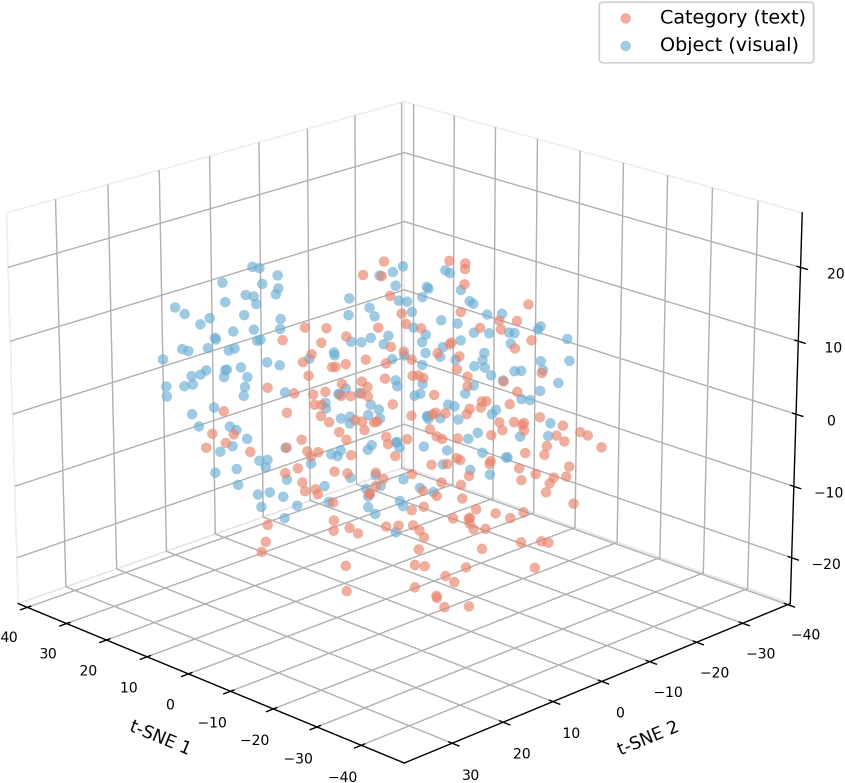}
        \vspace{-0.7em}
        \small (c) Aligned-FG
    \end{minipage}

    \vspace{0.6em}

    \caption{Visualization of visual-text alignment on CUB under different settings.}
    \label{fig:aligned_img_txt_visualization}
    \vspace{-0.4em}
\end{figure*}

\begin{figure}[t]
    \centering
    \includegraphics[width=0.7\columnwidth]{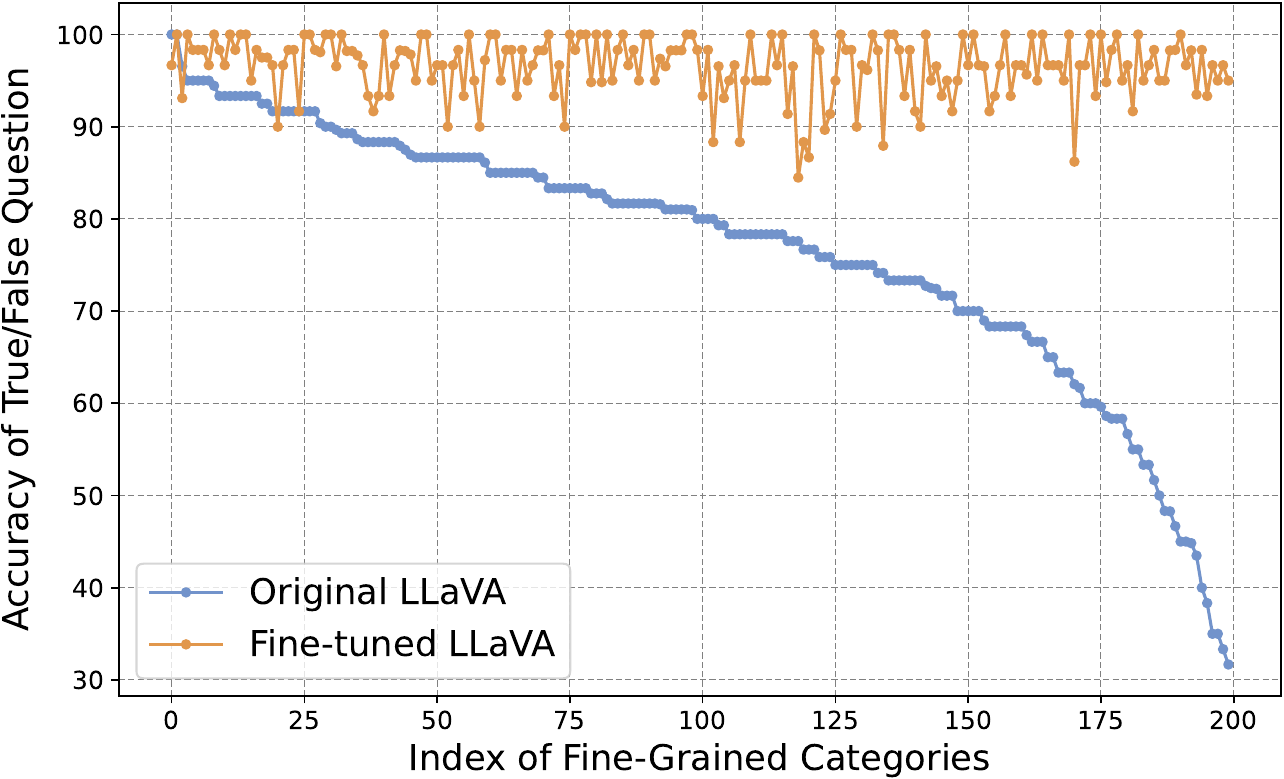}
    \caption{Comparison of the original (blue dots) and fine-tuned (yellow dots) LLaVA models on occurrence-balanced fine-grained bird categories. True/false accuracy per category is ranked.}
    \label{fig:sec42_each_species}
    \vspace{-1em}
\end{figure}

These analyses are supported by the feature-level linear probing and dialogue-based recognition comparison in Table~\ref{table:common_specialized_domain}, the unified-model probing and generation analysis in Table~\ref{table:generated_linear_probe} and Figure~\ref{fig:generated_fg_samples}, the alignment-stage analysis in Table~\ref{table:llava}, Table~\ref{table:llava_sft}, and Figure~\ref{fig:aligned_img_txt_visualization}, and the category-level long-tail analysis in Figure~\ref{fig:sec42_each_species}. Together, they show that LVLM failures in fine-grained recognition are not caused by a single bottleneck, but by the combined effects of visual discriminability limits, weak fine-grained semantic grounding, alignment-induced feature changes, and uneven category-level knowledge coverage.

\begin{findingbox}
    \textit{Finding 4:} Semantic understanding, rather than visual discrimination, becomes the bottleneck of LVLMs in specialized domains.
\end{findingbox}

\begin{table*}[t!]
    \caption{Comparison of LVLM Performance on Fine-Grained Datasets from Common and Specialized Domains. Results Are Reported in the Order of ``Multiple-Choice / True-False / Linear Probe''.}
    \setlength{\tabcolsep}{12pt}
    \centering
    \small
    \label{table:common_specialized_domain}
    \begin{tabular}{|c|c|c|c|c|c|}
        \hline
        \multirow{2}{*}{Models} & \multicolumn{2}{c|}{\rule{0pt}{8pt}Common domains} & \multicolumn{2}{c|}{Specialized domains} \\
        \cline{2-5}
         & \rule{0pt}{8pt}\emph{FGVC Aircraft} & \emph{Stanford Dogs} & \emph{SkinCon} & \emph{MTARSI} \\
        \hline
        \hline
        \rule{0pt}{8pt}LLaVA      & 58.75 / 77.62 / 62.46 & 68.81 / 77.45 / 80.73 & 41.81 / 59.62 / 81.29 & 60.32 / 71.60 / 94.79 \\
        Qwen2.5-VL & 94.84 / 89.56 / 62.07 & 96.74 / 94.50 / 79.07 & 60.81 / 60.10 / 70.89 & 70.11 / 65.08 / 93.20 \\
        Qwen3.0-VL & 92.29 / 81.43 / 55.65 & 96.28 / 92.48 / 77.29 & 66.51 / 67.70 / 72.48 & 71.34 / 62.87 / 96.03 \\ 
        InternVL3.0 & 85.48 / 86.92 / 45.42 & 92.02 / 92.48 / 73.90 & 60.81 / 66.75 / 69.72 & 64.11 / 73.99 / 88.71 \\  
        \hline
    \end{tabular}
\end{table*}

\begin{table*}[t!]
    \caption{Classification Accuracy of Unified Models on Real and Self-Generated Fine-Grained Images. ``Original'' Denotes Results on Original Images, while ``Generated'' Denotes Results on Images Synthesized by the Models Conditioned on Fine-Grained Category Names.}
    \setlength{\tabcolsep}{15pt}
    \centering
    \small
    \label{table:generated_linear_probe}
    \begin{tabular}{|c|c|c|c|c|c|c|}
        \hline
        \multirow{2}{*}{Models} & \multicolumn{2}{c|}{\rule{0pt}{8pt}\emph{CUB-200-2011}} & \multicolumn{2}{c|}{\emph{FGVC Aircraft}} & \multicolumn{2}{c|}{\emph{Flowers102}} \\
        \cline{2-7}
         & \rule{0pt}{8pt}Original & Generated & Original & Generated & Original & Generated \\ \hline
        \hline
        \rule{0pt}{8pt}UniWorld-V1 & 86.02 & 64.89 & 62.79 & 27.00 & 99.33 & 81.07 \\
        Bagel      & 82.43 & 36.59 & 53.70 & 13.33 & 98.95 & 65.82 \\
        BLIP3-o    & 85.79 & 65.29 & 79.53 & 36.39 & 99.62 & 85.49 \\
        \hline
    \end{tabular}
    \vspace{-0.1em}
\end{table*}

To localize the source of fine-grained recognition failures, we compare feature-level linear probing with dialogue-based recognition across common and specialized domains. This comparison allows us to examine whether failures come from insufficient visual feature discriminability or from the inability to map visual evidence to correct semantic concepts. As shown in Table~\ref{table:common_specialized_domain}, LVLMs exhibit different bottlenecks across common and specialized fine-grained domains.

On common datasets such as FGVC Aircraft and Stanford Dogs, visual feature discriminability remains an important limiting factor, consistent with prior findings~\cite{zhang2024visually}. For example, although Qwen2.5-VL achieves 94.84\% multiple-choice accuracy on FGVC Aircraft, its linear-probe accuracy is only 62.07\%, indicating that its visual representations are not sufficiently discriminative for fine-grained classification.

In contrast, we observe a different pattern in specialized domains such as remote sensing (MTARSI) and medical dermatology (SkinCon). Although LVLM visual features remain highly discriminative under linear classification, their dialogue-style recognition accuracy drops markedly. For instance, on MTARSI, LLaVA achieves 94.79\% linear-probe accuracy, but only 60.32\% and 71.60\% accuracy on multiple-choice and true/false questions, respectively. Similarly, Qwen3.0-VL reaches 96.03\% linear-probe accuracy on MTARSI, while its multiple-choice and true/false accuracies are only 71.34\% and 62.87\%.

This suggests that, in specialized domains, the limitation of LVLMs no longer primarily lies in visual discrimination; instead, the model struggles to map already discriminative visual cues to the correct semantic concepts under dialogue-based recognition. We attribute this gap to the scarcity of such domain-specific concepts in pre-training corpora, which prevents the model from forming sufficiently strong semantic priors for these categories. This interpretation is further supported by our appendix experiments, where fine-tuning on specialized-domain data substantially improves performance on multiple-choice and true/false questions.

\begin{findingbox}
    \textit{Finding 5:} Fine-grained visual representations do not imply fine-grained semantic grounding.
\end{findingbox}

The results in specialized domains reveal a clear mismatch between feature-level discriminability and dialogue-based recognition: fine-grained visual representations can be separable, yet the corresponding category semantics may still not be properly grounded. Building on this observation, we further examine the relation between visual representations and semantic grounding. Unified models provide a suitable testbed for this analysis, because their generation capability allows us to inspect whether a fine-grained category name can be translated into the corresponding visual concept. We therefore use linear probing on original fine-grained images to evaluate visual feature discriminability, and apply linear probing to category-conditioned generated images to test whether these models can ground fine-grained category names into corresponding visual concepts.

As shown in Table~\ref{table:generated_linear_probe}, unified models exhibit strong discriminability on original fine-grained images. However, when the images are replaced with the models' self-generated images conditioned on fine-grained category names, the linear-probe accuracy drops substantially. For example, BLIP3-o decreases from 89.92\% to 73.65\% on CUB-200-2011 and from 79.05\% to 55.87\% on FGVC Aircraft.

This gap is also evident from the generated images. As shown in Figure~\ref{fig:generated_fg_samples}, images synthesized from fine-grained category names often fail to reflect the defining characteristics of the target categories, and sometimes even contain incorrect visual content. These results indicate that unified models can distinguish fine-grained categories in original images, but may still fail to ground fine-grained category names into the corresponding visual semantics.

\begin{findingbox}
    \textit{Finding 6:} The alignment strategy in LVLMs might impair the fine-grained discriminability of visual features.
\end{findingbox}

\begin{table*}[t!]
    \caption{Performance of Different LLaVA Variants after Alignment Retraining and SFT on General and Fine-Grained Tasks. Improvements Are Reported Relative to the Original LLaVA.}
    \setlength{\tabcolsep}{11pt}
    \centering
    \small
    \label{table:llava_sft}
    \begin{tabular}{|l|c|c|c|c|c|c|c|c|}
        \hline
        \multirow{2}{*}{Models} & \multicolumn{3}{c|}{\rule{0pt}{8pt}\emph{POPE}} & \multirow{2}{*}{\emph{TextVQA}} & \multirow{2}{*}{\emph{GQA}} & \multirow{2}{*}{\emph{CUB}} & \multirow{2}{*}{\emph{Stanford Cars}} & \multirow{2}{*}{\emph{Stanford Dogs}} \\
        \cline{2-4}
         & \rule{0pt}{8pt}Rand & Pop & Adv &  &  &  &  &  \\
        \hline
        \hline
        \rule{0pt}{8pt}Original
            & 87.30 & 86.10 & 84.20 & 58.20 & 62.00 & 85.60 & 90.55 & 86.49 \\
        Aligned-ReCap
            & 88.86 & 87.46 & 86.26 & 58.70 & 62.20 & 85.80 & 90.83 & 86.60 \\
        \emph{$\Delta$ vs. Original}
            & \emph{+1.56} & \emph{+1.36} & \emph{+2.06} & \emph{+0.50} & \emph{+0.20} & \emph{+0.20} & \emph{+0.28} & \emph{+0.11} \\
        Aligned-FG
            & 87.40 & 86.30 & 84.70 & 58.30 & 62.10 & 86.32 & 91.73 & 87.58 \\
        \emph{$\Delta$ vs. Original}
            & \emph{+0.10} & \emph{+0.20} & \emph{+0.50} & \emph{+0.10} & \emph{+0.10} & \emph{+0.72} & \emph{+1.18} & \emph{+1.09} \\
        \hline
    \end{tabular}
    \vspace{-0.1em}
\end{table*}

After examining visual discriminability and semantic grounding, we next focus on visual-to-textual alignment, the stage where LVLMs connect visual features with language semantics. To investigate the effect of this alignment stage on fine-grained visual representations, we compare the linear-probe accuracy of LLaVA's~\cite{LLaVA15} original visual features with that of features after visual-to-textual alignment on fine-grained classification tasks. As shown in the first two columns of Table~\ref{table:llava}, the original features demonstrate superior classification performance, outperforming the aligned ones by an average of 3.39\%. This suggests that the standard alignment process may weaken the fine-grained discriminability of visual features.

This decline can be attributed to two key factors. First, aligning visual and textual features may introduce distortions due to inconsistencies between their respective feature spaces. Second, granularity inconsistencies in LVLMs' alignment data---where fine-grained objects in images are paired with coarse-grained textual descriptions, as demonstrated in our qualitative analysis in Appendix~\ref{app:quali_analy_in_granularity}---may negatively affect the discriminability of the aligned visual features.

To examine the impact of alignment-data granularity, we retrain the alignment module in LLaVA on two new alignment datasets: one with fine-grained category-level text matching the granularity of the objects in the images, and the other with recapped long captions that provide richer image descriptions. As shown in Table~\ref{table:llava}, both fine-grained category-level supervision and richer caption supervision improve the quality of aligned visual features: fine-grained category-level text significantly boosts classification accuracy, with gains of 2.55\% on \emph{Stanford Dogs} and 1.73\% on \emph{Stanford Cars}, while recapped long captions also bring marginal improvements.

We then compare the performance of different LLaVA variants after SFT on general and fine-grained tasks. As shown in Table~\ref{table:llava_sft}, LLaVA aligned with long captions consistently outperforms the original LLaVA, especially on general tasks(+1.66 on POPE), whereas LLaVA aligned with fine-grained content shows clearer gains on fine-grained tasks (+0.72 on CUB, +1.18 on Stanford Cars). This suggests that effective alignment data should be task-aware: detailed captions help improve general multimodal understanding, while fine-grained category-level supervision strengthens fine-grained capabilities.

To further understand how alignment benefits LVLM performance, we visualize the aligned fine-grained visual features and category text embeddings in the same representation space. As shown in Figure~\ref{fig:aligned_img_txt_visualization}, fine-grained category-level alignment brings visual features closer to their corresponding category embeddings, making it easier for the LVLM to associate visual evidence with the correct category semantics during fine-grained recognition. Further analysis is detailed in Appendix~\ref{app:granular_retrain}.

\begin{table}[t]
    \vspace{0.2em}
    \caption{Accuracy of LLaVA Visual Features Before and After Alignment. ``Origin'' Denotes Original Features from the Vision Encoder. ``Aligned'' Denotes Features Aligned to Text with Inconsistent Granularity, ``ReCap'' Denotes Features Aligned with Long Captions, while ``FG'' Denotes Those Aligned to Fine-Grained Text.}
    \label{table:llava}
    \setlength{\tabcolsep}{8pt}
    \centering
    \small
    \begin{tabular}{|c|c|c|c|c|}
        \hline
        \rule{0pt}{8pt}Datasets & Origin & Aligned & ReCap & FG \\
        \hline
        \hline
        \emph{\rule{0pt}{8pt}CUB-200-2011} & 79.77 & 73.17 & 73.89 & 75.06 \\
        \emph{Stanford Dogs} & 81.24 & 78.14 & 78.34 & 80.69 \\
        \emph{Stanford Cars} & 87.57 & 83.90 & 84.33 & 85.63 \\
        \emph{Food-101} & 94.27 & 93.35 & 93.65 & 94.32 \\
        \emph{DeepFashion} & 69.94 & 67.30 & 67.35 & 67.75 \\
        \hline
    \end{tabular}
    \vspace{-0.5em}
\end{table}

\begin{findingbox}
    \textit{Finding 7:} The inconsistent recognition accuracy of LVLMs across fine-grained categories can be attributed to the characteristics of their training data and the underlying LLM base.
\end{findingbox}

After analyzing representation- and alignment-level bottlenecks, we further examine whether LVLMs exhibit knowledge bias in recognizing different fine-grained categories. To this end, we rank fine-grained categories according to the model's accuracy on true/false questions. As shown in Figure~\ref{fig:sec42_each_species}, using LLaVA~\cite{LLaVA15} as an example, the model shows highly inconsistent recognition ability across categories, achieving nearly 90\% accuracy for some categories while dropping to approximately 30\% for others. This indicates a clear category-level long-tail pattern in fine-grained recognition.

We consider two possible explanations for this inconsistency: the training data may contain imbalanced fine-grained knowledge, or some fine-grained categories may be intrinsically more difficult for LVLMs to learn. To distinguish between these possibilities, we fine-tune LVLMs using data in which fine-grained categories appear in a balanced manner, and then re-evaluate their recognition performance. As indicated by the yellow dots in Figure~\ref{fig:sec42_each_species}, the fine-tuned LLaVA achieves consistently strong recognition across all fine-grained categories. This result suggests that the observed knowledge bias mainly stems from the uneven representation of fine-grained knowledge in training data, rather than from the inherent difficulty of learning particular categories.

To further trace the source of this imbalance, we examined the occurrence frequency of fine-grained categories in the LVLM training data. Interestingly, we found that these categories are almost absent from the training data. This suggests that the observed category-level inconsistency is not solely caused by the visual model or by category-specific learning difficulty, but is largely inherited from the language-side knowledge priors of the underlying LLM. Additional results for other LVLMs exhibit similar trends and can be found in Appendix~\ref{app:res_knowledge}.

\subsection{Training Designs for Better Fine-Grained LVLM Capabilities.}

After diagnosing the bottlenecks behind fine-grained LVLM failures, we further examine which training design factors can improve fine-grained capabilities. This analysis considers both the visual representation side, where feature separability provides the basis for fine-grained recognition, and the instruction-tuning side, where the model must acquire fine-grained knowledge without forgetting general multimodal capabilities. We therefore analyze LVLMs from the perspectives of training objective, feature quality, encoder and data scale, and SFT data composition.

We first examine how different training paradigms affect fine-grained visual discriminability by evaluating visual features on fine-grained classification and retrieval tasks. To understand where the performance differences come from, we further analyze their global feature distributions and local patch-level correspondences. We then investigate whether raw scale, including vision-encoder size and training-data scale, is sufficient to improve fine-grained visual representations. Finally, we examine whether fine-grained supervision can be incorporated during SFT without sacrificing general capabilities, by comparing direct fine-grained tuning with joint SFT on general and fine-grained data.

\begin{figure*}[t]
    \centering
    \begin{minipage}[b]{0.47\textwidth}
        \centering
        \includegraphics[width=0.7\textwidth]{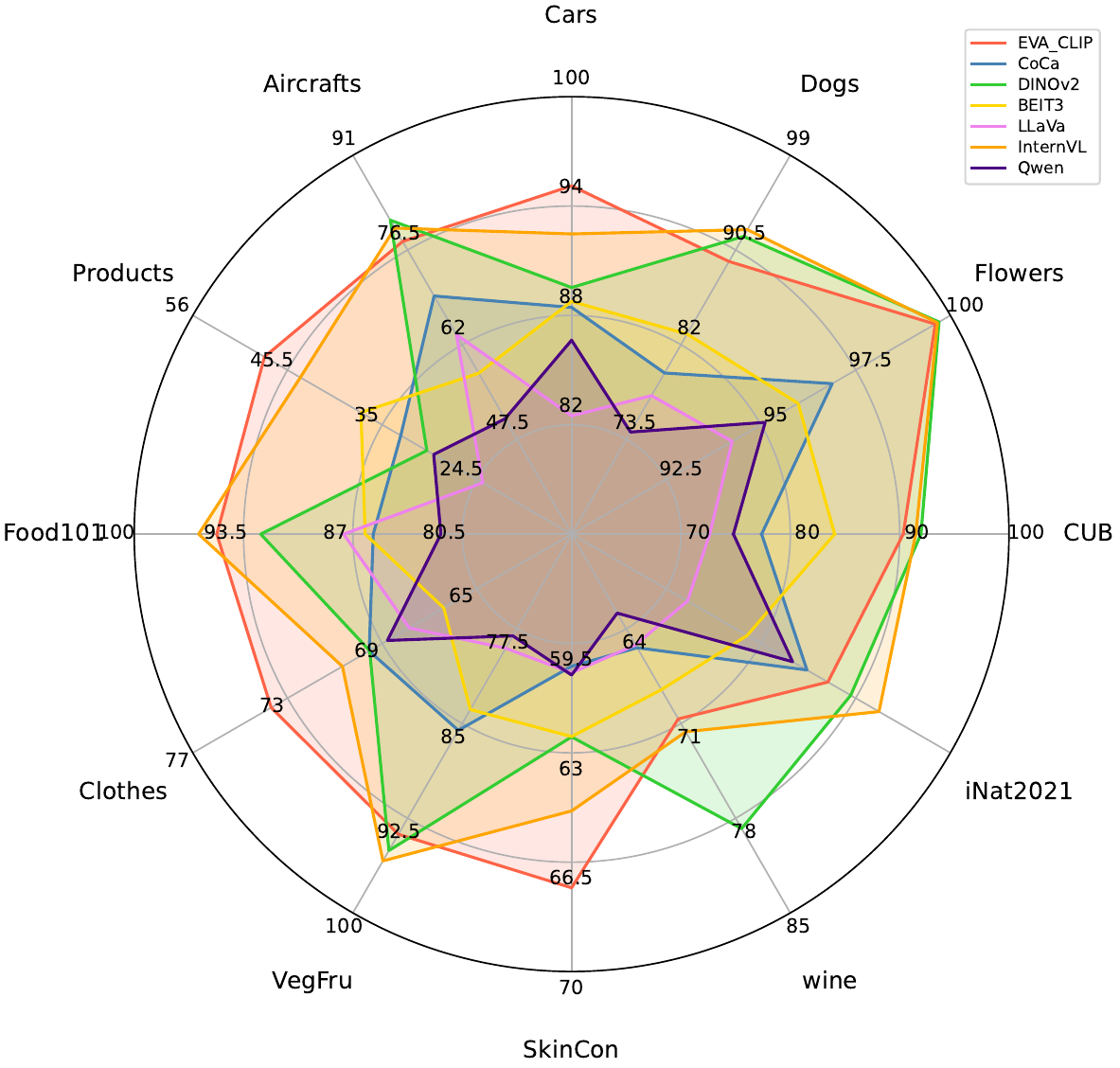}
        \caption{Retrieval results of LVLM visual features on twelve fine-grained datasets. Different colors represent different models.}
        \label{fig:rader_retrieval}
    \end{minipage}
    \hspace{0.04\textwidth}
    \begin{minipage}[b]{0.47\textwidth}
        \centering
        \includegraphics[width=0.7\textwidth]{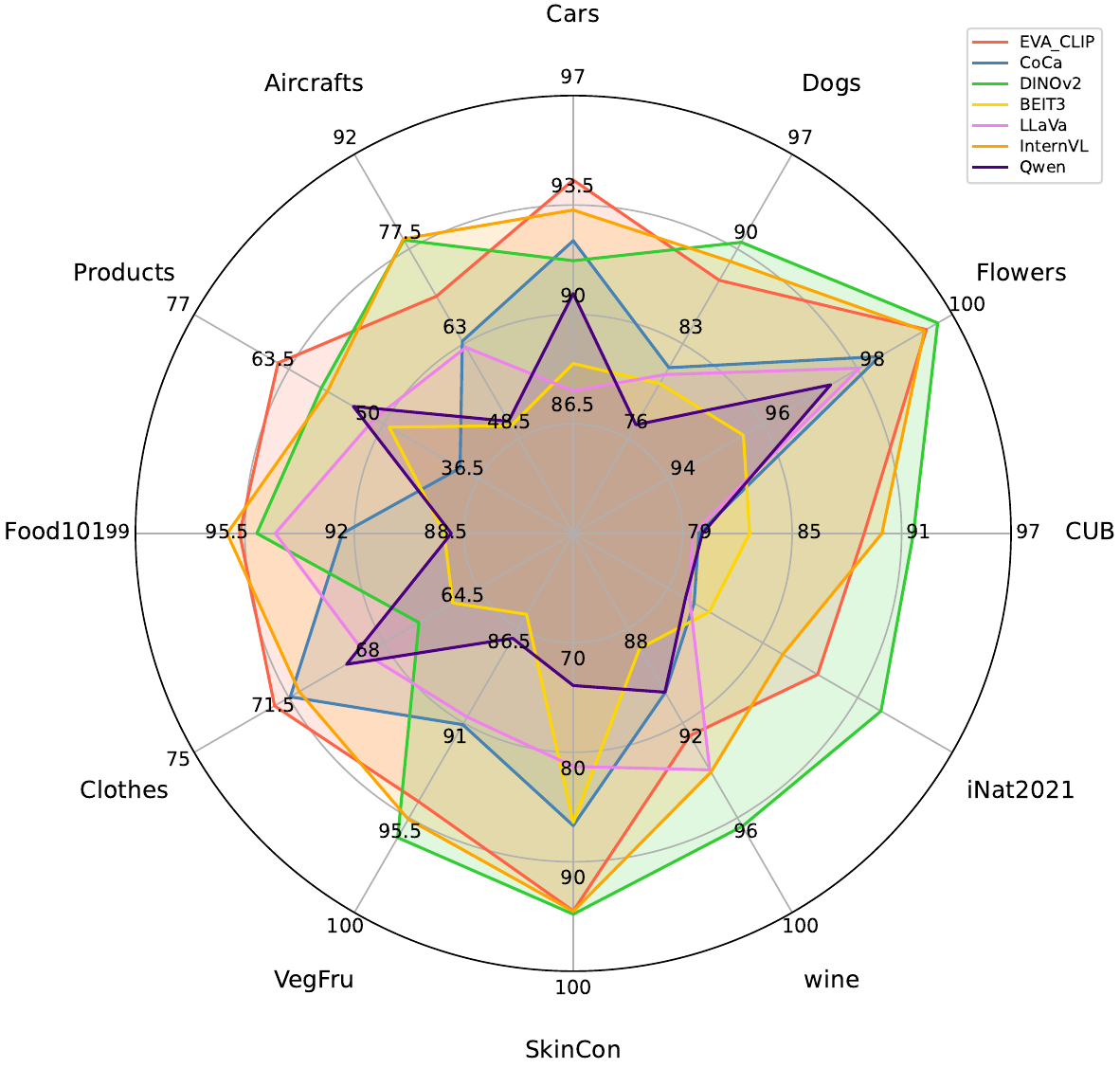}
        \caption{Classification results of LVLM visual features on twelve fine-grained datasets. Different colors represent different models.}
        \label{fig:rader_cls}
    \end{minipage}
    \vspace{-0.7em}
\end{figure*}

\begin{figure*}[t]
    \centering
    \begin{minipage}[b]{0.47\textwidth}
        \centering
        \includegraphics[width=0.7\textwidth]{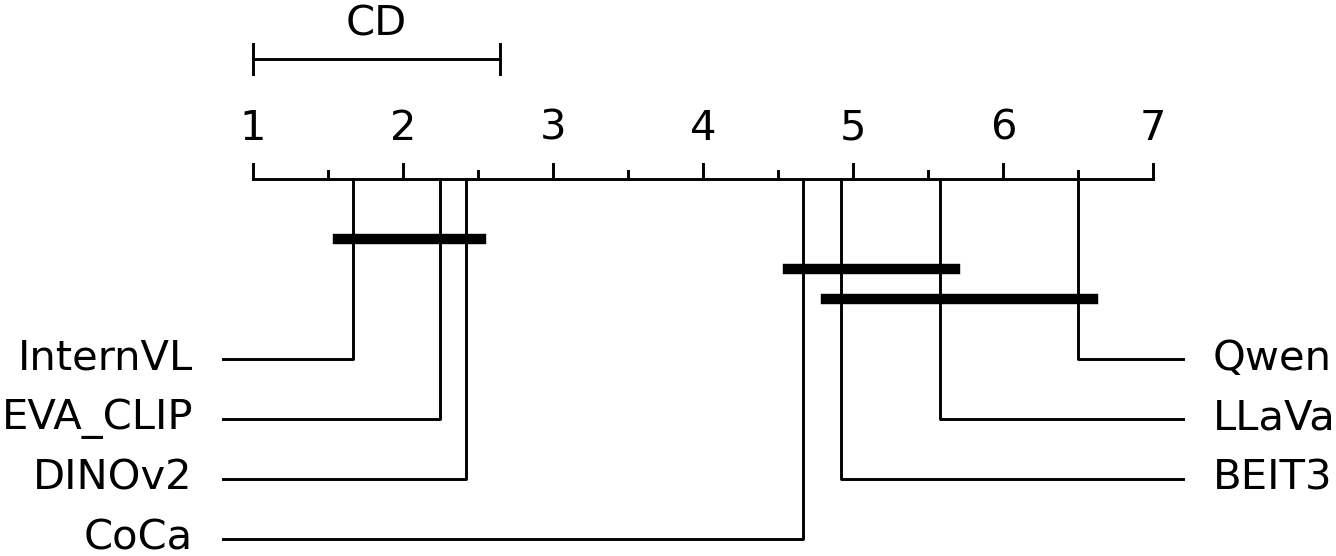}
        \caption{Nemenyi statistical test results for fine-grained retrieval. Black horizontal lines indicate the critical distance (CD), grouping models with no significant performance differences.}
        \label{fig:frd_retrieval}
    \end{minipage}
    \hspace{0.04\textwidth}
    \begin{minipage}[b]{0.47\textwidth}
        \centering
        \includegraphics[width=0.7\textwidth]{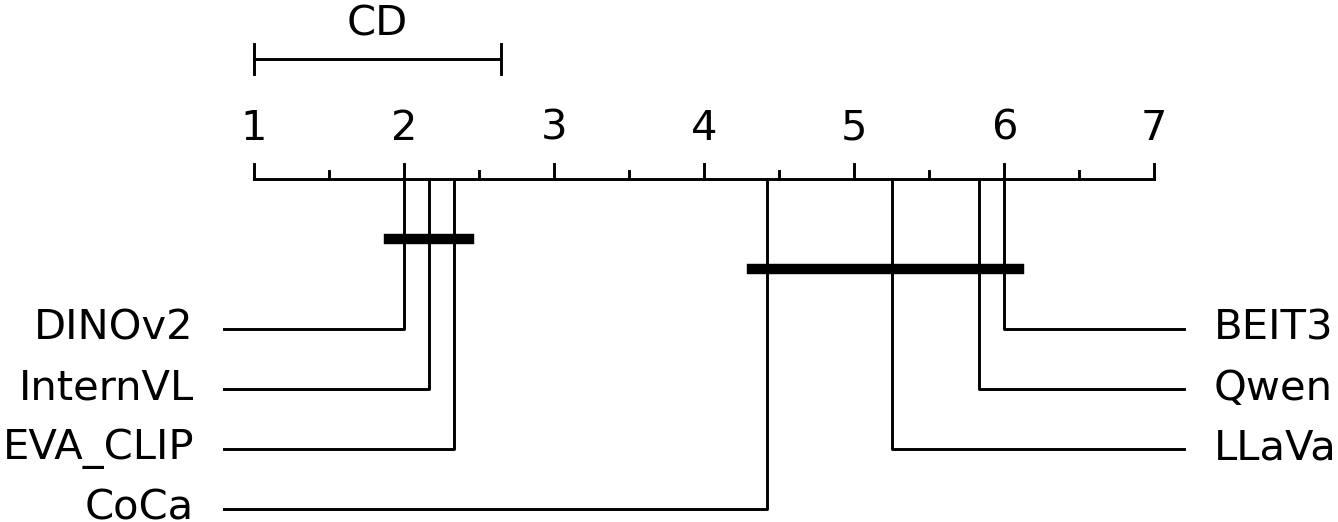}
        \caption{Nemenyi statistical test results for fine-grained recognition. Black horizontal lines indicate the critical distance (CD), grouping models with no significant performance differences.}
        \label{fig:frd_cls}
    \end{minipage}
    \vspace{-0.5em}
\end{figure*}

\begin{figure*}[t]
    \centering
    \begin{minipage}[b]{0.22\textwidth}
        \centering
        \includegraphics[width=\textwidth,height=7em]{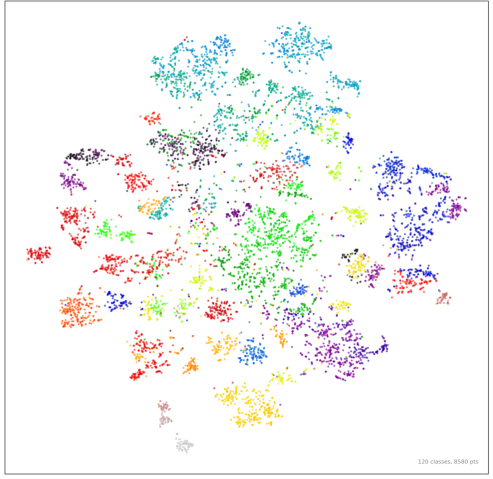}
        \vspace{-0.7em}
        \small (a) EVA-CLIP
    \end{minipage}
    \hfill
    \begin{minipage}[b]{0.22\textwidth}
        \centering
        \includegraphics[width=\textwidth,height=7em]{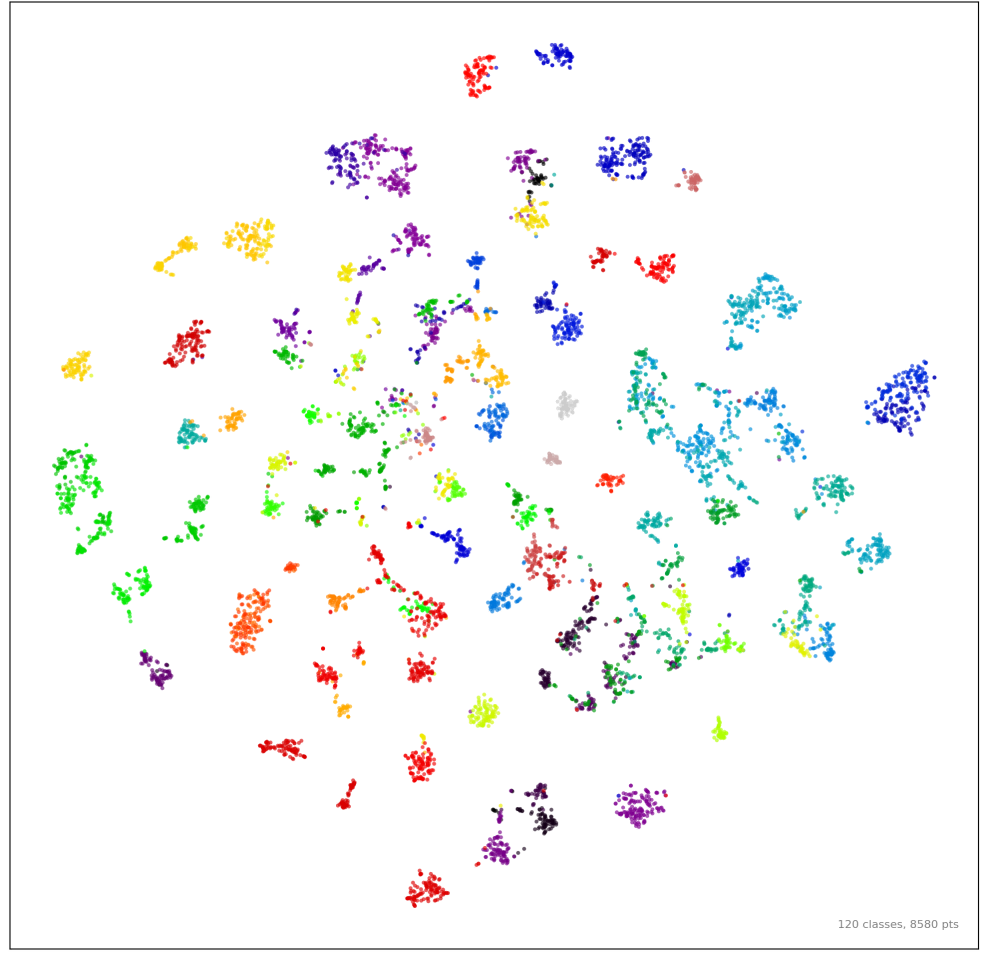}
        \vspace{-0.7em}
        \small (b) DinoV2
    \end{minipage}
    \hfill
    \begin{minipage}[b]{0.22\textwidth}
        \centering
        \includegraphics[width=\textwidth,height=7em]{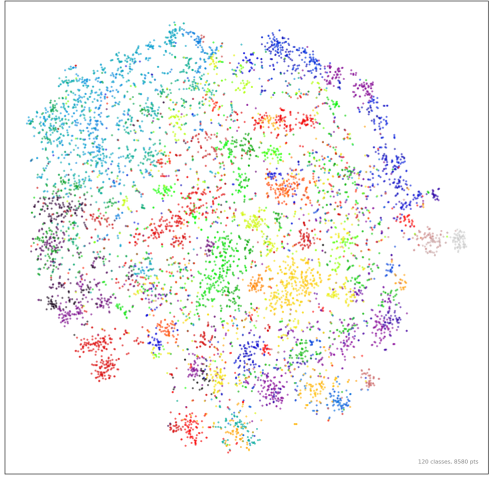}
        \vspace{-0.7em}
        \small (c) BEiT3
    \end{minipage}
    \hfill
    \begin{minipage}[b]{0.22\textwidth}
        \centering
        \includegraphics[width=\textwidth,height=7em]{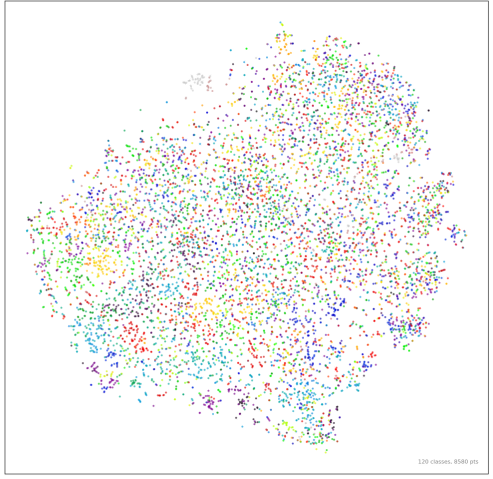}
        \vspace{-0.7em}
        \small (d) Qwen-VL
    \end{minipage}
    \vspace{0.6em}
    \caption{$t$-SNE visualization of visual features on Stanford Dogs.}
    \label{fig:t-SNE_visualization}
    \vspace{-1.0em}
\end{figure*}

\begin{figure*}[t!]
    \centering
    {\includegraphics[width=0.94\textwidth,height=8em]{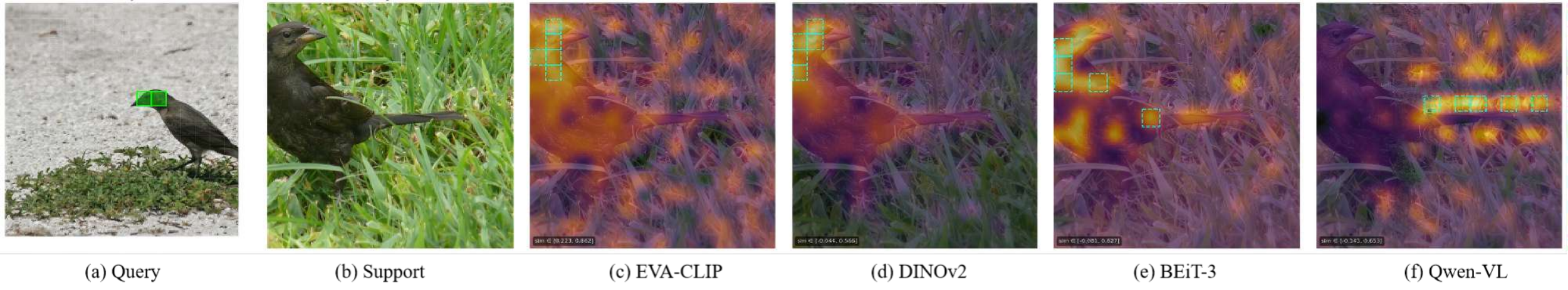}}
    \caption{Patch-level correspondence visualization on CUB datasets. Green boxes in the query images indicate the selected patches, and green boxes in the support images denote the most similar patches retrieved by different models.}
    \label{fig:patch_visualization}
    \vspace{-0.3em}
\end{figure*}

\begin{figure*}[t!]
    \centering
    {\includegraphics[width=0.94\textwidth,height=7em]{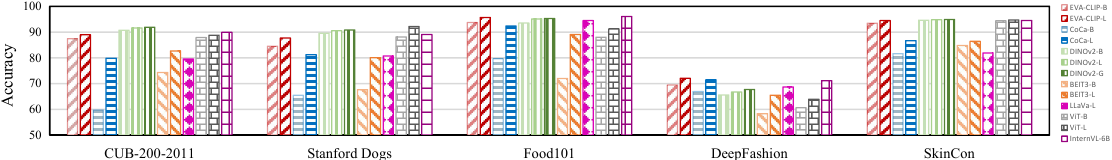}}
    \caption{Classification results with different vision encoder sizes. Bars filled with different patterns represent different models, with darker patterns indicating larger vision encoder sizes.}
    \vspace{-0.5em}
    \label{fig:size_cls}
\end{figure*}

\begin{figure*}[t!]
    \centering
    {\includegraphics[width=0.94\textwidth,height=10em]{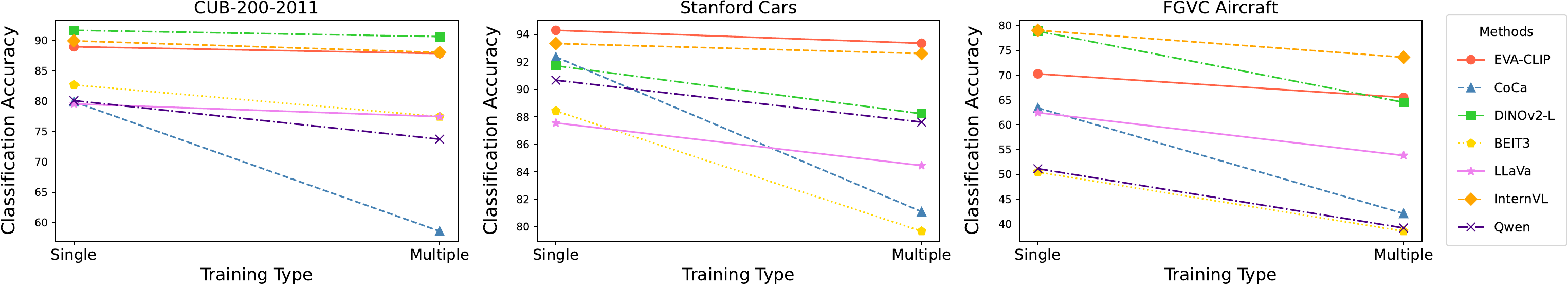}}
    \caption{Classification results of LVLM visual features on fine-grained datasets. ``Single" denotes accuracy from training on a single meta-category, while ``Multiple" reflects accuracy from training on a unified dataset combining multiple meta-categories.}
    \label{fig:combine_cls}
\end{figure*}

These analyses are supported by the fine-grained retrieval and classification results in Figure~\ref{fig:rader_retrieval} and Figure~\ref{fig:rader_cls}, the statistical comparisons in Figure~\ref{fig:frd_retrieval} and Figure~\ref{fig:frd_cls}, the multi-meta-category classification results in Figure~\ref{fig:combine_cls}, the global and local feature visualizations in Figure~\ref{fig:t-SNE_visualization} and Figure~\ref{fig:patch_visualization}, the encoder-size analysis in Figure~\ref{fig:size_cls}, and the SFT data-composition results in Table~\ref{table:fg_general_performance}. Together, they show that improving fine-grained LVLM capabilities requires more than raw scale: effective training objectives, high-quality data, and balanced SFT data composition are all important for strengthening fine-grained recognition while preserving general multimodal abilities.

\begin{findingbox}
    \textit{Finding 8:} The contrastive training paradigm in LVLMs effectively enhances the fine-grained discriminability of visual features.
\end{findingbox}

As shown in Figures~\ref{fig:rader_retrieval} and~\ref{fig:rader_cls}, visual encoders trained with contrastive objectives (\emph{e.g.}, EVA-CLIP, InternVL, and DINOv2) outperform those trained mainly with reconstruction-based objectives (BEiT3) or generative objectives (Qwen) on fine-grained retrieval and classification tasks. The Nemenyi test results in Figures~\ref{fig:frd_retrieval} and~\ref{fig:frd_cls} further show that InternVL, EVA-CLIP, and DINOv2 perform significantly better than Qwen and BEiT3. In multi meta-category classification (cf. Figure~\ref{fig:combine_cls}), EVA-CLIP maintains strong performance, with an average drop of only 1.96\% compared to the single-category setting, whereas Qwen and BEiT3 exhibit larger drops of 4.16\% and 7.41\%, respectively.

These quantitative results are further supported by qualitative visualizations. As shown in Figure~\ref{fig:t-SNE_visualization}, contrastive features form more compact and better-separated clusters on fine-grained datasets, indicating stronger global category separability. At the local level, Figure~\ref{fig:patch_visualization} shows that contrastive features produce more semantically consistent patch correspondences across images, while reconstruction- and generation-based features are more easily distracted by background textures or irrelevant regions. These observations suggest that contrastive training benefits fine-grained recognition not only by improving global feature separability, but also by preserving more reliable local discriminative cues.

We further examine whether this advantage simply comes from larger vision encoders. As shown in Figure~\ref{fig:size_cls}, DINOv2-B, despite using a smaller vision encoder, achieves higher classification accuracy than the larger BEiT3-L, outperforming it by 8.08\% on \emph{CUB-200-2011} and 9.49\% on \emph{Stanford Dogs}. This suggests that training paradigm can be more critical than encoder scale for fine-grained feature learning. A possible reason is that reconstruction- and generation-based objectives do not explicitly enforce inter-category separation and intra-category compactness among visually similar categories, thereby limiting their effectiveness on fine-grained tasks. More results are detailed in Appendix~\ref{app:quali_analy_contras_vs_gen}.

\begin{findingbox}
    \textit{Finding 9:} Scaling vision encoders or web data alone brings limited gains in fine-grained visual discriminability.
\end{findingbox}

After observing the strong effect of training paradigm, we further examine whether raw scale can compensate for limited fine-grained visual discriminability. Regarding vision encoder size, as shown in Figure~\ref{fig:size_cls}, scaling DINOv2's vision encoder from DINOv2-B to DINOv2-L improves the average classification accuracy by only 0.6\%, and further scaling it from DINOv2-L to DINOv2-G brings another marginal gain of only 0.3\%. Moreover, the classification accuracy obtained from InternVL-6B visual features is not higher than that of DINOv2-L, suggesting that merely enlarging the vision encoder is insufficient to substantially improve fine-grained discriminability.

Regarding training-data scale, as shown in Figure~\ref{fig:frd_cls}, EVA-CLIP, whose vision encoder is trained on over 2 billion samples, does not outperform DINOv2, which is trained on 142 million samples, in fine-grained classification and retrieval tasks. We attribute this difference to training-data quality: DINOv2's dataset is carefully curated from a large pool of data, whereas EVA-CLIP relies on crawled web data. A similar trend is observed when comparing DINOv2 with InternVL, whose vision encoder is trained on 6B samples. These results suggest that simply increasing the scale of the vision encoder or training data, without considering objective design and data quality, offers limited gains in fine-grained visual feature discriminability.

\begin{findingbox}
    \textit{Finding 10:} Mixing general and fine-grained data during SFT improves fine-grained recognition while preserving general capabilities.
\end{findingbox}

After examining visual representation factors, we further ask whether fine-grained LVLM capabilities can be improved through supervised fine-tuning. A straightforward strategy is to continue fine-tuning an already SFT-trained LVLM on fine-grained data. As shown in Table~\ref{table:fg_general_performance}, this strategy improves fine-grained recognition performance, but substantially degrades general multimodal capabilities. For example, compared with the model trained on general SFT data only (\#1), the model further tuned on fine-grained data (\#2) drops from 65.31 to 48.67 on AI2D, from 27.36 to 13.45 on ChartQA, and from 42.43 to 20.36 on DocVQA. This indicates that post-hoc fine-grained tuning can introduce severe forgetting of general capabilities.

To mitigate this trade-off, we mix general SFT data and fine-grained data during the SFT stage with a 1:1 sampling ratio. As shown by setting \#3 in Table~\ref{table:fg_general_performance}, joint SFT with general and fine-grained data largely preserves general performance compared with the general-only SFT baseline (\#1), while still achieving strong fine-grained recognition accuracy. For example, its general performance remains close to the baseline on AI2D (65.02 vs. 65.31), ChartQA (26.79 vs. 27.36), MathVista (23.2 vs. 22.6), and POPE (87.6 vs. 87.6). Meanwhile, its short-answer fine-grained results are comparable to the model further tuned on fine-grained data (\#2), and even slightly higher on CUB, Food-101, and Stanford Dogs.

These results suggest that fine-grained supervision is beneficial, but its placement and composition during SFT are critical. Directly tuning an already instruction-tuned LVLM on fine-grained data improves task-specific recognition at the cost of general ability, whereas mixing general and fine-grained data during SFT provides a better balance. This indicates that fine-grained LVLM improvement should not rely on isolated task-specific tuning alone; instead, fine-grained data should be incorporated together with general instruction data so that the model can acquire fine-grained knowledge while maintaining broad multimodal competence.

\subsection{Robustness of Fine-Grained LVLM Recognition under Visual and Linguistic Perturbations}

After analyzing the performance gaps, underlying bottlenecks, and representation-learning factors of LVLMs, we finally examine whether their fine-grained recognition ability is robust under perturbations. This question is particularly important for fine-grained tasks, where predictions often depend on subtle visual cues and precise grounding between visual evidence and category semantics. Small disturbances may therefore weaken the discriminative visual evidence or bias the model toward incorrect semantic decisions.

We evaluate robustness from both the visual and linguistic sides. On the visual side, we first perturb visual inputs using projected gradient descent~\cite{PGD} to examine whether fine-grained representations are more fragile than generic representations, and further apply image-level corruptions to test how degraded visual evidence affects both feature discriminability and dialogue-based recognition. On the linguistic side, we introduce misleading textual cues into the prompt to examine whether language priors can override visual evidence during fine-grained recognition. We also compare different question formats to understand when such linguistic perturbations become more effective.

\begin{table*}[t!]
    \caption{Results of InternVL Trained under Different Settings on Fine-Grained and General Tasks. The ``558k'' Represents the Alignment Data, ``665k'' Represents the Generic Fine-Tuning Data, while ``fg'' Represents the Fine-Grained Data Used in Training. ``Short Answer'' Represents the Results on Questions About the Object Fine-Grained Category.}
    \setlength{\tabcolsep}{7pt}
    \centering
    \small
    \label{table:fg_general_performance}
    \resizebox{\textwidth}{!}{
    \begin{tabular}{|c|c|c|c|c|c|c|c|c|c|}
        \hline
        \multirow{2}{*}{Setting} 
            & \multicolumn{3}{c|}{\rule{0pt}{8pt}Training Process} 
            & \multicolumn{6}{c|}{General Capabilities} \\
        \cline{2-10}
            & \rule{0pt}{8pt}Alignment 
            & FT 
            & FT 
            & \emph{AI2D} 
            & \emph{ChartQA} 
            & \emph{DocVQA} 
            & \emph{InfographicsVQA} 
            & \emph{MathVista} 
            & \emph{POPE} \\
        \hline
        \hline
        \rule{0pt}{8pt}\#1 & 558k & 665k    & -- & 65.31 & 27.36 & 42.43 & 30.27 & 22.6 & 87.6  \\
        \#2               & 558k & 665k    & fg & 48.67 & 13.45 & 20.36 & 18.89 & 16.7 & 83.39 \\
        \#3               & 558k & 665k+fg & -- & 65.02 & 26.79 & 41.11 & 28.34 & 23.2 & 87.6  \\
        \#4               & 558k & fg      & -- & --    & --    & --    & --    & --   & --    \\
        \hline
        \hline
        \multirow{2}{*}{Setting} 
            & \multicolumn{3}{c|}{\rule{0pt}{8pt}Training Process} 
            & \multicolumn{6}{c|}{Fine-grained Recognition Capabilities -- Short Answer} \\
        \cline{2-10}
            & \rule{0pt}{8pt}Alignment 
            & FT 
            & FT 
            & \emph{Aircraft} 
            & \emph{CUB} 
            & \emph{Flowers102} 
            & \emph{Food-101} 
            & \emph{Dog} 
            & \emph{VegFru} \\
        \hline
        \hline
        \rule{0pt}{8pt}\#1 & 558k & 665k    & -- & --    & --    & --    & --    & --    & --    \\
        \#2               & 558k & 665k    & fg & 68.4  & 83.32 & 92.66 & 94.03 & 84.51 & 91.65 \\
        \#3               & 558k & 665k+fg & -- & 66.03 & 83.84 & 92.19 & 94.46 & 85.33 & 90.77 \\
        \#4               & 558k & fg      & -- & 69.45 & 83.43 & 93.54 & 94.25 & 84.41 & 91.79 \\
        \hline
    \end{tabular}
    }
    \vspace{-1em}
\end{table*}

These analyses are supported by the image perturbation results in Table~\ref{table:white-box attack}, the visual corruption results in Table~\ref{table:fg_attack} and Table~\ref{tab:appendix_gb_sp}, and the language-side perturbation analysis in Table~\ref{table:fg_attack}. Together, they show that fine-grained LVLM recognition is vulnerable not only to weakened visual evidence, but also, and more severely, to misleading linguistic cues that directly bias the final semantic decision.

\begin{findingbox}
    \textit{Finding 11:} Visual features in LVLMs are more susceptible to perturbations in fine-grained tasks.
\end{findingbox}

We first examine the robustness of visual representations under white-box image perturbations. Specifically, we use gradients computed from visual features to update the input pixels, and compare the resulting accuracy drop on fine-grained and generic classification tasks. As shown in Table~\ref{table:white-box attack}, applying such perturbations to images encoded by EVA-CLIP sharply reduces the classification accuracy on the fine-grained dataset \emph{CUB-200-2011}, from 88.95\% to 24.94\%. By comparison, the accuracy drop on the generic dataset \emph{CIFAR-100}~\cite{cifar} is less severe, decreasing from 93.05\% to 50.76\%. Similar trends are observed for CoCa and DINOv2, indicating that fine-grained visual representations are more fragile under adversarial image perturbations than generic representations.

This vulnerability may be related to the limited fine-grained discriminability of visual features learned from coarse-grained or noisy training data. Since fine-grained categories often differ only in subtle visual cues, perturbations that slightly shift the visual representation can make closely related categories much harder to distinguish. In contrast, the Vision Transformer~\cite{ViT} trained on the curated \emph{ImageNet}~\cite{deng2009imagenet} dataset with cross-entropy loss demonstrates stronger robustness, showing only minor declines in classification accuracy on both fine-grained and generic datasets. This suggests that adopting alternative training paradigms or incorporating high-quality, fine-grained data (as seen in \emph{ImageNet}) during training could help improve the robustness of visual features in LVLMs.

\begin{findingbox}
    \textit{Finding 12:} Language-side perturbations can override visual evidence more effectively than feature-side perturbations.
\end{findingbox}

Having shown that fine-grained visual representations are vulnerable to visual perturbations, we next compare how perturbations from the visual and linguistic sides affect LVLM predictions. We first apply a range of visual corruptions to the input images, including salt-and-pepper noise, Gaussian blur, background removal, and object-level color shift. As shown in Table~\ref{table:fg_attack} and Table~\ref{tab:appendix_gb_sp}, these perturbations consistently degrade LVLM performance at both the feature and response levels: the discriminability of visual features declines, and the accuracy on fine-grained recognition questions also drops.

However, we find that perturbations on the language side are substantially more effective. When misleading linguistic cues are appended to the prompt (e.g., “the bird in the image seems to be a black-footed albatross”), Qwen2.5-VL drops from 74.04\%/71.49\% to 63.01\%/28.69\% on CUB, corresponding to a 42.80\% drop on true/false questions. A similar trend is observed for InternVL3, whose true/false accuracy decreases by 20.94\%. We attribute this asymmetry to the fact that the final output space of LVLMs is fundamentally linguistic. Visual perturbations mainly weaken the strength of perceptual evidence, which still needs to be interpreted by the language model before producing the final answer. In contrast, language-side perturbations inject an explicit prior directly into the inference process, biasing the model’s decision rule rather than merely degrading its evidence. From a causal perspective, linguistic perturbations are closer to the final prediction, and are therefore more likely to override the effect of visual evidence.

We further observe that the effect of linguistic perturbations depends strongly on the question format. On coarse-grained tasks, misleading prompts have little impact on multiple-choice questions, but still remain highly effective for true/false questions. We attribute this difference to the structure of the answer space. In multiple-choice settings, the correct answer is guaranteed to appear among the options, allowing the model to rely on relative comparison among candidates and partially compensate for the bias introduced by the prompt. In contrast, true/false questions are closer to semantic verification: the model must determine whether a given statement is correct, without the benefit of a constrained candidate set. As a result, when the model’s semantic understanding is weak, misleading linguistic cues can more easily distort its final judgment, which explains the observed trend. More results are detailed in Appendix~\ref{app:res_visual_side_and_language-side_perturbations}

\begin{table*}[t!]
    \caption{Classification Results of LVLMs' Original and Perturbed Visual Features on the Fine-Grained Dataset \emph{CUB-200-2011} and the Generic Dataset \emph{CIFAR-100}. ``Origin'' Refers to Results with Original Features, while ``Perturbed'' Indicates Results with Perturbed Features.}
    \setlength{\tabcolsep}{11pt}
    \centering
    \small
    \label{table:white-box attack}
    \begin{tabular}{|c|c|c|c|c|c|c|c|c|}
        \hline
        \multirow{2}{*}{Datasets} & \multicolumn{2}{c|}{\rule{0pt}{8pt}EVA-CLIP} & \multicolumn{2}{c|}{CoCa} & \multicolumn{2}{c|}{DINOv2} & \multicolumn{2}{c|}{ViT} \\
        \cline{2-9} 
                                  & \rule{0pt}{8pt}Origin      & Perturbed       & Origin     & Perturbed    & Origin      & Perturbed     & Origin & Perturbed \\ \hline
        \hline
        \emph{ \rule{0pt}{8pt}CIFAR-100}                  & 93.05       & 50.76           & 86.94      & 52.23        & 93.38       & 42.39         & 89.81  & 72.15                     \\
        \emph{CUB-200-2011}                       & 88.95       & 24.94           & 79.89      & 23.40        & 91.64       & 25.94         & 88.83  & 73.85                     \\
        \hline 
    \end{tabular}
\end{table*}

\begin{table*}[t!]
    \caption{Robustness of LVLMs under Different Perturbations on Fine-Grained Datasets. Each Entry Reports ``Multiple-Choice and True/False'' Accuracy. GB and SP Denote Gaussian Blur and Salt-and-Pepper Noise; BG-Gray, Color, and Mislead Denote Background, Object-Color, and Textual Perturbations, Respectively. The $\Delta$ Rows Report Performance Drops from Original Inputs.}
    \setlength{\tabcolsep}{3.4pt}
    \centering
    \small
    \label{table:fg_attack}
    \begin{tabular}{|l|c|c|c|c|c|c|c|c|}
        \hline
        \multirow{2}{*}{Models} & \multicolumn{3}{c|}{\rule{0pt}{8pt}\emph{Aircraft}} & \multicolumn{3}{c|}{\emph{Stanford Dogs}} & \multicolumn{2}{c|}{\emph{CUB}} \\
        \cline{2-9}
         & \rule{0pt}{8pt}Original & GB & SP & Original & BG-gray & Color & Original & Mislead \\
        \hline
        \hline
        \rule{0pt}{8pt}Qwen2.5-VL
            & 94.84 / 89.56 & 87.67 / 71.02 & 87.46 / 80.35 & 96.74 / 94.50 & 95.55 / 92.80 & 90.12 / 86.19 & 74.04 / 71.49 & 63.01 / 28.69 \\
        \emph{$\Delta$ vs. Original}
            &  & \emph{7.17 / 18.54} & \emph{7.38 / 9.21} &  & \emph{1.19 / 1.70} & \emph{6.62 / 8.31} &  & \emph{11.03 / 42.80} \\
        InternVL3
            & 85.48 / 86.92 & 80.17 / 84.01 & 79.45 / 83.68 & 93.11 / 92.02 & 91.50 / 90.99 & 83.90 / 85.07 & 61.18 / 62.48 & 51.71 / 41.54 \\
        \emph{$\Delta$ vs. Original}
            &  & \emph{5.31 / 2.91} & \emph{5.03 / 3.24} &  & \emph{1.61 / 1.03} & \emph{9.21 / 6.95} &  & \emph{9.47 / 20.94} \\
        \hline
    \end{tabular}
    \vspace{-1em}
\end{table*}

\section{Concluding Remarks}

In this work, we introduced \texttt{FG-BMK}, a comprehensive benchmark and diagnostic framework for evaluating LVLMs on fine-grained image tasks. Rather than treating fine-grained evaluation as a conventional classification problem, our study examines how LVLMs perceive subtle visual evidence, preserve fine-grained discriminability in their representations, align such evidence with language semantics, and finally produce category-level decisions through dialogue. By jointly considering human-oriented semantic recognition and machine-oriented visual discriminability, \texttt{FG-BMK} provides a structured lens for understanding not only whether LVLMs fail on fine-grained tasks, but also where such failures originate.

The broader implication of our study is that fine-grained visual understanding exposes a fundamental capability boundary of current LVLMs. Existing LVLMs have made substantial progress in open-ended multimodal interaction, but fine-grained tasks require a different level of visual-semantic precision: models must attend to local attributes, compare subtle part-level differences, associate them with subordinate concepts, and resist misleading linguistic priors when visual evidence is weak or ambiguous. Our results suggest that strong general-purpose multimodal ability does not automatically translate into reliable fine-grained understanding. In particular, a model may learn visually separable representations without grounding fine-grained category concepts, or may possess relevant visual evidence but fail to express it correctly through the language interface. This distinction is important for future LVLM research, because many real-world applications---such as biodiversity monitoring, industrial inspection, medical image analysis, remote sensing, and product recognition---depend precisely on this ability to connect subtle visual patterns with specialized semantic knowledge.

Our findings further indicate that improving fine-grained LVLMs requires more than simply scaling model size or training data. Future LVLMs should incorporate granularity-aware vision-language alignment, stronger local and part-level visual modeling, and fine-grained instruction data that can enrich category-level knowledge without compromising general multimodal capabilities. For specialized domains, models also need mechanisms for acquiring and updating domain-specific visual semantics, so that discriminative representations can be effectively translated into meaningful decisions. Moreover, robustness to linguistic priors should become an important evaluation criterion, since LVLM outputs are produced through a language-centric interface that can override visual evidence in fine-grained reasoning.

Looking forward, \texttt{FG-BMK} can serve as a foundation for studying fine-grained multimodal intelligence beyond static recognition. Promising directions include building fine-grained LVLMs with explicit attribute- and part-aware reasoning, developing alignment strategies that preserve visual discriminability while strengthening semantic grounding, extending fine-grained evaluation to more open-world and dynamic scenarios, and exploring how unified understanding-generation models can learn category concepts that are both visually faithful and semantically precise. We hope this work encourages the community to view fine-grained visual understanding not as a narrow downstream task, but as a critical testbed for whether LVLMs can achieve reliable, grounded, and domain-aware multimodal intelligence.

\bibliography{main}
\bibliographystyle{IEEEtran}

\clearpage
\appendices

\renewcommand{\thesubsection}{\thesection.\arabic{subsection}}
\renewcommand{\thesubsubsection}{\thesubsection.\arabic{subsubsection}}

\renewcommand{\thesubsectiondis}{\thesubsection}
\renewcommand{\thesubsubsectiondis}{\thesubsubsection}

\begin{strip}
\vspace*{-0.5em}
\begin{center}
{\normalfont\LARGE  Supplementary Material of Benchmarking Large Vision-Language Models on Fine-Grained Image Tasks: From Evaluation to Diagnosis\par}
\end{center}
\vspace{0.8em}
\end{strip}

\section{The Evaluation Benchmark}\label{app:benchmark}

\subsection{Evaluation Task Details}\label{app:eval_details}

In Section~\ref{subsec:task_and_metrics}, we have described each evaluation task. Here, we provide further details. In the Knowledge Bias Estimation task, to uncover potential knowledge biases across different fine-grained categories, we pair each image with its corresponding fine-grained label to generate positive samples for true/false questions. For constructing negative samples, each image is paired with a single fine-grained label randomly selected from other subcategories within the same super-category. For each fine-grained category, we calculate the LVLM's accuracy on all coresponding true/false questions as a measure of its understanding of that category's knowledge.

In the cross meta-class classification task, we follow the DINOv2~\cite{DINOv2} method to train the model on a unified training set where fine-grained categories from different datasets are combined. The model is then tested on each individual dataset to evaluate its performance.

\subsection{Data Curation}\label{app:data_curate_samples}

\paragraph{Dataset}
We source images for the \texttt{FG-BMK} benchmark from 13 fine-grained datasets. These datasets cover a wide range of meta-classes, with different categories and sample, providing a comprehensive assessment of LVLMs capabilities on fine-grained tasks across different domains. Table~\ref{table:appendix_fg-datasets} indicates their meta-classes, the amount of samples, the number of categories. For all datasets, we construct human-oriented evaluation questions based on their test sets. We use the original labels directly from the datasets for the machine-oriented evaluation.

\begin{figure*}[h]
    \centering
    {\includegraphics[width=0.99\textwidth]{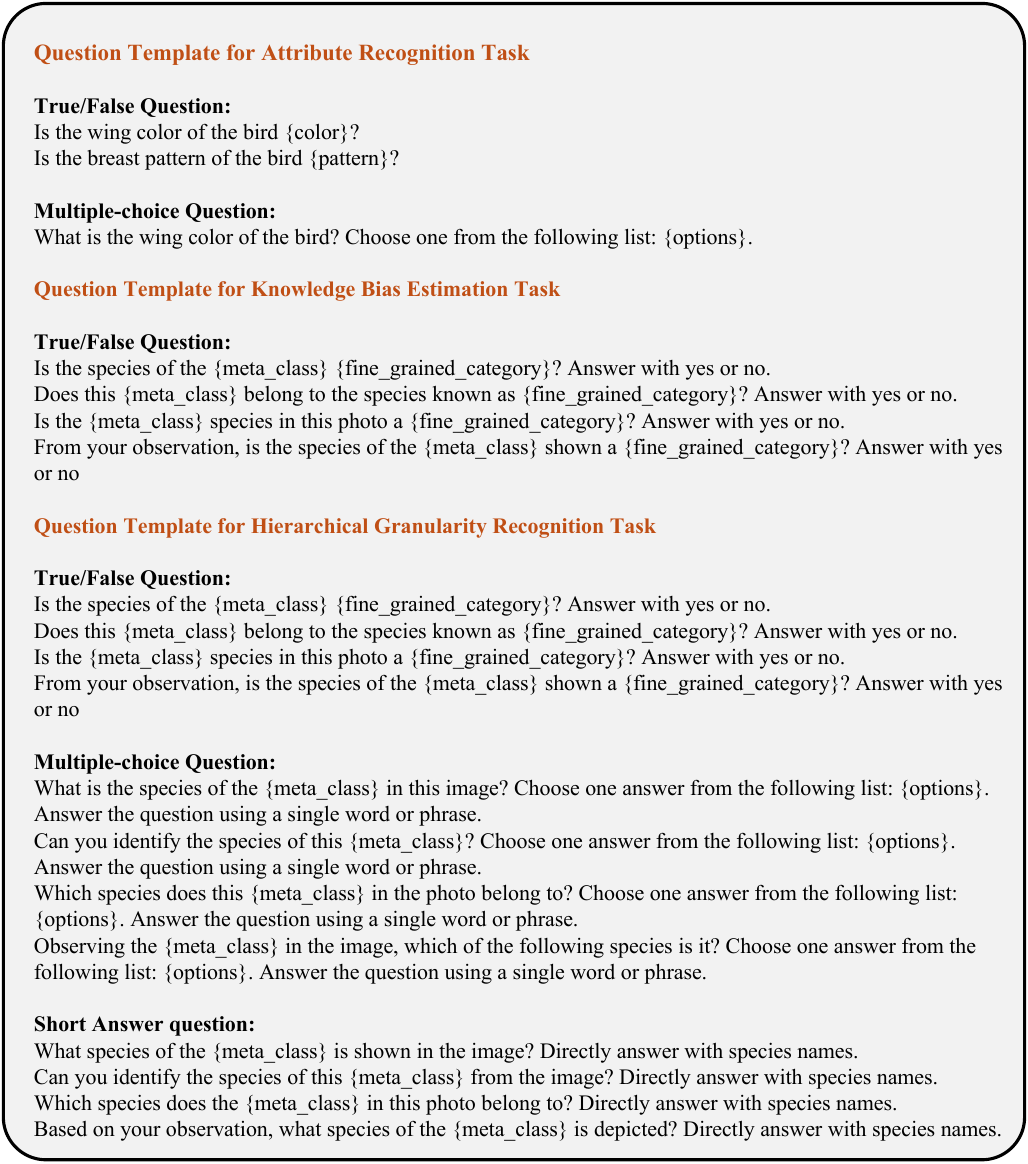}}
    \caption{Question templates for each task in huamn-oriented evaluation.}
    \vspace{-1em}  
    \label{fig:app_prompt_template}
\end{figure*}

\paragraph{Human-oriented Question Templates}
When constructing true/false, multiple-choice, short answer questions for each task in human-oriented evaluation, we manually design several question templates to ensure both diversity and comprehensive coverage. Figure~\ref{fig:app_prompt_template} illustrates the question templates we use for generating the tasks.

We also expanded the original template set to 10 diverse human-written prompts and reconstructed the multiple-choice questions in the human-oriented benchmark to examine the potential impact of linguistic diversity. As shown in Table~\ref{table:attributes_InternVL3_appendix} and Table~\ref{table:tf_mc_InternVL3_appendix}, increasing the number of templates leads to only minor changes in accuracy, and the overall LVLM behavior and observed trends remain consistent. Therefore, as long as the template clearly states the question, the effect of the template quantity on the results is negligible.

\begin{table}[!t]
    \caption{Details of 13 Fine-Grained Datasets Sorted by Their Numbers of Categories. ``Meta-Class'' Refers to a High-Level Categorization of the Dataset. ``Categories'' Refers to the Number of Fine-Grained Categories. ``Samples'' Refers to the Total Number of Samples in Each Dataset.}
    \label{table:appendix_fg-datasets}
    \centering
    \setlength{\tabcolsep}{4pt}
    \begin{tabular}{|c|c|c|c|}
        \hline
        Datasets                     & Meta-class  & Categories & Samples \\
        \hline
        \hline
        \emph{Wine~\cite{baijiu}}          & Industrial  & 11     & 4,516       \\
        \emph{MTARSI-Fixed~\cite{MTARSI}}          & Aircraft  & 27     & 9,114       \\
        \emph{DeepFashion~\cite{liu2016deepfashion}}   & Clothes     & 46     & 18,000      \\
        \emph{SkinCon~\cite{daneshjou2022skincon}}       & Dermatology & 48     & 3,866       \\
        \emph{Flowers102~\cite{nilsback2008automated}}    & Flower      & 102    & 7,169       \\
        \emph{Food101~\cite{food101}}       & Food        & 101    & 101,000     \\
        \emph{FGVC Aircraft~\cite{aircraft}} & Aircraft    & 100    & 6,667       \\
        \emph{Stanford Dogs~\cite{stanforddog}} & Dog         & 120    & 20,580      \\
        \emph{Stanford Cars~\cite{stanfordcar}} & Car         & 196    & 16,185      \\
        \emph{CUB-200-2011~\cite{CUB}}  & Bird        & 200    & 11,788      \\
        \emph{VegFru~\cite{hou2017vegfru}}        & Vegetable   & 292    & 146,131     \\
        \emph{Products-10K~\cite{bai2020products}}  & Retail      & 9,691  & 197,307     \\
        \emph{iNat2021~\cite{iNat2021}}      & Plants      & 10,000 & 2,786,843   \\
        \hline
    \end{tabular}
\end{table}

\begin{table*}[!t]
    \caption{Attribute recognition accuracy of InternVL3~\cite{zhu2025internvl3} using original and extended prompts on the \emph{CUB-200-2011}~\cite{CUB} dataset (values in parentheses represent the average accuracy for each attribute). Accuracy are shown in the format ``original / extended'', with the left representing accuracy using the original prompt and the right using the extended prompt.}
    \setlength{\tabcolsep}{5pt}
    \vspace{-0.5em}
    \label{table:attributes_InternVL3_appendix}
    \centering
    \small
    \begin{tabular}{|c|c|c|c|c|c|}
        \hline
        \multicolumn{6}{|c|}{\textbf{\rule{0pt}{8pt}Color Attribute (47.40 / 47.45)}} \\
        \hline\hline
        \rule{0pt}{8pt}belly color       & 58.49 / 60.04 & back color       & 34.98 / 36.33 & bill color       & 51.31 / 49.64 \\
        breast color     & 54.25 / 55.91 & crown color      & 55.30 / 54.01 & eye color        & 84.59 / 82.96 \\
        forehead color   & 53.32 / 51.90 & leg color        & 44.01 / 45.67 & nape color       & 39.24 / 38.02 \\
        throat color     & 52.77 / 54.53 & under tail color & 34.69 / 35.80 & underparts color & 56.20 / 55.08 \\
        upper tail color & 37.30 / 38.77 & upperparts color & 28.75 / 27.50 & wing color       & 30.16 / 31.88 \\
        primary color    & 43.05 / 41.29 &                 &               &                 &               \\
        \hline\hline

        \multicolumn{6}{|c|}{\textbf{\rule{0pt}{8pt}Pattern Attribute (50.13 / 50.28)}} \\
        \hline\hline
        \rule{0pt}{8pt}back pattern   & 40.94 / 39.38 & belly pattern  & 68.13 / 67.00 & breast pattern & 65.12 / 66.87 \\
        head pattern     & 35.92 / 34.66 & tail pattern   & 41.64 / 42.93 & wing pattern   & 49.04 / 50.84 \\
        \hline\hline

        \multicolumn{6}{|c|}{\textbf{\rule{0pt}{8pt}Shape Attribute (30.95 / 31.01)}} \\
        \hline\hline
        \rule{0pt}{8pt}bill shape & 37.61 / 36.41 & shape     & 52.37 / 50.60 & tail shape & 10.42 / 12.04 \\
        wing shape        & 23.39 / 24.98 &           &               &           &               \\
        \hline\hline

        \multicolumn{3}{|c|}{\textbf{\rule{0pt}{8pt}Length Attribute (71.03 / 69.71)}} & \multicolumn{3}{c|}{\textbf{Size Attribute (52.55 / 54.21)}} \\
        \hline\hline
        \multicolumn{2}{|c|}{\rule{0pt}{8pt}bill length} & \multicolumn{1}{c|}{71.03 / 69.71} &
        \multicolumn{2}{c|}{size} & \multicolumn{1}{c|}{52.55 / 54.21} \\
        \hline
    \end{tabular}
\end{table*}

\begin{table*}[!t]
    \caption{Results of InternVL3 using original and extended prompts on true/false (TF) and multiple-choice (MC) questions across different levels of granularity on the \emph{CUB-200-2011} dataset. Results are shown in the format ``original / extended''.}
    \setlength{\tabcolsep}{13pt}
    \vspace{-0.5em}
    \label{table:tf_mc_InternVL3_appendix}
    \centering
    \small
    \begin{tabular}{|c|c|c|c|c|c|}
        \hline
        \multicolumn{6}{|c|}{\textbf{\rule{0pt}{8pt}TF}} \\
        \hline\hline
        \rule{0pt}{8pt}Class       & 98.79 / 98.03 & Genus       & 85.69 / 86.19 & Species       & 61.88 / 62.34 \\
        \hline\hline

        \multicolumn{6}{|c|}{\textbf{\rule{0pt}{8pt}MC}} \\
        \hline\hline
        \rule{0pt}{8pt}Class       & 99.42 / 99.58 & Genus       & 88.13 / 87.62 & Species       & 60.15 / 59.23 \\
        \hline
    \end{tabular}
\end{table*}

\section{Evaluated Models}\label{app:evaluated_models}
As shown in Table~\ref{table:model_detail_appendix1}, we select nine widely-used open-source LVLMs, two closed-source models (GPT-5.4~\cite{GPT4} and Gemini-3.5-flash~\cite{Gemini}) and one purely visual model, each of which employs a distinctive training recipes, including variations in vision encoder, language model, training losses and data.

\begin{table*}[!htb]
    \centering
    \caption{Configurations of the evaluated models. ``DINOv2-L'' is a purely visual model. ``Con'' stands for the contrastive loss, ``Gen'' for the generative loss, ``Mat'' for the image-text matching loss, ``Rec'' for the reconstruction loss as used in BEiT3~\cite{BEiT3}, and ``Dis'' for the distillation loss as applied in DINOv2~\cite{DINOv2}.}
    \small 
    \vspace{-0.5em}
    \label{table:model_detail_appendix1}
    \begin{tabular}{|c|c|c|c|c|c|c|c|}
        \hline
        \multirow{2}{*}{Model}       & \multicolumn{2}{c|}{\rule{0pt}{8pt}Component} & \multicolumn{5}{c|}{Loss Function}                                   \\
        \cline{2-8}
                               & \rule{0pt}{8pt}Vision Model  & Language Model & Con & Gen & Mat & Rec & Dis \\ \hline
        \hline

        \rule{0pt}{8pt}InternV3-7B     & InternViT-L     & Qwen2.5-7B      & \checkmark           & \checkmark          & \checkmark        &                &  \checkmark            \\
        
        InternVL-Chat-V1.1     & InternViT-6B     & LLaMA2-13B      & \checkmark           & \checkmark          & \checkmark        &                &              \\
        LLaVA-1.5-7B           & CLIP-L           & Vicuna-7B       &             & \checkmark          &          &                &              \\
        Qwen2.5-VL                & CLIP-600M       & Qwen2.5-7B         &  \checkmark           & \checkmark          &  \checkmark        &                &  \checkmark            \\
        Qwen-VL                & Openclip-G       & Qwen-7B         &             & \checkmark          &          &                &              \\
        BLIP-2-FLAN-T5-XL      & EVA-CLIP-G       & FlanT5-XL       & \checkmark            & \checkmark           & \checkmark         &                &              \\
        EVA02-CLIP-L           & EVA02-L          & CLIP-L          & \checkmark           &            &          &                &              \\
        BEiT3-L-ITC            & CLIP-L           & CLIP-L          &             &            &          & \checkmark              &              \\
        CoCa-L                 & CLIP-L           & CLIP-L          & \checkmark           & \checkmark          &          &                &              \\
        DINOv2-L               & CLIP-L           & /               & \checkmark           &            &          &                & \checkmark            \\      
        \hline
    \end{tabular}
\end{table*}

\begin{itemize}[leftmargin=*, itemsep=0pt, topsep=0pt]

\item \textbf{EVA-CLIP}~\cite{EVA-CLIP} aligns visual and textual features using contrastive loss, leveraging over 2 billion web image-text pairs and advanced optimization techniques.
\item \textbf{InternVL3}~\cite{zhu2025internvl3} adopts a unified pre-training approach over both multimodal and pure-text data, enhanced by variable visual position encoding (V2PE) and advanced post-training strategies for improved scalability and effectiveness.
\item \textbf{InternVL}~\cite{chen2024internvl} leverages contrastive, matching, and generative losses in a multi-stage training process, with a large-scale vision encoder and over 6 billion image-text pairs to align visual and textual representation.
\item \textbf{BLIP-2}~\cite{BLIP2} bridges the modality gap between frozen image encoders and LLMs using a lightweight Q-Former, leveraging contrastive, matching, and generative loss in a two-stage pre-training process over 129 million data with fewer trainable parameters.
\item \textbf{Qwen2.5-VL}~\cite{Qwen2.5-VL} combines dynamic-resolution Vision Transformer with Window Attention to reduce computational cost while preserving native image resolution.
\item \textbf{Qwen-VL}~\cite{Qwen-VL} employs a three-stage training process with generative loss, using a VL adapter to align visual and textual features while reducing computational cost over 1.4 billion image-text pairs.
\item \textbf{CoCa}~\cite{CoCa} adopts task-specific attentional pooling to tailor visual representations for different training objectives, applying contrastive loss to train the first half of the decoder and generative loss to train the full decoder in an end-to-end manner over 5 billion image-text pairs.
\item \textbf{BEIT3}~\cite{BEiT3} treats images as a foreign language, leveraging a mask-then-predict objective over 36 million image-text pairs to unify vision and language pretraining, and introduces a multiway transformer architecture for general-purpose modeling.
\item \textbf{LLaVA}~\cite{LLaVA15} aligns visual and textual features using a simple MLP with generative loss, leveraging 1.2 million GPT-4~\cite{GPT4} generated multimodal instruction-following data for training.
\item \textbf{DINOv2}~\cite{DINOv2} uses a self-supervised learning approach, leveraging knowledge distillation and a mask-then-predict strategy over 142 million images to train the vision encoder.
\end{itemize}

For all our evaluated model, we follow their official configurations to run the inference. We set the temperature of all open-source models to 0, while keeping the default for closed-source APIs. 

\section{Human-oriented Evaluations}\label{app:human_oriented_evaluations}

\subsection{Results of Hierarchical Granularity Recognition}\label{app:res_hierar}
\begin{figure*}[!t]
    \centering
     \begin{minipage}{0.45\textwidth}
        \centering
        \includegraphics[width=\textwidth,height=4cm]{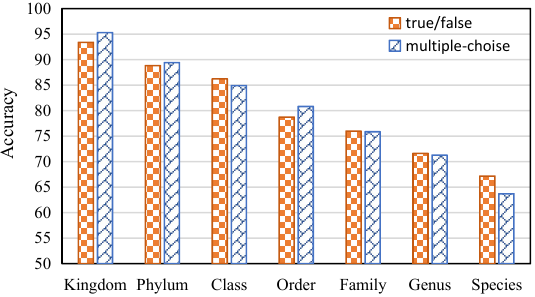}  
        \subcaption{GPT-5.4 on \emph{iNat2021}}  
        \label{fig:app_gpt54_iNat}
    \end{minipage}
    \begin{minipage}{0.45\textwidth}
        \centering
        \includegraphics[width=\textwidth,height=4cm]{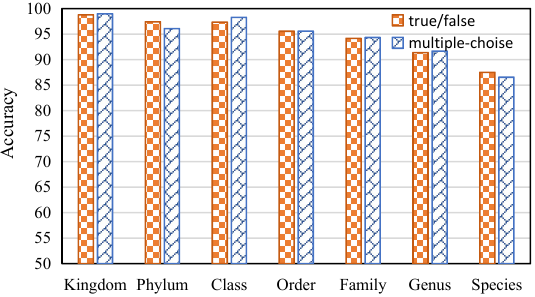}
        \subcaption{Genimi-3.5-flash on \emph{iNat2021}}  
        \label{fig:app_gemini35_iNat}
    \end{minipage}
    
     \begin{minipage}{0.45\textwidth}
        \centering
        \includegraphics[width=\textwidth,height=4cm]{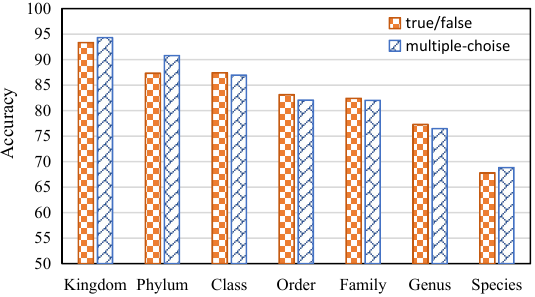}  
        \subcaption{GPT-4o on \emph{iNat2021}}  
        \label{fig:app_gpt_iNat}
    \end{minipage}
    \begin{minipage}{0.45\textwidth}
        \centering
        \includegraphics[width=\textwidth,height=4cm]{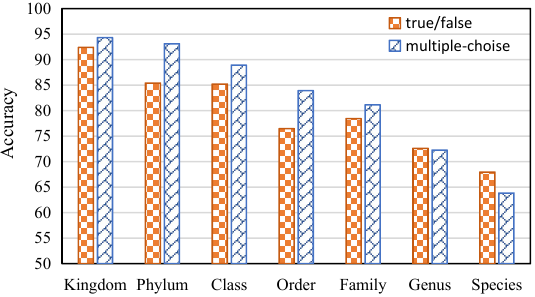}
        \subcaption{Genimi-2.0-flash on \emph{iNat2021}}  
        \label{fig:app_gemini_iNat}
    \end{minipage}
    
    
     \begin{minipage}{0.45\textwidth}
        \centering
        \includegraphics[width=\textwidth,height=4cm]{figure/app_hierarchical_llava_iNat.pdf}  
        \subcaption{LLaVA on \emph{iNat2021}}  
        \label{fig:app_llava_iNat}
    \end{minipage}
    \begin{minipage}{0.45\textwidth}
        \centering
        \includegraphics[width=\textwidth,height=4cm]{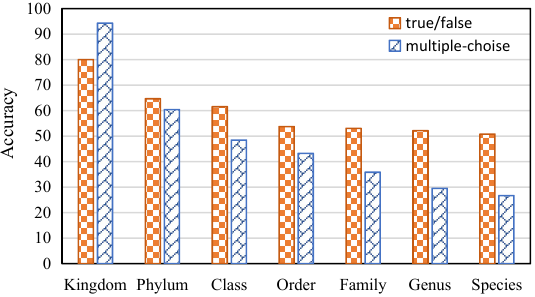}
        \subcaption{InternVL on \emph{iNat2021}}  
        \label{fig:app_intern_iNat}
    \end{minipage}
    

     \begin{minipage}{0.45\textwidth}
        \centering
        \includegraphics[width=\textwidth,height=4cm]{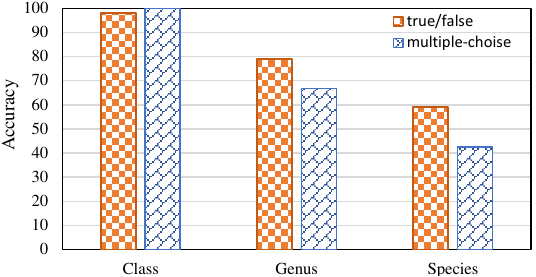}  
        \subcaption{LLaVA on \emph{CUB-200-2011}}  
        \label{fig:app_llava_cub}
    \end{minipage}
    \begin{minipage}{0.45\textwidth}
        \centering
        \includegraphics[width=\textwidth,height=4cm]{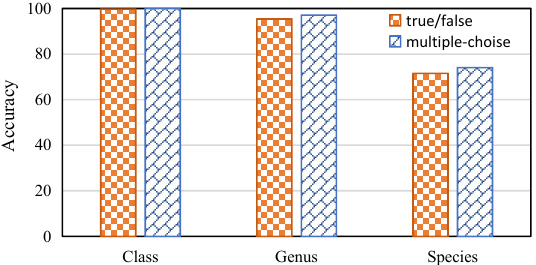}
        \subcaption{Qwen2.5-VL on \emph{CUB-200-2011}}  
        \label{fig:app_qwen25_cub}
    \end{minipage}
    

    \begin{minipage}{0.45\textwidth}
        \centering
        \includegraphics[width=\textwidth,height=4cm]{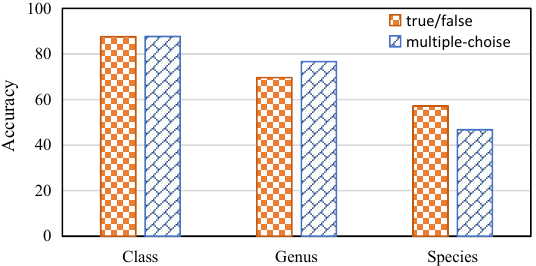}
        \subcaption{InternVL on \emph{CUB-200-2011}}  
        \label{fig:app_intern_cub}
    \end{minipage}
    \caption{Results of GPT-5.4~\cite{GPT4}, GPT-4o~\cite{GPT4}, Gemini-3.5-flash~\cite{Gemini}, Gemini-2.0-flash~\cite{Gemini}, Qwen2.5-VL~\cite{Qwen2.5-VL}, LLaVA~\cite{LLaVA15} and InternVL~\cite{chen2024internvl} on true/false and multiple-choice questions across different levels of granularity on \emph{CUB-200-2011}~\cite{CUB} and \emph{iNat2021}~\cite{iNat2021} dataset. The x-axis denotes the granularity of the recognition questions.}
    \label{fig:app_all_hierarchical}
    \vspace{-0.5em}
\end{figure*}

Figure~\ref{fig:sec42_granularity} shows InternVL3's~\cite{zhu2025internvl3} accuracy in answering true/false and multiple-choice questions within hierarchical granularity recognition task on \emph{CUB-200-2011} dataset. In Figure~\ref{fig:app_all_hierarchical}, we present additional results for GPT-5.4~\cite{GPT4}, GPT-4o~\cite{GPT4}, Gemini-3.5-flash~\cite{Gemini}, Gemini-2.0-flash~\cite{Gemini}, Qwen2.5-VL~\cite{Qwen2.5-VL}, LLaVA~\cite{LLaVA15} and InternVL~\cite{chen2024internvl} on \emph{CUB-200-2011}~\cite{CUB} and \emph{iNat2021}~\cite{iNat2021} datasets. As shown in the experiments, the accuracy of all models decreases as the granularity becomes finer. When the granularity level reaches the finest level, the models struggle to distinguish between closely related species.

\subsection{Results of Knowledge Bias Estimation}\label{app:res_knowledge}
\begin{figure*}[t!]
    \centering
    \begin{minipage}{0.45\textwidth}
        \centering
        \includegraphics[width=\textwidth,height=4.8cm]{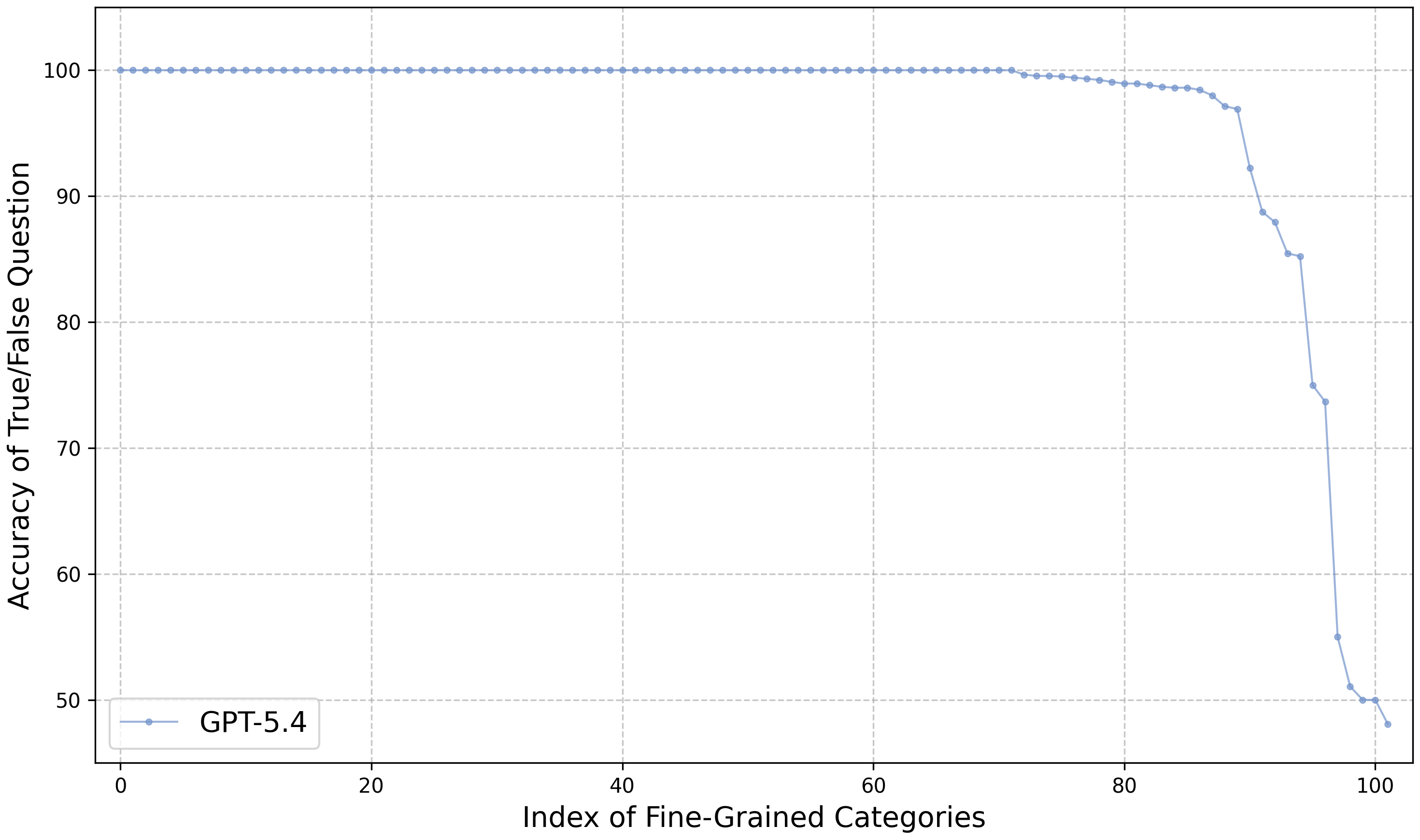}
        \subcaption{GPT-4o on \emph{Flowers102}}  
        \label{fig:app_gpt4o_flowers102_each}
    \end{minipage}
    \begin{minipage}{0.45\textwidth}
        \centering
        \includegraphics[width=\textwidth,height=4.8cm]{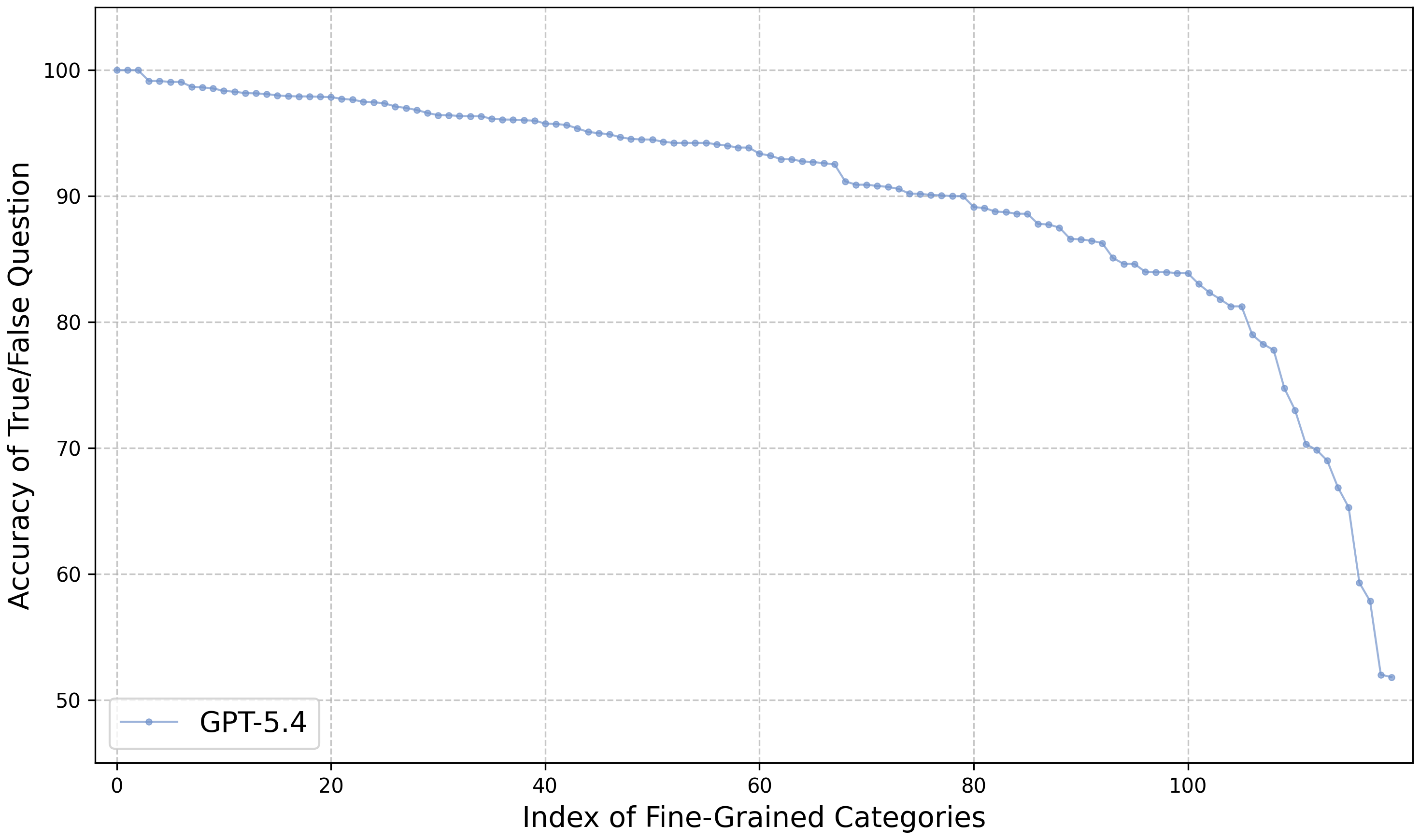}
        \subcaption{Gemini-2.0-flash on \emph{Stanford Dog}}  
        \label{fig:app_gemini_stanforddog_each}
    \end{minipage}
    \caption{Knowledge bias estimation results of two closed-source models. True/false question accuracy for each category is ranked, with blue dots representing the original model.}
    \label{fig:app_inconsist_close_models}
    \vspace{-0.6em}
\end{figure*}

\begin{figure}[t!]
    \centering

    \includegraphics[width=0.45\textwidth,height=4.8cm]{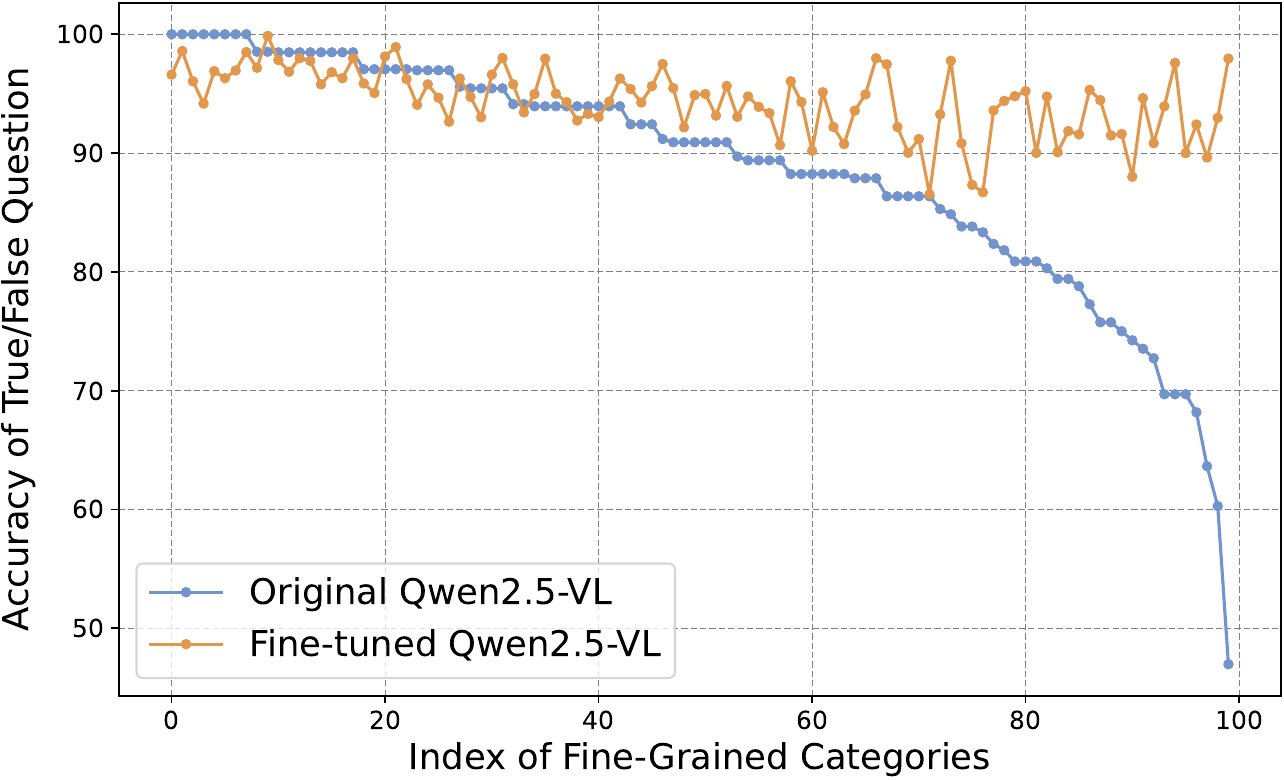} 

    \caption{Comparison of the original and fine-tuned Qwen2.5-VL~\cite{Qwen2.5-VL} models on occurrence-balanced fine-grained aircraft categories. True/false question accuracy for each category is ranked, with blue dots representing the original model and yellow dots the fine-tuned model.}
    \label{fig:app_inconsistence_Qwen25_aircraft}
    \vspace{-0.6em}
\end{figure}

In Figure~\ref{fig:sec42_each_species}, we observe that LLaVA exhibit highly inconsistent recognition abilities across categories. We also conduct experiments with Qwen2.5-VL, GPT-5.4, GPT-4o, Gemini-3.5-flash and Gemini-2.0-flash on fine-grained datasets such as Aircraft~\cite{aircraft}, Flowers102~\cite{nilsback2008automated} and Stanford Dogs~\cite{stanforddog}. As shown in Figure~\ref{fig:app_inconsist_close_models} and Figure~\ref{fig:app_inconsistence_Qwen25_aircraft}, all LVLMs display similar trends, indicating inconsistent recognition abilities across fine-grained categories. However, after fine-tuned on datasets with balanced occurrences of fine-grained categories, LVLMs demonstrate remarkable recognition abilities across all fine-grained categories.

To construct datasets with balanced occurrences of fine-grained categories, we select an equal number of images from each category. Then we generate the same number of true/false questions for each fine-grained category, thereby fine-tuning the LVLMs in a way that each category receives balanced representation.

\begin{figure*}[t!]
    \centering
    \begin{minipage}{0.45\textwidth}
        \centering
        \includegraphics[width=\textwidth,height=4.8cm]{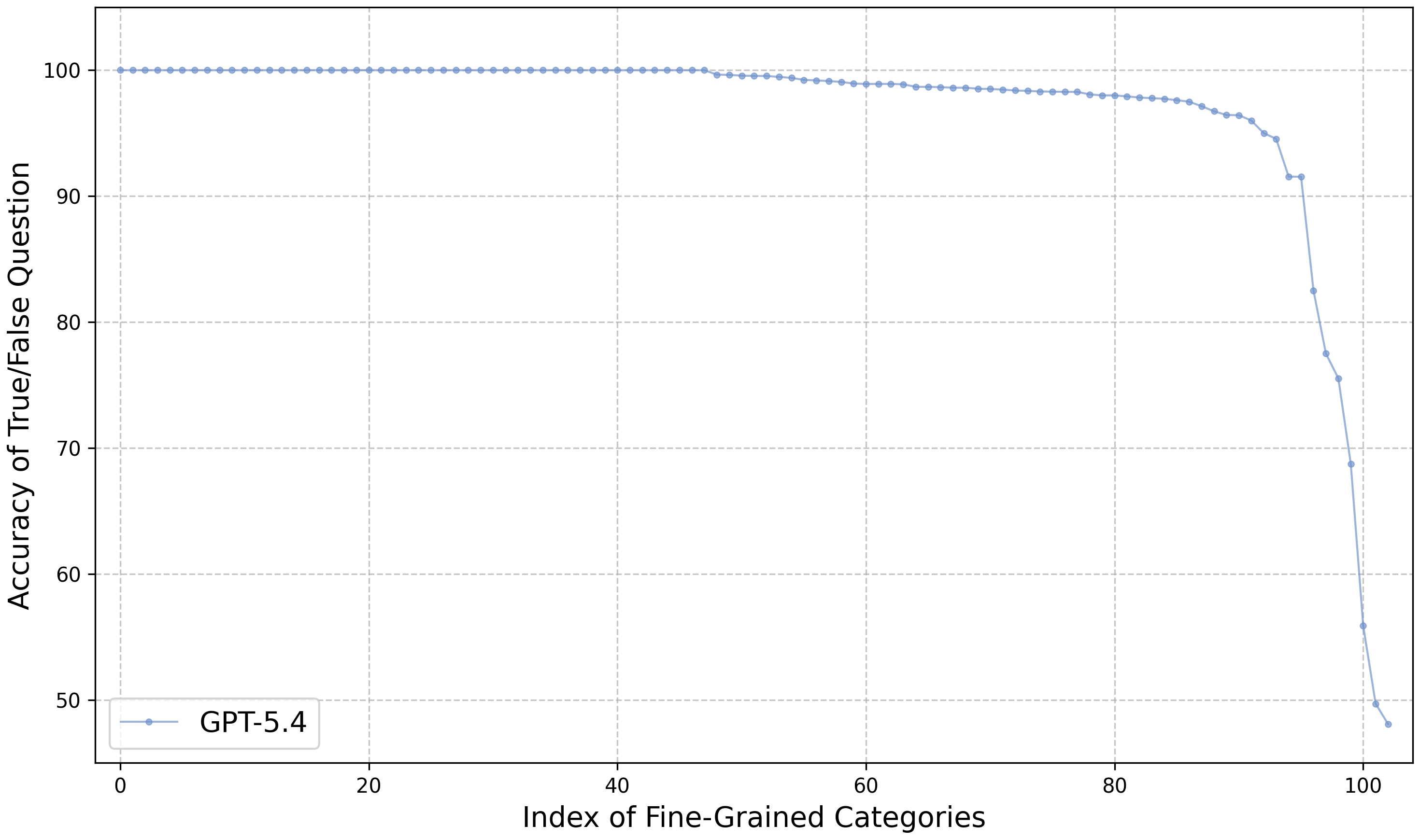}
        \subcaption{GPT-5.4 on \emph{Flowers102}}  
        \label{fig:app_gpt54_flowers102_each}
    \end{minipage}
    \begin{minipage}{0.45\textwidth}
        \centering
        \includegraphics[width=\textwidth,height=4.8cm]{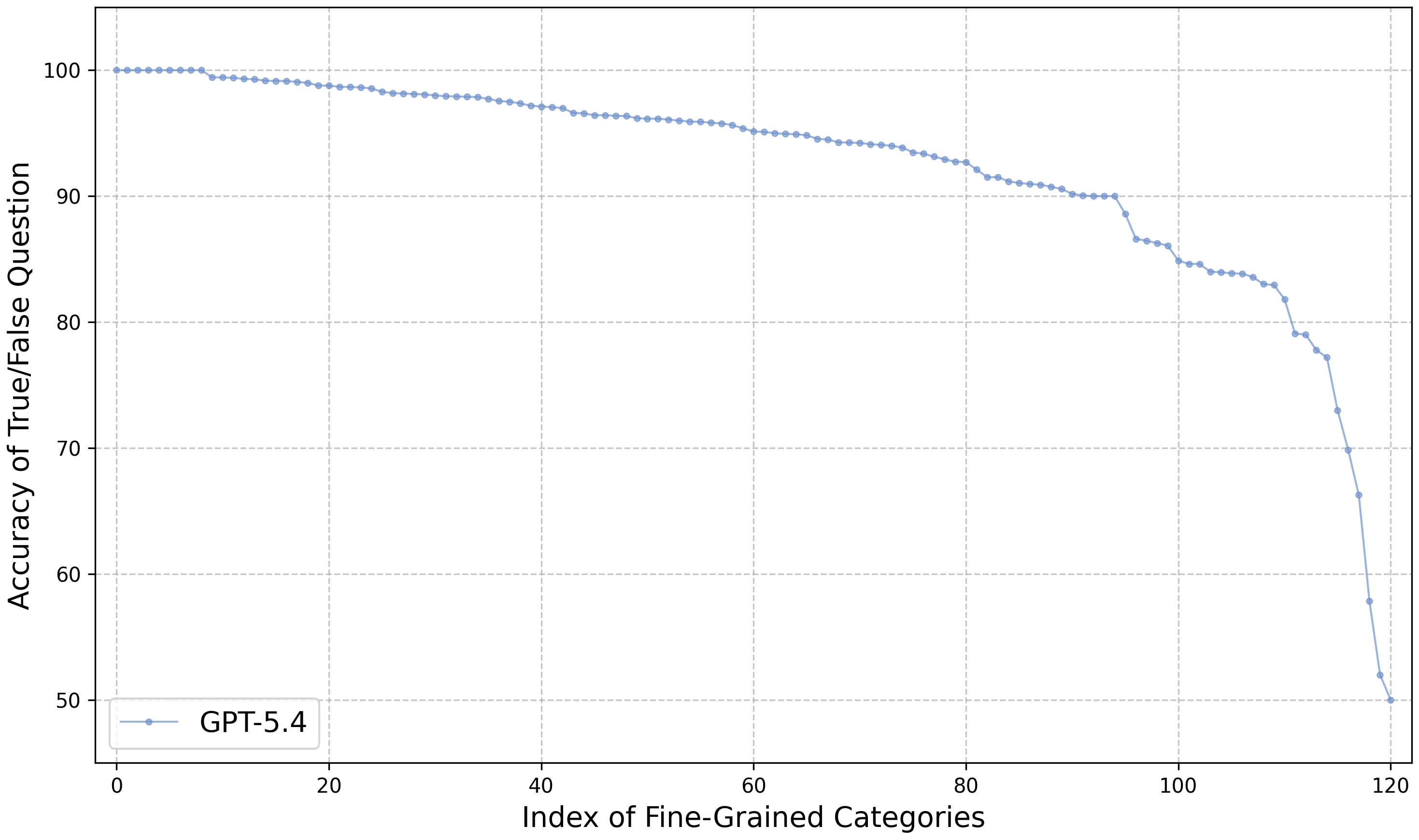}
        \subcaption{Gemini-3.5-flash on \emph{Stanford Dog}}  
        \label{fig:app_gemini_3.5_stanforddog_each}
    \end{minipage}
    \caption{Knowledge bias estimation results of two closed-source models. True/false question accuracy for each category is ranked, with blue dots representing the original model.}
    \label{fig:app_inconsist_close_models_gemini3.5}
    \vspace{-0.6em}
\end{figure*}

\begin{table*}[]
    \centering
    \caption{Linear prob classification results of LLaVA visual features and fine-tuned results of two variants of LLaVA on fine-grained short asnwer questions.}
    \vspace{-0.5em}
    \label{table:llava_granular_retrain_appendix}
    \begin{tabular}{|c|c|c|c|c|c|}
        \hline
        \multirow{2}{*}{\rule{0pt}{8pt}Datasets} & \multicolumn{3}{c|}{\rule{0pt}{8pt}Linear}                                                          & \multicolumn{2}{c|}{Fine-tuned}  \\
        \cline{2-6}
                                  & \rule{0pt}{8pt}Origin                     & Aligned                    & Aligned-FG                 & Vanilla LLaVA & Retrained LLaVA \\ \hline
        \hline
        \emph{\rule{0pt}{8pt}CUB-200-2011}              & \multicolumn{1}{c|}{79.77} & \multicolumn{1}{c|}{73.17} & \multicolumn{1}{c|}{75.06} & 85.60         & 86.32           \\
        \emph{Stanford Dogs}             & \multicolumn{1}{c|}{81.24} & \multicolumn{1}{c|}{78.14} & \multicolumn{1}{c|}{80.69} & 86.49         & 87.58           \\
        \emph{Stanford Cars}             & \multicolumn{1}{c|}{87.57} & \multicolumn{1}{c|}{83.90} & \multicolumn{1}{c|}{85.63} & 90.55         & 91.73           \\
        \emph{Food-101}                  & \multicolumn{1}{c|}{94.27} & \multicolumn{1}{c|}{93.35} & \multicolumn{1}{c|}{94.32} & 95.25         & 95.74          \\
        \hline
    \end{tabular}
\end{table*}

\subsection{Results of Attribute Recognition}\label{app:res_attribute}
Table~\ref{table:attributes_InternVL3} and Table~\ref{table:app_attributes_Qwen2.5_VL} shows the attribute recognition accuracy of InternVL3 and Qwen2.5-VL on the \emph{CUB-200-2011} dataset. The results of LLaVA, BLIP2, InternVL, Gemini-3.5-flash and Gemini-2.0-flash are shown in Table~\ref{table:attributes_LLaVa}, Table~\ref{table:app_attributes_BLIP2}, Table~\ref{table:app_attributes_InternVL}, Table~\ref{table:app_attributes_Gemini35}, and Table~\ref{table:app_attributes_Gemini}.

\begin{table*}[t!]
    \caption{Attribute Recognition Accuracy of LLaVA~\cite{LLaVA15} on the \emph{CUB-200-2011}~\cite{CUB} Dataset (Values in Parentheses Represent the Average Accuracy for Each Attribute).}
    \setlength{\tabcolsep}{4pt}
    \label{table:attributes_LLaVa}
    \centering
    \small
    \begin{tabular}{|c|c|c|c|c|c|c|c|}
        \hline
        \multicolumn{8}{|c|}{\textbf{\rule{0pt}{8pt}Color Attribute (44.34)}} \\
        \hline
        \hline
        \rule{0pt}{8pt}belly color        & 54.79 & back color        & 41.90 & bill color        & 41.44 & breast color      & 49.56 \\
        crown color        & 48.71 & eye color         & 69.27 & forehead color    & 47.03 & leg color         & 35.37 \\
        nape color         & 40.51 & throat color      & 35.40 & under tail color  & 38.88 & underparts color  & 54.81 \\
        upper tail color   & 41.41 & upperparts color  & 34.00 & wing color        & 34.60 & primary color     & 41.77 \\
        \hline
        \hline
        \multicolumn{8}{|c|}{\textbf{\rule{0pt}{8pt}Pattern Attribute (23.69)}} \\
        \hline
        \hline
        \rule{0pt}{8pt}back pattern       & 27.27 & belly pattern     & 26.41 & breast pattern    & 24.24 & head pattern      & 11.35 \\
        tail pattern       & 23.19 & wing pattern      & 29.67 &                   &       &                   &       \\
        \hline
        \hline
        \multicolumn{8}{|c|}{\textbf{\rule{0pt}{8pt}Shape Attribute (14.05)}} \\
        \hline
        \hline
        \rule{0pt}{8pt}bill shape         & 1.39  & shape             & 18.59 & tail shape        & 9.89  & wing shape        & 26.34 \\
        \hline
        \hline
        \multicolumn{4}{|c|}{\textbf{\rule{0pt}{8pt}Length Attribute (15.71)}} & \multicolumn{4}{c|}{\textbf{Size Attribute (49.47)}} \\
        \hline
        \hline
        \multicolumn{2}{|c|}{\rule{0pt}{8pt}bill length} & \multicolumn{2}{c|}{15.71} & \multicolumn{2}{c|}{size} & \multicolumn{2}{c|}{49.47}   \\
        \hline
    \end{tabular}
\end{table*}

\begin{table*}[!htb]
    \caption{Attribute recognition accuracy of BLIP2~\cite{BLIP2} on the \emph{CUB-200-2011}~\cite{CUB} dataset (values in parentheses represent the average accuracy for each attribute).}
    \setlength{\tabcolsep}{4pt}
    \vspace{-0.5em}
    \label{table:app_attributes_BLIP2}
    \centering
    \small
    \begin{tabular}{|c|c|c|c|c|c|c|c|}
        \hline
        \multicolumn{8}{|c|}{\textbf{Color Attribute (37.94)}} \\
        \hline
        \hline
        belly color        & 51.15 & back color        & 39.64 & bill color        & 23.42 & breast color      & 50.17 \\
        crown color        & 42.59 & eye color         & 23.59 & forehead color    & 43.18 & leg color         & 18.77 \\
        nape color         & 41.55 & throat color      & 53.81 & under tail color  & 37.98 & underparts color  & 41.60 \\
        upper tail color   & 37.52 & upperparts color  & 33.01 & wing color        & 31.25 & primary color     & 33.48 \\
        \hline
        \hline
        \multicolumn{8}{|c|}{\textbf{Pattern Attribute (11.34)}} \\
        \hline
        \hline
        back pattern       & 14.66 & belly pattern     & 7.82  & breast pattern    & 9.48  & head pattern      & 2.14  \\
        tail pattern       & 14.21 & wing pattern      & 19.73 &                   &       &                   &       \\
        \hline
        \hline
        \multicolumn{8}{|c|}{\textbf{Shape Attribute (25.05)}} \\
        \hline
        \hline
        bill shape         & 8.84  & shape             & 34.69 & tail shape        & 13.51 & wing shape        & 43.19 \\
        \hline
        \hline
        \multicolumn{4}{|c|}{\textbf{Length Attribute (30.11)}} & \multicolumn{4}{c|}{\textbf{Size Attribute (27.62)}} \\
        \hline
        \hline
        \multicolumn{2}{|c|}{bill length} & \multicolumn{2}{c|}{30.11} & \multicolumn{2}{c|}{size} & \multicolumn{2}{c|}{27.62}   \\
        \hline
    \end{tabular}

\end{table*}

\begin{table*}[!htb]
    \caption{Attribute recognition accuracy of InternVL~\cite{chen2024internvl} on the \emph{CUB-200-2011}~\cite{CUB} dataset (values in parentheses represent the average accuracy for each attribute).}
    \setlength{\tabcolsep}{4pt}
    \vspace{-0.5em}
    \label{table:app_attributes_InternVL}
    \centering
    \small
    \begin{tabular}{|c|c|c|c|c|c|c|c|} 
        \hline
        \multicolumn{8}{|c|}{\textbf{Color Attribute (35.78)}} \\ 
        \hline 
        \hline 
        belly color        & 52.09 & back color        & 33.89 & bill color        & 26.59 & breast color      & 46.58 \\ 
        crown color        & 39.91 & eye color         & 23.68 & forehead color    & 40.83 & leg color         & 32.75 \\ 
        nape color         & 29.66 & throat color      & 30.21 & under tail color  & 32.31 & underparts color  & 50.57 \\ 
        upper tail color   & 33.42 & upperparts color  & 29.64 & wing color        & 27.17 & primary color     & 40.15 \\ 
        \hline
        \hline
        \multicolumn{8}{|c|}{\textbf{Pattern Attribute (34.71)}} \\ 
        \hline 
        \hline 
        back pattern       & 35.57 & belly pattern     & 44.14 & breast pattern    & 42.22 & head pattern      & 11.81 \\ 
        tail pattern       & 35.86 & wing pattern      & 37.31 &                   &       &                   &       \\ 
        \hline
        \hline
        \multicolumn{8}{|c|}{\textbf{Shape Attribute (23.03)}} \\ 
        \hline 
        \hline 
        bill shape         & 12.16 & shape             & 38.08 & tail shape        & 15.49 & wing shape        & 26.43 \\ 
        \hline
        \hline
        \multicolumn{4}{|c|}{\textbf{Length Attribute (29.31)}} & \multicolumn{4}{c|}{\textbf{Size Attribute (47.70)}} \\ 
        \hline 
        \hline 
        \multicolumn{2}{|c|}{bill length}   & \multicolumn{2}{c|}{29.31}  & \multicolumn{2}{c|}{size} & \multicolumn{2}{c|}{47.70}    \\ 
        \hline
    \end{tabular}
\end{table*}

\begin{table*}[!htb]
    \caption{Attribute recognition accuracy of Gemini-3.5-flash~\cite{Gemini} on the \emph{CUB-200-2011}~\cite{CUB} dataset (values in parentheses represent the average accuracy for each attribute).}
    \setlength{\tabcolsep}{4pt}
    \vspace{-0.5em}
    \label{table:app_attributes_Gemini35}
    \centering
    \small
    \begin{tabular}{|c|c|c|c|c|c|c|c|} 
        \hline
        \multicolumn{8}{|c|}{\textbf{Color Attribute (61.18)}} \\ 
        \hline 
        \hline 
        belly color        & 73.40 & back color        & 60.46 & bill color        & 58.99 & breast color      & 68.64 \\ 
        crown color        & 67.71 & eye color         & 50.06 & forehead color    & 67.20 & leg color         & 55.23 \\ 
        nape color         & 61.55 & throat color      & 70.71 & under tail color  & 52.69 & underparts color  & 71.04 \\ 
        upper tail color   & 59.49 & upperparts color  & 51.57 & wing color        & 50.72 & primary color     & 59.40 \\ 
        \hline
        \hline
        \multicolumn{8}{|c|}{\textbf{Pattern Attribute (64.96)}} \\ 
        \hline 
        \hline 
        back pattern       & 64.27 & belly pattern     & 77.81 & breast pattern    & 76.41 & head pattern      & 47.44 \\ 
        tail pattern       & 66.59 & wing pattern      & 57.24 &                   &       &                   &       \\ 
        \hline
        \hline
        \multicolumn{8}{|c|}{\textbf{Shape Attribute (47.15)}} \\ 
        \hline 
        \hline 
        bill shape         & 65.66 & shape             & 60.56 & tail shape        & 23.48 & wing shape        & 38.91 \\ 
        \hline
        \hline
        \multicolumn{4}{|c|}{\textbf{Length Attribute (86.00)}} & \multicolumn{4}{c|}{\textbf{Size Attribute (55.93)}} \\ 
        \hline 
        \hline 
        \multicolumn{2}{|c|}{bill length}   & \multicolumn{2}{c|}{86.00}  & \multicolumn{2}{c|}{size} & \multicolumn{2}{c|}{55.93}    \\ 
        \hline
    \end{tabular}
\end{table*}

\begin{table*}[!htb]
    \caption{Attribute recognition accuracy of Gemini-2.0-flash~\cite{Gemini} on the \emph{CUB-200-2011}~\cite{CUB} dataset (values in parentheses represent the average accuracy for each attribute).}
    \setlength{\tabcolsep}{4pt}
    \vspace{-0.5em}
    \label{table:app_attributes_Gemini}
    \centering
    \small
    \begin{tabular}{|c|c|c|c|c|c|c|c|} 
        \hline
        \multicolumn{8}{|c|}{\textbf{Color Attribute (47.22)}} \\ 
        \hline 
        \hline 
        belly color        & 62.09 & back color        & 36.51 & bill color        & 52.31 & breast color      & 56.01 \\ 
        crown color        & 56.44 & eye color         & 59.57 & forehead color    & 53.55 & leg color         & 40.66 \\ 
        nape color         & 40.40 & throat color      & 60.23 & under tail color  & 40.60 & underparts color  & 59.65 \\ 
        upper tail color   & 39.99 & upperparts color  & 29.66 & wing color        & 29.21 & primary color     & 38.69 \\ 
        \hline
        \hline
        \multicolumn{8}{|c|}{\textbf{Pattern Attribute (56.14)}} \\ 
        \hline 
        \hline 
        back pattern       & 56.26 & belly pattern     & 70.51 & breast pattern    & 66.89 & head pattern      & 39.56 \\ 
        tail pattern       & 52.33 & wing pattern      & 51.26 &                   &       &                   &       \\ 
        \hline
        \hline
        \multicolumn{8}{|c|}{\textbf{Shape Attribute (48.75)}} \\ 
        \hline 
        \hline 
        bill shape         & 61.62 & shape             & 68.20 & tail shape        & 32.13 & wing shape        & 33.04 \\ 
        \hline
        \hline
        \multicolumn{4}{|c|}{\textbf{Length Attribute (71.82)}} & \multicolumn{4}{c|}{\textbf{Size Attribute (52.72)}} \\ 
        \hline 
        \hline 
        \multicolumn{2}{|c|}{bill length}   & \multicolumn{2}{c|}{71.82}  & \multicolumn{2}{c|}{size} & \multicolumn{2}{c|}{52.72}    \\ 
        \hline
    \end{tabular}
    \vspace{-1.2em}
\end{table*}

\begin{table*}[!t]
    \caption{Attribute recognition accuracy of Qwen2.5-VL~\cite{Qwen2.5-VL} on the \emph{CUB-200-2011}~\cite{CUB} dataset (values in parentheses represent the average accuracy for each attribute).}
    \setlength{\tabcolsep}{4pt}
    \vspace{-0.5em}
    \label{table:app_attributes_Qwen2.5_VL}
    \centering
    \small
    \begin{tabular}{|c|c|c|c|c|c|c|c|} 
        \hline
        \multicolumn{8}{|c|}{\textbf{Color Attribute (40.39)}} \\ 
        \hline 
        \hline 
        belly color        & 51.11 & back color        & 32.89 & bill color        & 46.50 & breast color      & 44.84 \\ 
        crown color        & 46.54 & eye color         & 54.85 & forehead color    & 44.57 & leg color         & 37.79 \\ 
        nape color         & 36.49 & throat color      & 40.74 & under tail color  & 34.60 & underparts color  & 50.20 \\ 
        upper tail color   & 34.92 & upperparts color  & 27.20 & wing color        & 26.03 & primary color     & 36.96 \\ 
        \hline
        \hline
        \multicolumn{8}{|c|}{\textbf{Pattern Attribute (45.12)}} \\ 
        \hline 
        \hline 
        back pattern       & 42.66 & belly pattern     & 64.58 & breast pattern    & 59.79 & head pattern      & 14.57 \\ 
        tail pattern       & 45.04 & wing pattern      & 44.11 &                   &       &                   &       \\ 
        \hline
        \hline
        \multicolumn{8}{|c|}{\textbf{Shape Attribute (29.30)}} \\ 
        \hline 
        \hline 
        bill shape         & 15.30 & shape             & 58.17 & tail shape        & 5.63 & wing shape        & 38.10 \\ 
        \hline
        \hline
        \multicolumn{4}{|c|}{\textbf{Length Attribute (63.20)}} & \multicolumn{4}{c|}{\textbf{Size Attribute (52.56)}} \\ 
        \hline 
        \hline 
        \multicolumn{2}{|c|}{bill length}   & \multicolumn{2}{c|}{63.20}  & \multicolumn{2}{c|}{size} & \multicolumn{2}{c|}{52.56}    \\ 
        \hline
    \end{tabular}
\end{table*}

\subsection{Results of visual-side and language-side perturbations.}\label{app:res_visual_side_and_language-side_perturbations}

Table~\ref{table:fg_attack} summarizes LVLM robustness under a representative subset of perturbations; additional Gaussian blur/noise sweeps, background/color corruptions on more datasets, and misleading-prompt evaluations are reported in Appendix Tables~\ref{tab:appendix_bg_color}, \ref{tab:appendix_gb_sp}, and~\ref{tab:appendix_misleading_prompt}.

\begin{table*}[t]
  \centering
  \caption{Extended robustness of LVLMs to \emph{background grayscale} (\mbox{BG-gray}) and \emph{object-centric color jitter} (Color) on Flowers-102 and Stanford Dogs. Each cell reports accuracy as ``multiple-choice / true/false'' (\%). BG-gray grayscale the background region while preserving the segmented foreground; Color perturbs hues/saturation on the foreground object.}
  \label{tab:appendix_bg_color}
  \small
  \setlength{\tabcolsep}{10pt}
  \begin{tabular}{|l|c|c|c|c|c|c|}
    \hline
    \multirow{2}{*}{\rule{0pt}{9pt}Models}
      & \multicolumn{3}{c|}{Flowers-102}
      & \multicolumn{3}{c|}{Stanford Dogs} \\
    \cline{2-7}
    & \rule{0pt}{9pt}Original & BG-gray & Color
      & Original & BG-gray & Color \\
    \hline\hline
    \rule{0pt}{9pt}Qwen2.5-VL
      & 95.69/93.32 & 94.00/92.01 & 83.09/83.62
      & 96.74/94.50 & 95.55/92.80 & 90.12/86.19 \\
    InternVL3
      & 88.24/88.75 & 82.42/84.71 & 74.06/81.05
      & 93.11/92.02 & 91.50/90.99 & 83.90/85.07 \\
    LLaVA-1.5
      & 66.81/76.53 & 62.55/74.11 & 55.37/67.10
      & 68.81/77.45 & 66.60/77.09 & 60.75/73.58 \\
    \hline
  \end{tabular}
\end{table*}

\begin{table*}[t]
  \centering
  \footnotesize
  \setlength{\tabcolsep}{5pt}
  \caption{%
    Robustness of Qwen2.5-VL and InternVL3 under Gaussian blur (GB; \emph{GB-}$k$ with $k\!\in\!\{1,3,5\}$ denotes increasing blur strength) and salt-and-pepper noise (SP; \emph{SP-}$r$ with $r\!\in\!\{5,10,15\}$ denotes noise density in percentage points).
    \emph{Linear} rows report Top-1 accuracy and
    \emph{QA} rows report accuracy as ``multiple-choice\,/\,true-false''.
  }
  \label{tab:appendix_gb_sp}
  \vspace{0.25em}
  \begin{tabular}{|l|c|c|c|c|c|c|c|}
    \hline
    CUB & Ori. & GB-1 & GB-3 & GB-5 & SP-5 & SP-10 & SP-15 \\
    \hline\hline
    \rule{0pt}{7pt}Qwen2.5-VL (\emph{Linear}) & 85.62 & 85.55 & 79.92 & 70.45 & 80.15 & 75.54 & 71.78 \\
    Qwen2.5-VL (\emph{QA}) & 74.04/71.49 & 72.89/71.10 & 70.50/71.06 & 64.29/65.55 & 70.78/70.35 & 68.16/68.33 & 65.64/67.48 \\
    \hline
    InternVL3 (\emph{Linear}) & 69.41 & 68.89 & 54.43 & 40.00 & 67.31 & 61.68 & 56.28 \\
    InternVL3 (\emph{QA}) & 61.18/62.48 & 60.41/62.01 & 59.96/61.93 & 55.45/61.85 & 61.01/62.43 & 60.80/62.30 & 59.51/62.10 \\
    \hline
  \end{tabular}

  \vspace{0.85em}

  \vspace{0.25em}
  \begin{tabular}{|l|c|c|c|c|c|c|c|}
    \hline
    FGVC Aircraft & Ori. & GB-1 & GB-3 & GB-5 & SP-5 & SP-10 & SP-15 \\
    \hline\hline
    \rule{0pt}{7pt}Qwen2.5-VL (\emph{Linear}) & 62.07 & 62.07 & 59.49 & 55.11 & 60.87 & 59.28 & 56.76 \\
    Qwen2.5-VL (\emph{QA}) & 94.84/89.56 & 93.79/87.52 & 87.67/71.02 & 80.83/61.69 & 90.16/83.53 & 87.46/80.35 & 84.07/77.71 \\
    \hline
    InternVL3 (\emph{Linear}) & 45.42 & 45.18 & 42.81 & 34.98 & 43.17 & 40.05 & 38.16 \\
    InternVL3 (\emph{QA}) & 85.48/86.92 & 84.58/86.35 & 80.17/84.01 & 75.46/82.84 & 81.76/84.94 & 79.45/83.68 & 78.25/83.17 \\
    \hline
  \end{tabular}

  \vspace{0.85em}

  \vspace{0.25em}
  \begin{tabular}{|l|c|c|c|c|c|c|c|}
    \hline
    Stanford Dogs & Ori. & GB-1 & GB-3 & GB-5 & SP-5 & SP-10 & SP-15 \\
    \hline\hline
    \rule{0pt}{7pt}Qwen2.5-VL (\emph{Linear}) & 79.07 & 77.82 & 67.07 & 55.48 & 69.11 & 62.89 & 57.16 \\
    Qwen2.5-VL (\emph{QA}) & 96.74/94.50 & 96.60/94.20 & 91.13/87.37 & 82.88/78.38 & 91.43/89.92 & 86.84/86.15 & 81.66/82.77 \\
    \hline
    InternVL3 (\emph{Linear}) & 73.90 & 72.20 & 57.43 & 43.00 & 64.90 & 56.67 & 49.33 \\
    InternVL3 (\emph{QA}) & 93.13/92.02 & 92.93/91.57 & 87.23/88.61 & 78.62/83.24 & 90.71/90.05 & 87.88/88.11 & 84.50/86.06 \\
    \hline
  \end{tabular}
\end{table*}

\begin{table*}[t]
  \centering
  \footnotesize
  \setlength{\tabcolsep}{4pt}
  \caption{%
    Effect of misleading prompts on multiple-choice/true-false accuracy. Cells are formatted as ``multiple-choice\,/\,true-false''.
  }
  \label{tab:appendix_misleading_prompt}

  \begin{tabular}{|l|c|c|c|c|c|c|c|c|c|c|}
    \hline
    \multirow{2}{*}{\rule{0pt}{7pt}Model}
      & \multicolumn{2}{c|}{\rule{0pt}{7pt}CUB}
      & \multicolumn{2}{c|}{FGVC Aircraft}
      & \multicolumn{2}{c|}{Stanford Dogs}
      & \multicolumn{2}{c|}{Caltech-101}
      & \multicolumn{2}{c|}{CIFAR-100} \\
    \cline{2-11}
    & \rule{0pt}{8pt}Original & Misleading
      & Original & Misleading
      & Original & Misleading
      & Original & Misleading
      & Original & Misleading \\
    \hline\hline
    \rule{0pt}{8pt}Qwen2.5-VL
      & 74.04/71.49 & 63.01/28.69
      & 94.84/89.56 & 82.12/51.37
      & 96.74/94.50 & 92.45/53.10
      & 99.66/97.80 & 98.35/50.76
      & 91.24/81.87 & 64.64/58.15 \\
    InternVL3
      & 61.18/62.48 & 51.71/41.54
      & 85.48/86.92 & 77.77/60.73
      & 93.13/92.02 & 88.45/74.90
      & 99.62/98.31 & 98.71/80.14
      & 93.81/90.06 & 89.60/54.41 \\
    LLaVA-1.5
      & 44.55/58.84 & 3.59/18.57
      & 58.75/77.62 & 4.35/32.13
      & 68.81/77.45 & 6.12/28.87
      & 92.67/86.49 & 68.58/45.60
      & 92.67/86.49 & 68.58/45.60 \\
    \hline
  \end{tabular}
\end{table*}

\section{Machine-oriented Evaluations}\label{app:machine_oriented_evaluations}

\subsection{Qualitative Analysis of Features from Contrastive Training Paradigms and others} \label{app:quali_analy_contras_vs_gen}

\begin{figure*}[!htb]
    \centering
    {\includegraphics[width=0.9\textwidth]{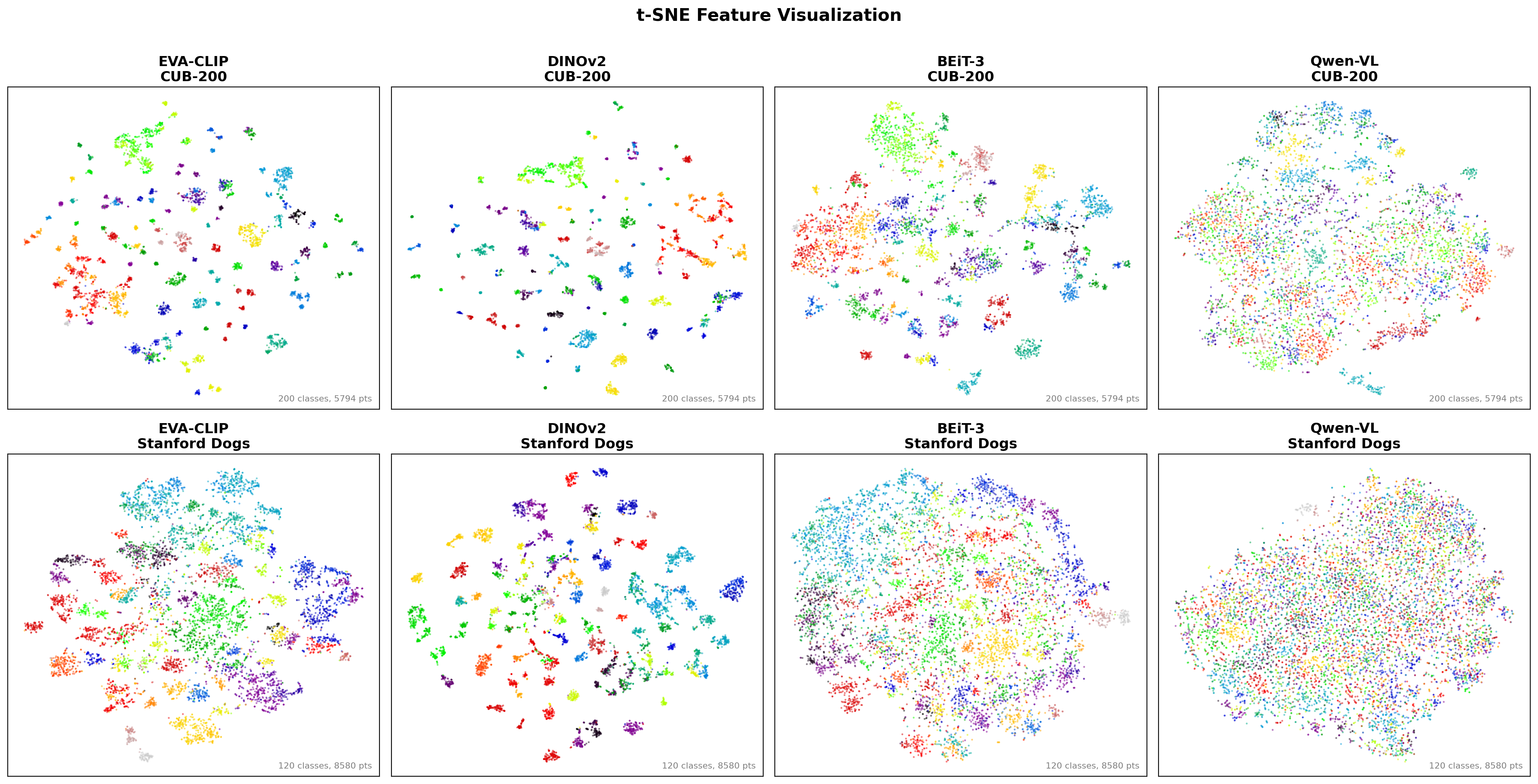}}
    \caption{$t$-SNE visualization of visual features on \emph{CUB-200-2011} and \emph{Stanford Dogs}. Features learned with contrastive paradigms (\emph{e.g.}, EVA-CLIP and DINOv2) form more compact and better-separated class clusters than those learned with reconstruction- or generation-based paradigms (\emph{e.g.}, BEiT-3 and Qwen-VL), indicating stronger fine-grained discriminability in the embedding space.}
    \label{fig:t-SNE_contras_vs_gen_and_recon}
\end{figure*}

\begin{figure*}[t]
    \centering
    \includegraphics[width=0.95\textwidth]{figure/patch_sim_1.pdf}

    \vspace{0.8em}

    \includegraphics[width=0.95\textwidth]{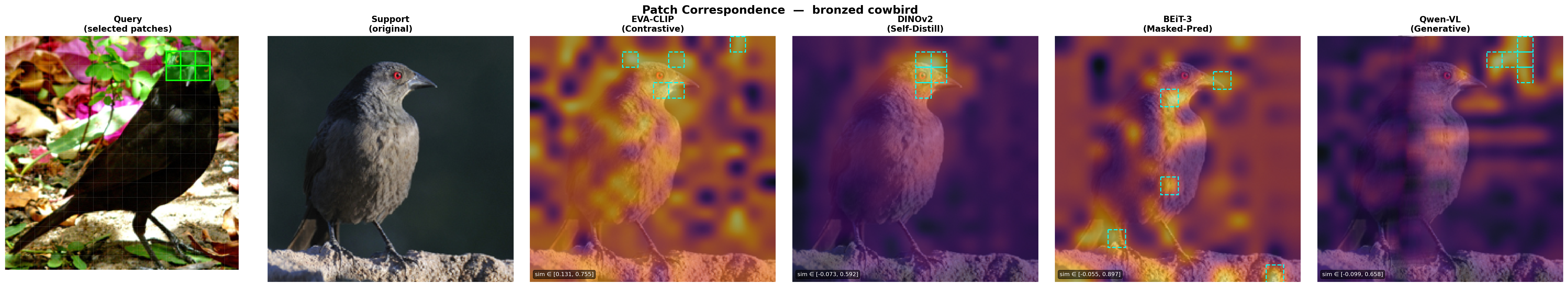}
    \caption{Patch-level correspondence analysis on fine-grained bird images. Given selected query patches, contrastive features (\emph{e.g.}, EVA-CLIP) retrieve semantically more consistent corresponding regions in support images, while reconstruction- and generation-based features (\emph{e.g.}, BEiT-3 and Qwen-VL) are more easily distracted by background patterns or semantically irrelevant regions. This suggests that contrastive learning yields more stable part-level semantic representations for fine-grained recognition.}
    \label{fig:patch_corespondence}
    \vspace{-1.0em}
\end{figure*}

Figure~\ref{fig:t-SNE_visualization} and Figure~\ref{fig:patch_visualization} illustrate how alternative training paradigms reshape learned visual representations, while Figure~\ref{fig:t-SNE_contras_vs_gen_and_recon} and Figure~\ref{fig:patch_corespondence} provide complementary visualization examples under the same setting.

\subsection{Qualitative Analysis of Granularity Inconsistency in LVLM Alignment Data} \label{app:quali_analy_in_granularity}

\begin{figure*}[!htb]
    \centering
    {\includegraphics[width=0.8\textwidth]{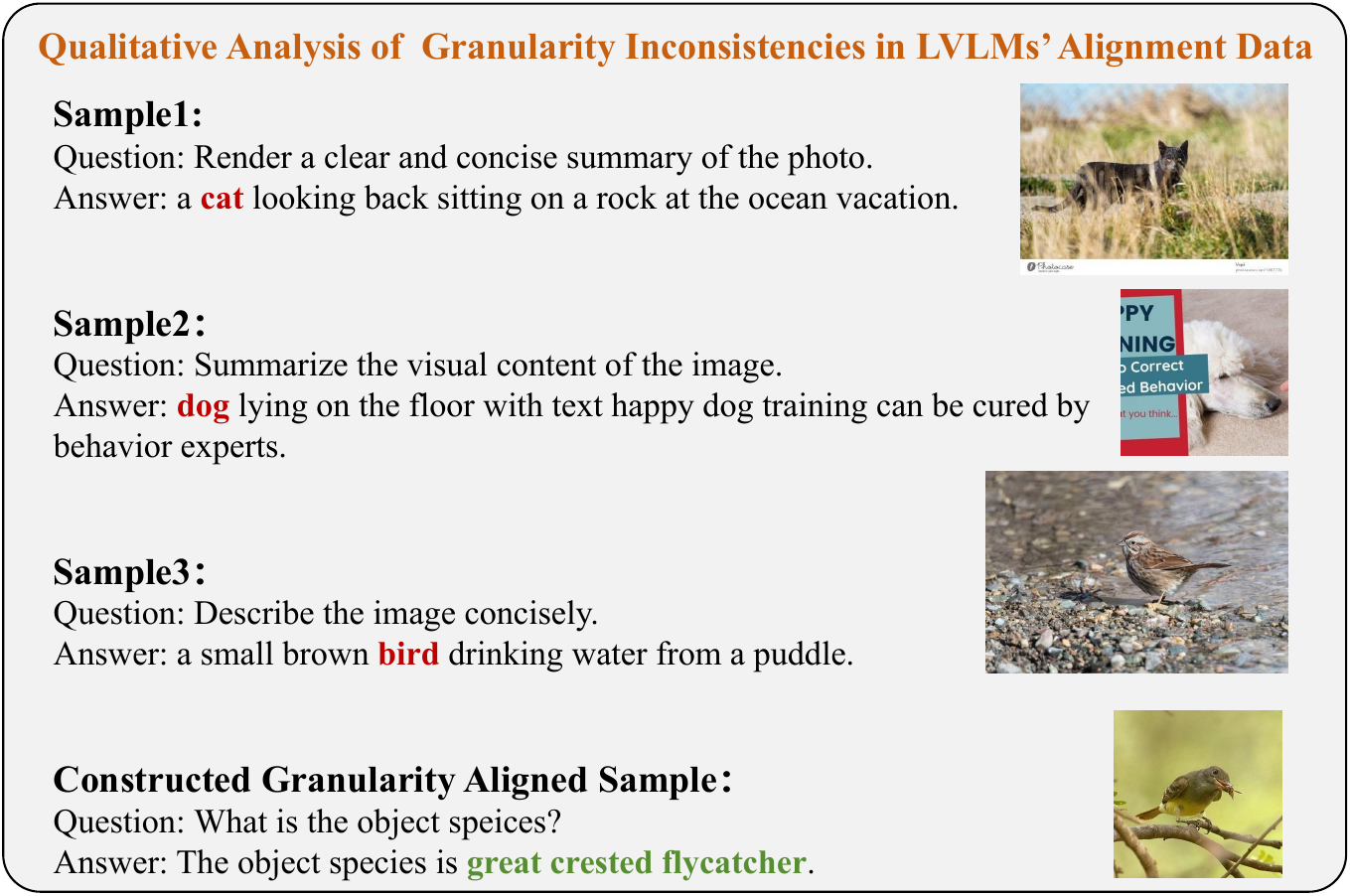}}
    \caption{Qualitative analysis of granularity inconsistencies in LVLMs’ alignment data and a constructed sample of properly aligned granularity.}
    \label{fig:app_dis_match_granularity}
\end{figure*}

In the LVLM's alignment data, we observe a phenomenon of granularity inconsistency, where fine-grained objects in images are paired with coarse-grained textual descriptions. Figure~\ref{fig:app_dis_match_granularity} shows some examples of granularity inconsistency, as well as a constructed sample of properly aligned granularity.

In practice, ensuring fully consistent fine-grained granularity across all image-text pairs is often infeasible, especially when relying on web-scale or weakly labeled data. In our retraining experiment in Table~\ref{table:llava}, we made efforts to construct more consistent alignment data, but some residual mismatch may still exist.

\subsection{Improving the fine-grained discriminability of visual features during the alignment stage can enhance LVLM performance on fine-grained tasks.} \label{app:granular_retrain}

In Table~\ref{table:llava}, we can find that the alignment strategy might impair the fine-grained discriminability of visual features. We then conduct further analysis and find that improving the fine-grained discriminability of visual features during the alignment stage can enhance LVLM performance on fine-grained tasks.

Specifically, we compare the two variants of LLaVA from Table~\ref{table:llava} on fine-grained short-answer questions: (1) Vanilla LLaVA, where the vision-language alignment is trained on image-text pairs with granularity inconsistencies, (2) Retrained LLaVA, where the alignment module is trained on data with matched granularity.

The results in Table~\ref{table:llava_granular_retrain_appendix} show that Retrained LLaVA consistently outperforms Vanilla LLaVA over all datasets, indicating that improving the fine-grained discriminability of visual features during the alignment stage can enhance LVLM performance on fine-grained tasks.

Building on this finding, we believe that incorporating contrastive learning objectives (e.g., patch- or region-level contrastive loss) during the alignment stage may further help preserve discriminative visual information.

\subsection{Results of Classification Across Multi-categories} \label{app:res_cls_multi_categories}

\begin{figure*}[!htb]
    \centering
    {\includegraphics[width=0.99\textwidth]{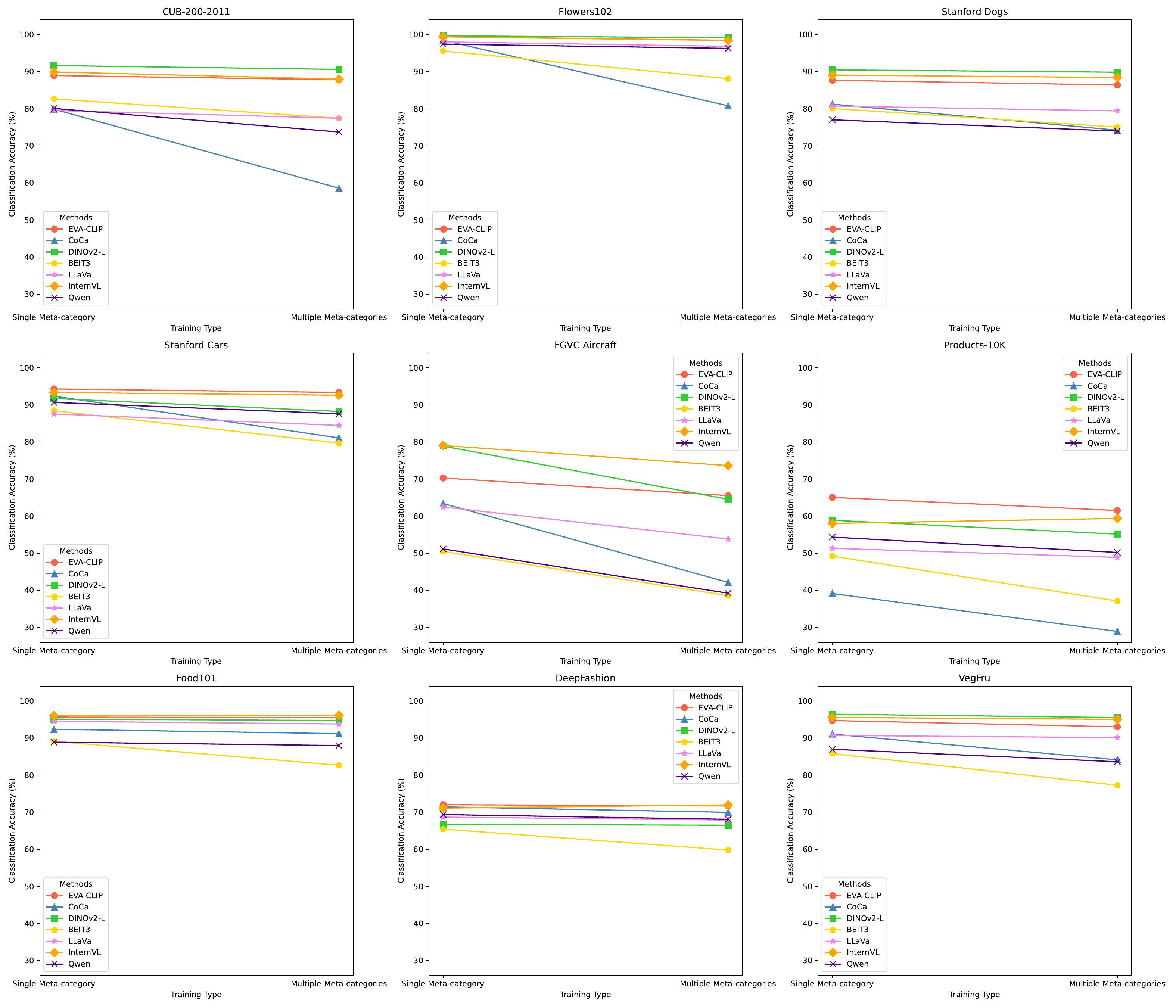}}
    \caption{Classification results of LVLM visual features on fine-grained datasets. ``Single" denotes accuracy from training on a single meta-category, while ``Multiple" reflects accuracy from training on a unified dataset combining multiple meta-categories.}
    \label{fig:combine_cls_9}
    \vspace{-0.2em}
\end{figure*}

In Figure~\ref{fig:combine_cls}, we have shown the classification accuracy both within a single super-category and across multiple meta-categories in three datasets. Here, in Figure~\ref{fig:combine_cls_9}, we include more results on nine fine-grained datasets. As shown in the results, EVA-CLIP, trained with contrastive paradigm, maintains a higher score in classification across multiple meta-categories compared to Qwen and BEiT3, which are trained with generative and reconstruction paradigms.

\begin{figure*}[t]
    \centering

    \begin{minipage}[b]{0.30\textwidth}
        \centering
        \includegraphics[width=\textwidth]{figure/aligned_img_txt_visualization/cub_mlp_vs_text_3d_558k_short.png}
        \vspace{0.15em}
        
        \small (a) CUB: Original
    \end{minipage}
    \hspace{0.02\textwidth}
    \begin{minipage}[b]{0.30\textwidth}
        \centering
        \includegraphics[width=\textwidth]{figure/aligned_img_txt_visualization/cub_mlp_vs_text_3d_558k_long.png}
        \vspace{0.15em}
        
        \small (b) CUB: Aligned-Recap
    \end{minipage}
    \hspace{0.02\textwidth}
    \begin{minipage}[b]{0.30\textwidth}
        \centering
        \includegraphics[width=\textwidth]{figure/aligned_img_txt_visualization/cub_mlp_vs_text_3d_fg.png}
        \vspace{0.15em}
        
        \small (c) CUB: Aligned-FG
    \end{minipage}

    \vspace{0.8em}

    \begin{minipage}[b]{0.30\textwidth}
        \centering
        \includegraphics[width=\textwidth]{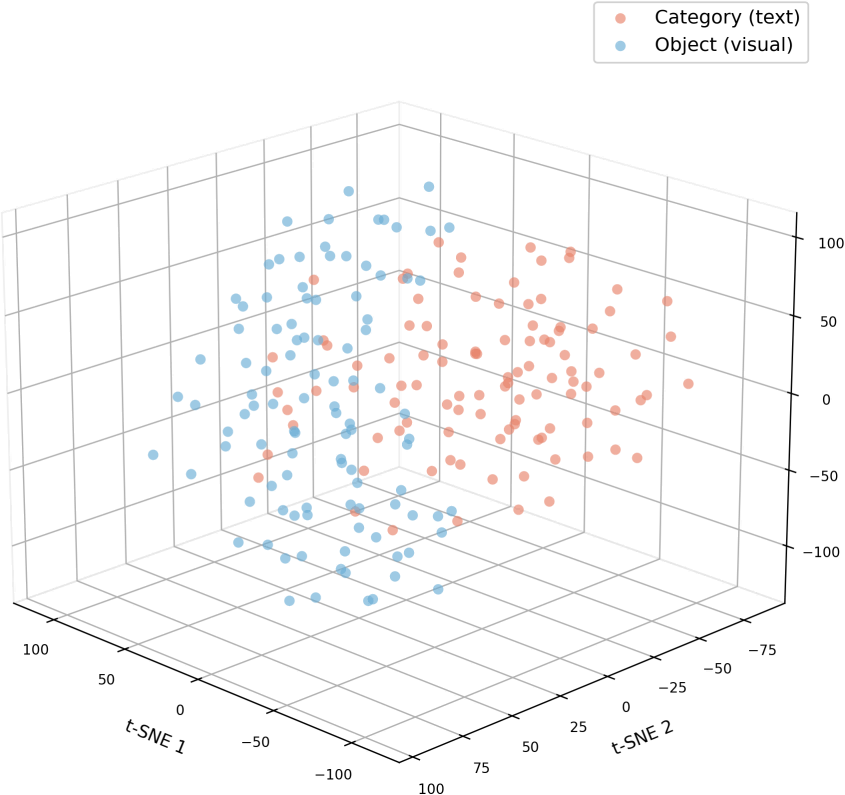}
        \vspace{0.15em}
        
        \small (d) Flowers102: Original
    \end{minipage}
    \hspace{0.02\textwidth}
    \begin{minipage}[b]{0.30\textwidth}
        \centering
        \includegraphics[width=\textwidth]{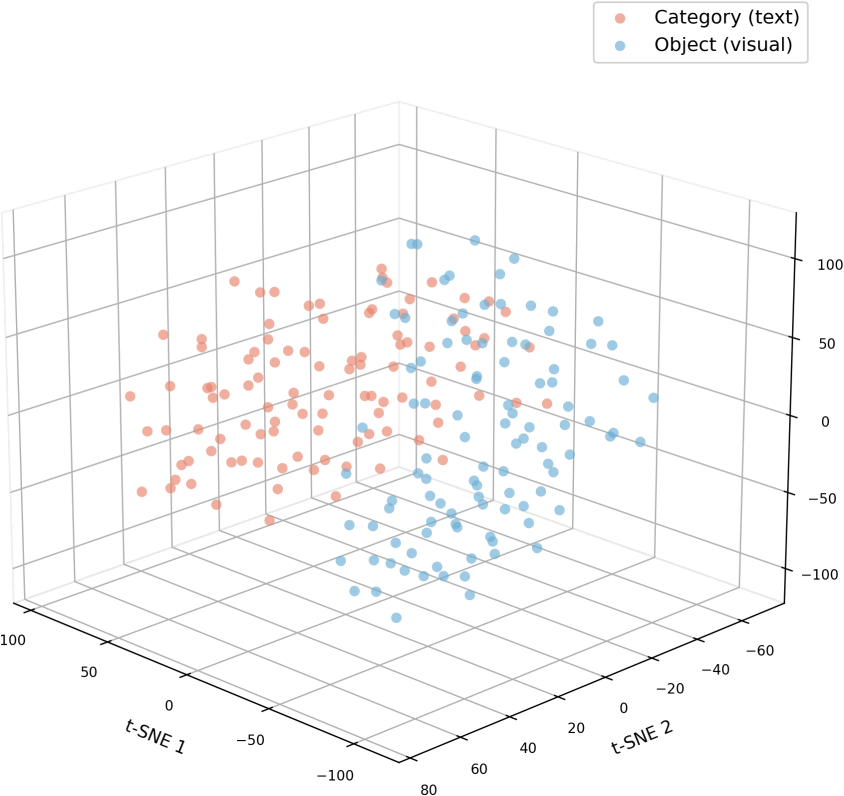}
        \vspace{0.15em}
        
        \small (e) Flowers102: Aligned-Recap
    \end{minipage}
    \hspace{0.02\textwidth}
    \begin{minipage}[b]{0.30\textwidth}
        \centering
        \includegraphics[width=\textwidth]{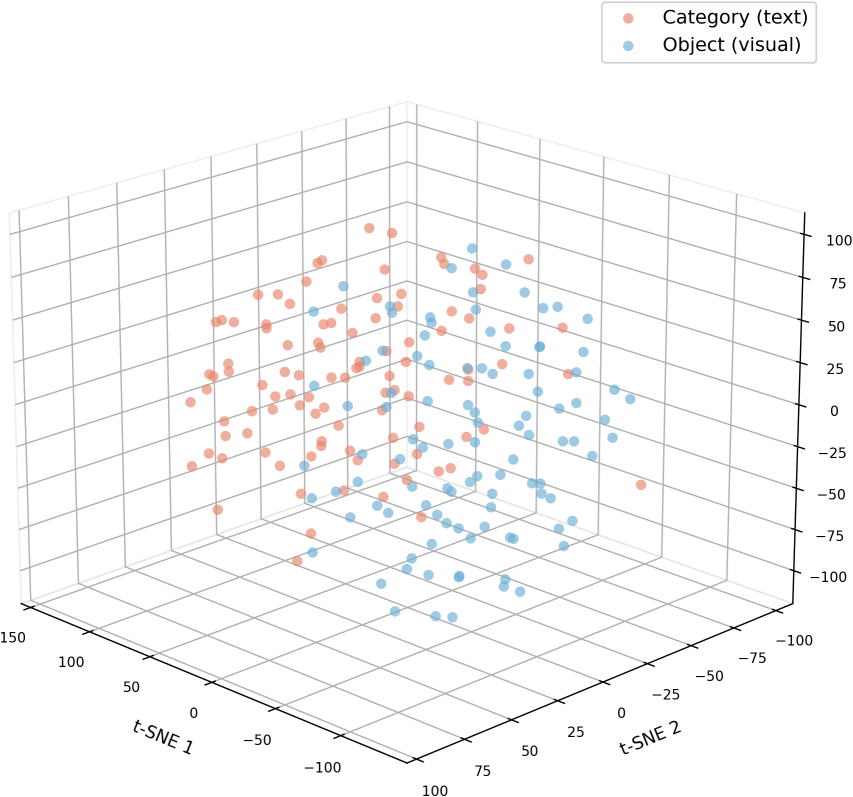}
        \vspace{0.15em}
        
        \small (f) Flowers102: Aligned-FG
    \end{minipage}

    \vspace{0.8em}

    \begin{minipage}[b]{0.30\textwidth}
        \centering
        \includegraphics[width=\textwidth]{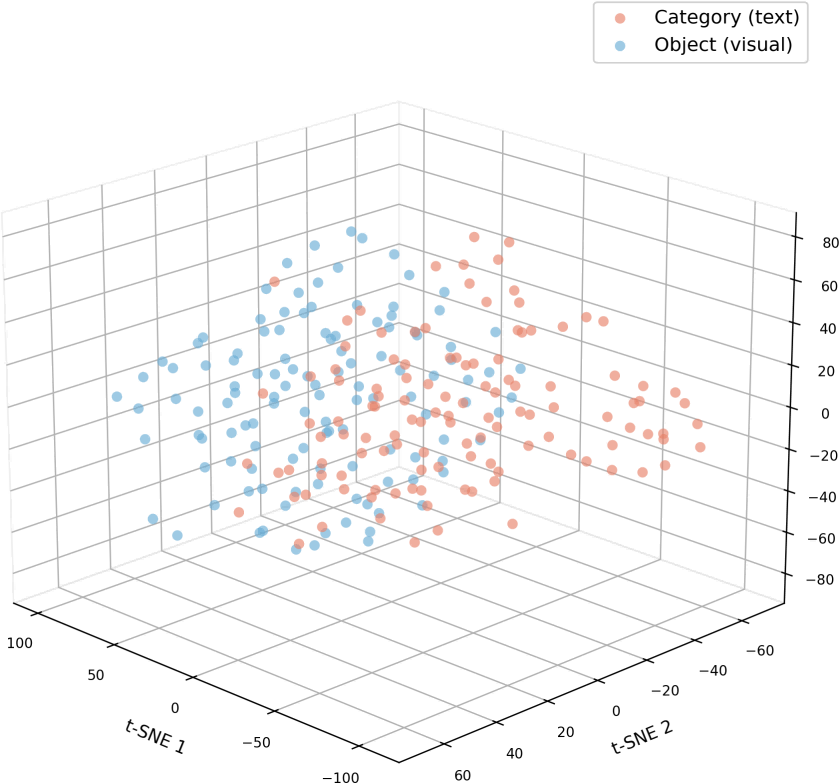}
        \vspace{0.15em}
        
        \small (g) Stanford Dogs: Original
    \end{minipage}
    \hspace{0.02\textwidth}
    \begin{minipage}[b]{0.30\textwidth}
        \centering
        \includegraphics[width=\textwidth]{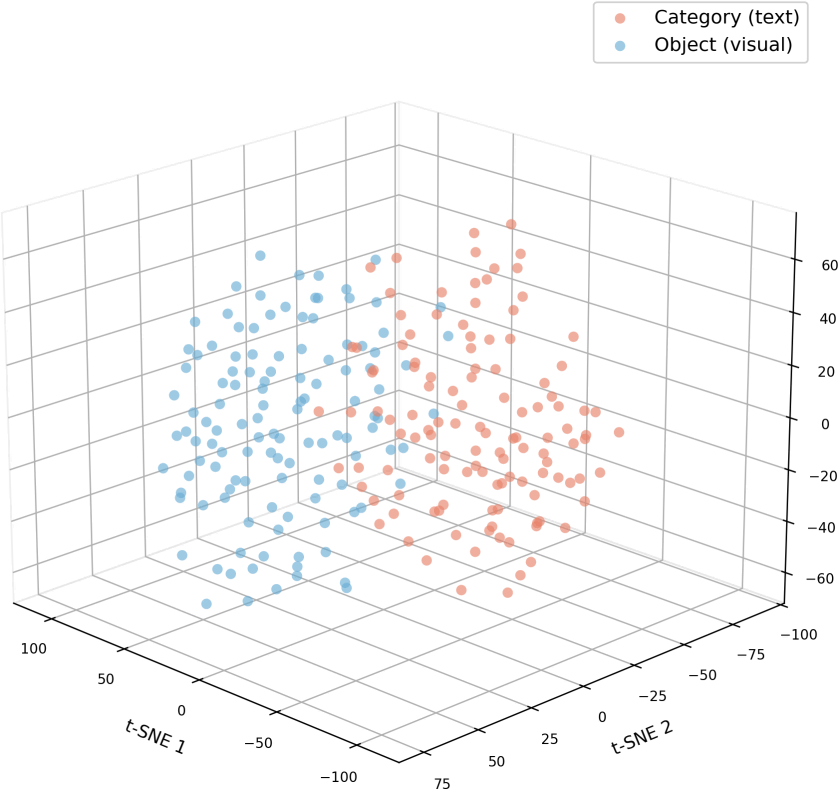}
        \vspace{0.15em}
        
        \small \mbox{(h) Stanford Dogs: Aligned-Recap}
    \end{minipage}
    \hspace{0.02\textwidth}
    \begin{minipage}[b]{0.30\textwidth}
        \centering
        \includegraphics[width=\textwidth]{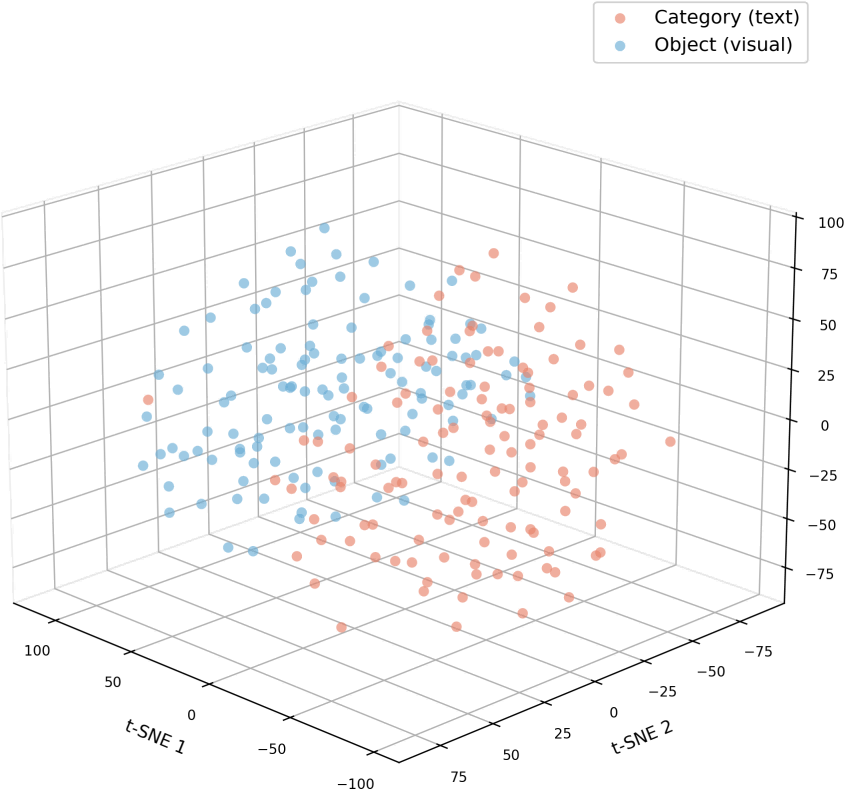}
        \vspace{0.15em}
        
        \small (i) Stanford Dogs: Aligned-FG
    \end{minipage}

    \caption{Visualization of aligned visual features and category text embeddings under different alignment settings. Fine-grained category-level alignment brings visual features closer to their corresponding category embeddings, improving semantic association in fine-grained recognition.}
    \label{fig:appdix_aligned_img_txt_visualization}
    \vspace{-1.0em}
\end{figure*}

\end{document}